\documentclass[12pt]{article}
\usepackage{amsthm,amsmath,amsfonts,amssymb,mathrsfs}
\usepackage{graphicx}
\usepackage{enumerate}
\usepackage{tikz}
\usepackage[numbers]{natbib}
\usepackage{url} 
\usepackage{algorithm}
\usepackage{algpseudocode}
\usepackage{multirow}
\usepackage{epstopdf}
\usepackage{epsfig}
\usepackage{subcaption,float}
\usepackage{pgfplots}
\usepackage{dsfont,stmaryrd,bm,comment}
\usepackage{booktabs}
\usepackage{caption}
\usepackage{subcaption}
\usepackage{tabularx}   
\usepackage{adjustbox}  
\newcolumntype{Y}{>{\raggedright\arraybackslash\hspace{0pt}}X}
\usepackage{rotating}

\usepackage[toc,page,header]{appendix}
\usepackage{minitoc}

\usepackage{hyperref}
\hypersetup{
    colorlinks=true,
    linkcolor=blue,
    filecolor=magenta,
    urlcolor=cyan,
}

\usepackage{cleveref}
\numberwithin{equation}{section}

\theoremstyle{plain}
\newtheorem{theorem}{Theorem}[section]
\newtheorem{lemma}{Lemma}[section] 
\newtheorem{corollary}{Corollary}[section]

\theoremstyle{definition}
\newtheorem{assumption}{Assumption}
\newtheorem{definition}{Definition}

\crefname{theorem}{Theorem}{Theorems}
\crefname{section}{Section}{Sections}
\crefname{figure}{Figure}{Figures}
\crefname{subfigure}{Figure}{Figures}
\crefname{table}{Table}{Tables}
\crefname{algorithm}{Algorithm}{Algorithms}
\crefname{lemma}{Lemma}{Lemmas}
\crefname{assumption}{Assumption}{Assumptions}
\crefname{corollary}{Corollary}{Corollaries}
\crefname{definition}{Definition}{Definitions}
\crefname{example}{Example}{Examples}
\crefname{proposition}{Proposition}{Propositions}
\crefname{appendix}{Appendix}{Appendices}
\crefname{subappendix}{Appendix}{Appendices}

\newcommand{\bed}{\begin{definition}}
\newcommand{\eed}{\end{definition}}

\newcommand{\bitem}{\begin{itemize}}
\newcommand{\eitem}{\end{itemize}}

\newcommand{\beqn}{\begin{equation}}
\newcommand{\eeqn}{\end{equation}}
\newcommand{\balign}{\begin{align}}
\newcommand{\ealign}{\end{align}}

\newcommand{\beq}{\begin{equation}}
\newcommand{\eeq}{\end{equation}}

\newcommand{\diag}{\mathrm{diag}}

\newcommand{\blind}{1}

\newcommand{\calR}{{\cal R}}

\newcommand{\resetspacing}{\spacingset{1.75}}

\usepackage[normalem]{ulem} 

\allowdisplaybreaks

\makeatletter
\renewcommand{\section}{\@startsection{section}{1}{0pt}%
  {\baselineskip}{0.5\baselineskip}{\large\bfseries}}
\renewcommand{\subsection}{\@startsection{subsection}{2}{0pt}%
  {0.5\baselineskip}{0.3\baselineskip}{\normalsize\bfseries}}
\makeatother

\begin{document}



\def\spacingset#1{\renewcommand{\baselinestretch}%
{#1}\small\normalsize} \spacingset{1}


\if1\blind  
{
  \title{\bf Poisson-Process Topic Model for Integrating Knowledge from Pre-trained Language Models}
  \author{Morgane Austern, Yuanchuan Guo, Zheng Tracy Ke, and Tianle Liu\\ \\
  Harvard University}
} 
\maketitle
\fi

\if0\blind 
{
  \bigskip
  \bigskip
  \bigskip
  \begin{center}
    {\Large\bf Poisson-Process Topic Model for Integrating Knowledge from Pre-trained Language Models}
\end{center}
  \medskip
} \fi

\bigskip
\begin{abstract}

Topic modeling is traditionally applied to word counts without accounting for the context in which words appear. Recent advancements in large language models (LLMs) offer contextualized word embeddings, which capture deeper meaning and relationships between words. We aim to leverage such embeddings to improve topic modeling.

We use a pre-trained LLM to convert each document into a sequence of word embeddings. This sequence is then modeled as a Poisson point process, with its intensity measure expressed as a convex combination of $K$ base measures, each corresponding to a topic. To estimate these topics, we propose a flexible algorithm that integrates  traditional topic modeling methods, enhanced by net-rounding applied before and kernel smoothing applied after. One advantage of this framework is that it treats the LLM as a black box, requiring no fine-tuning of its parameters. Another advantage is its ability to seamlessly integrate any traditional topic modeling approach as a plug-in module, without the need for modifications. 

Assuming each topic is a $\beta$-H\"{o}lder smooth intensity measure on the embedded space, we establish the rate of convergence of our method. We also provide a minimax lower bound and show that the rate of our method matches with the lower bound when $\beta\leq 1$. Additionally, we apply our method to two real datasets, providing evidence that it offers an advantage over traditional topic modeling approaches.
\end{abstract}

\vfill

\newpage
\spacingset{1.75} 

\doparttoc 
\faketableofcontents 

\part{} 

\section{Introduction} \label{sec:Intro}
Topic modeling \citep{hofmann1999,blei2003latent} seeks to uncover latent thematic structures in text. Traditional approaches to topic modeling operate on word count data, treating topics as categorical distributions over the vocabulary and assuming a bag-of-words model for each document. Specifically, consider a corpus of \( n \) documents with a vocabulary of \( p \) unique words. Let \( X = [X_1, X_2, \ldots, X_n] \in \mathbb{R}^{p \times n} \) be the word count matrix, where \( X_i(j) \) is the count of the \( j \)-th word in the \( i \)-th document. Suppose the corpus discusses $K$ different topics represented by \( A_1, A_2, \ldots, A_K \in \mathbb{R}^p \), where each $A_k$ is the probability mass function (PMF) of a categorical distribution over the vocabulary. The classical topic model assumes that \( X_1, X_2, \ldots, X_n \) are generated independently satisfying that 
\begin{equation}\label{traditional-TM}
\textstyle X_i \sim \mathrm{Multinomial}(N_i, \; \Omega_i), \qquad \text{with} \quad \Omega_i = \sum_{k=1}^K w_i(k) A_k. 
\end{equation}
In this model, \( N_i \) is the length of the \( i \)-th document, and \( \Omega_i \in \mathbb{R}^p \) contains the population word frequencies; $\Omega_i$  is expressed as a convex combination of the $K$ topic vectors, where \( w_i \in \mathbb{R}^K \) is the weight vector whose entries are the fractional weights that this document puts on different topics. Many methods have been proposed for estimating parameters of this model (e.g., \cite{blei2003latent,arora2013practical,bing2020fast,ke2022using, ke2023recent}). 

However, the reliance on word counts in traditional topic modeling approaches limits their ability to capture the rich semantic information of words and the contexts in which they occur. In this paper, we address this limitation by leveraging advancements in pre-trained large language models (LLMs). LLMs are large, transformer-based neural networks pre-trained on extensive text corpora using self-supervised learning techniques, such as masked token prediction or next-token prediction. Many pre-trained LLMs (e.g., BERT \citep{devlin2019bert}, LLaMA \citep{touvron2023llama}) are open-source, providing access to their neural architectures and trained parameters, making them highly adaptable for downstream tasks. Notably, models like BERT have proven highly effective in generating contextualized embeddings for individual words, enabling richer semantic representation within a text corpus.

To illustrate, each document \( i \) is transformed by a pre-trained LLM into a sequence of \( N_i \) word embeddings, \( \ z_{i1}, z_{i2}, \ldots, z_{iN_i}\in\mathbb{R}^d \), which are extracted from the outputs of final layers of the LLM's neural network. These embeddings offer two key advantages. First, they capture semantic relationships between words. For example, \cite{mikolov2013distributed} observe that  relationships such as ``country-currency'' and ``female-male'' are well-preserved in the embedded space.
Second, they retain contextual information \citep{vaswani2017attention},  
where a same word in the vocabulary can be mapped to different embedding vectors across documents, reflecting its varying contexts. 
Thanks to these benefits, word embeddings generated by an LLM encapsulate substantially richer information than word-count-based representations.

To incorporate the concept of contextualized word embeddings, we propose a new topic modeling framework, where a ``document'' is represented as a sequence of vectors drawn from a point process.
Specifically, let \( \mathcal{Z} \subset \mathbb{R}^d \) denote the space containing all embedding vectors. In this framework, each ``topic'' is treated as a probability measure on \( \mathcal{Z} \), denoted by \( \mathcal{A}_k(z) \), which satisfies \( \int_{\mathcal{Z}} \mathcal{A}_k(z) \, dz = 1 \) for \( 1 \leq k \leq K \). For each document \( i \), let \( w_i \in \mathbb{R}^K \) denote its weight vector, and define:  
\vspace{-.5em}
\begin{equation} \label{PTM-1}
\textstyle\Omega_i(z) = \sum_{k=1}^K w_i(k)\mathcal{A}_k(z), \qquad \mbox{for } 1 \leq i \leq n.  
\end{equation}  
Then, \( \Omega_i(\cdot) \) is also a probability measure on \( \mathcal{Z} \).  
Let $ N >0$ be a document length parameter. 
We assume that the \( n \) documents are generated independently, 
and for each document $i$, 
\vspace{-.6em}
\begin{equation} \label{PTM-2}
z_{i1}, z_{i2}, \ldots z_{iN_i} \text{ follow a Poisson point process with an intensity measure } N\Omega_i(\cdot).
\vspace{-.4em}
\end{equation}  
We refer to the model defined by \eqref{PTM-1} and \eqref{PTM-2} as the \emph{Poisson-Process Topic Model (PPTM)}.  
 

PPTM integrates two complementary modeling strategies. The first is {\it context-aware topic definitions}. In this model, a topic is no longer a discrete distribution over a fixed vocabulary; instead, it is modeled as a continuous distribution in the embedding space. Because the embeddings are contextualized, the same word may be associated with different embedding vectors $z$ across contexts, leading to different values of $\mathcal{A}_1(z), \mathcal{A}_2(z), \ldots, \mathcal{A}_K(z)$. Consequently, a word's relevance to different topics becomes context-dependent. This context awareness not only distinguishes our approach from classical word-count-based topic modeling but also provides a key advantage over recent neural topic modeling (see Section~\ref{subsec:literature} for further discussion).

The second strategy is {\it point-process modeling of word embeddings}. Most pre-trained LLMs are based on the transformer architecture \citep{vaswani2017attention}, which exhibits permutation invariance after positional encoding---a property we discuss in more detail in Section~\ref{sec:Model}. This invariance motivates treating the transformer-generated word embeddings within a document as a ``bag of vectors." Point processes provide a natural framework for modeling such unordered collections. In this work, we adopt the Poisson point process due to its simplicity and widespread use, though other point-process models could be incorporated into our framework as well.

Estimating the parameters of PPTM is a nontrivial problem. As we show in Section~\ref{sec:Method}, it can be viewed as an nonparametric version of the linear unmixing problem \citep{bioucas2012hyperspectral},  in which a large collection of nonparametric densities must be decomposed into a small number of underlying basis densities. This task is substantially more challenging than classical nonparametric density estimation.

\subsection{Our Results and Contributions} \label{subsec:outline}

\begin{figure}[tb!]
\centering
\includegraphics[width=.75\textwidth, trim=0 5 0 0, clip=true]{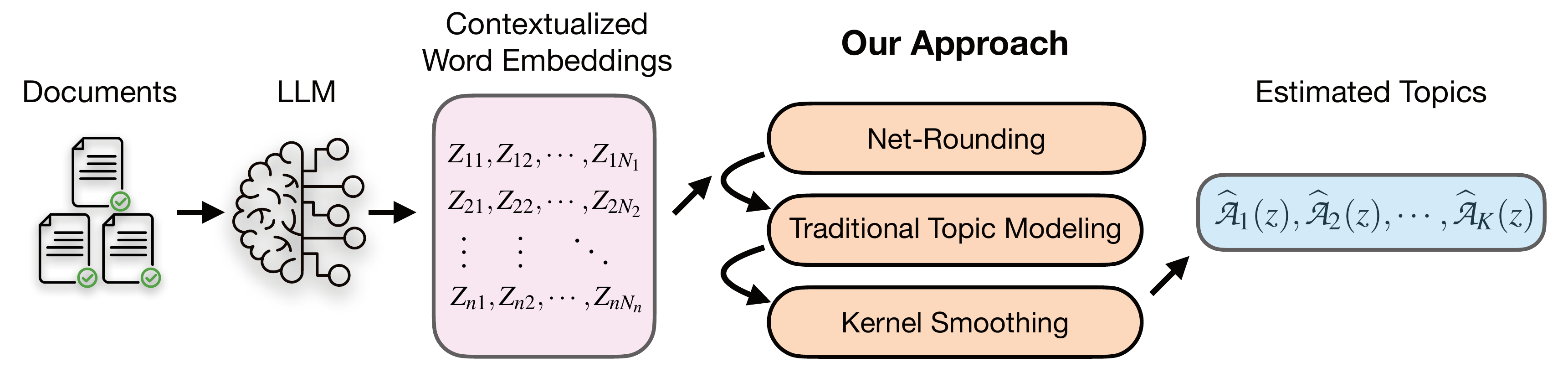}
\vspace{-.5cm}
\caption{The proposed topic modeling approach.} \label{fig:CTM}
\end{figure}

We propose a flexible approach to address 
parameter estimation 
in the PPTM framework, inspired by a key observation: Given an integer \( M \geq 2 \), for any partition \( \mathcal{Z} = \mathcal{R}_1 \sqcup \mathcal{R}_2 \sqcup \cdots \sqcup \mathcal{R}_M \)
where each \( \mathcal{R}_j \) is referred to as a ``hyperword'', we define \( X^{\text{net}} \in \mathbb{R}^{M \times n} \) as the hyperword count matrix, with \( X_i^{\text{net}}(j) \) representing the total number of embeddings from document \( i \) that fall into \( \mathcal{R}_j \). We show that \( X^{\text{net}} \) follows the traditional topic model \eqref{traditional-TM} with an effective vocabulary size of \( M \). Consequently, any traditional topic modeling approach can be applied to \( X^{\text{net}} \) to estimate \( K \) topic vectors \( \widehat{A}_1^{\text{net}}, \widehat{A}_2^{\text{net}}, \ldots, \widehat{A}_K^{\text{net}} \in \mathbb{R}^M \). This insight motivates a three-step method:
\vspace*{-5pt}
\begin{enumerate}
    \setlength{\itemsep}{-7pt}
  \item Partition \( \mathcal{Z} \) into a net \( \{\mathcal{R}_j\}_{j=1}^M \), with a properly chosen \( M \), and construct \( X^{\text{net}} \).
  \item Apply a traditional topic modeling approach to \( X^{\text{net}} \) to obtain \( \widehat{A}_1^{\text{net}}, \widehat{A}_2^{\text{net}}, \ldots, \widehat{A}_K^{\text{net}} \).
  \item Use kernel smoothing to construct \( \widehat{\mathcal{A}}_k(\cdot) \) from the vector \( \widehat{A}_k^{\text{net}} \) for each \( 1 \leq k \leq K \).
\end{enumerate}
\vspace*{-5pt}
We call this approach \emph{TRACE (Topic Representation Assisted by Contextualized Embeddings)} (see \cref{fig:CTM}). TRACE is a flexible algorithm that allows users to seamlessly integrate any traditional topic modeling method without further modification.

To facilitate theoretical analysis, we employ Topic-SCORE \citep{ke2022using} as the plug-in traditional topic modeling algorithm. Topic-SCORE has a theoretical guarantee for estimating traditional topic models \citep{ke2022using,ke2024entry}, hence permitting us to derive explicit error rates for our TRACE algorithm. 
Assuming \( \beta \)-Hölder smoothness for \( \mathcal{A}_k(\cdot) \), we establish the rate as a function of the number of documents \( n \), the average document length \( N \), the embedding dimensionality \( d \), and the smoothness parameter \( \beta \). We also provide a minimax lower bound and show that the error rate of our method matches this lower bound, up to a logarithmic factor, when \( \beta \leq 1 \). 
Our analysis is technically challenging. The difficulty in deriving the minimax upper bound stems from the fact that our method applies kernel smoothing to the output matrix of Topic-SCORE, whose entries are {\it not} independent samples. The challenge in establishing the minimax lower bound lies in constructing least-favorable configurations for \( K \) ``separable" base densities, rather than a single density.

We apply our method to two datasets: a news corpus from the Associated Press and a paper abstract corpus called MADStat. The results illustrate the advantages of our approach. Compared with traditional word-count-based topic models, our method can uncover ``weak" topics by leveraging the richer information encoded in word embeddings. Compared with recent neural topic models (NTMs), our approach provides context-dependent topic relevance for each word. For example, it estimates that the word {\it bond} is relevant to the {\it Finance} topic when appearing in {\it ``government bond"} and to the {\it Entertainment} topic when appearing in {\it ``James Bond."} In contrast, NTMs produce only a single topic relevance vector for each word, ignoring contextual variation.

\vspace*{-5pt}
\subsection{Comparison with Existing Studies in Literature} \label{subsec:literature}

\vspace*{-5pt}

Previous methods integrating word or document embeddings into topic modeling have mainly focused on the neural topic modeling (NTM) framework \citep{wu2024survey, wu2022mitigating, pham2024neuromax, nguyen2024glocom}. The basic NTM still adopts the word-count-based topic model \eqref{traditional-TM}, while neural networks are used in the inference procedure, typically via a variational autoencoder \citep{wu2024survey}.
Building on the basic NTM, non-contextualized word embeddings have been incorporated to reparameterize the topics.
Specifically, let $z_1,z_2,\ldots,z_p\in\mathbb{R}^d$ be the non-contextualized word embeddings for the $p$ unique words in the vocabulary. 
Each topic is represented by a vector $t_k\in\mathbb{R}^d$, and each $A_k$ in \eqref{traditional-TM} is determined by the relationships between $t_k$ (topic embedding) and $z_1,z_2,\ldots,z_p$ (word embeddings). 
A widely used model in this category is the Embedded Topic Model (ETM) \citep{dieng2020topic}, where $A_k= \mathrm{Softmax}(z_1't_k, z_2't_k,\ldots,z_p't_k)$, for $1\leq k\leq K$.  
Another popular idea is to define $A_k$'s by using an optimal transport plan between topic embeddings and word embeddings \citep{zhao2021neural}. 
A major limitation of these approaches is that they are still using word count data to train the model, without  incorporating contextual information. 


In the NTM literature, the predominant approach to incorporating contextual information is to use document embeddings from a pre-trained LLM. 
BERTopic \citep{grootendorst2022bertopic} has been popular due to its computational efficiency and accessible software implementation. It clusters document embeddings and treats each cluster as a ``topic," with representative words selected based on their frequencies within each document cluster. However, this approach yields only topic-relevant word lists rather than explicit topic embeddings. 
Alternatively, Contextualized Topic Model (CTM) \citep{bianchi2021pre} integrates   document embeddings into the inference procedure of the basic NTM, allowing the posterior distribution of document-level topic weights to depend on document embeddings.

More recently, several approaches embed documents, words, and topics into a shared latent space \citep{zhao2021neural, wang2022representing, wu2024fastopic}, where the document embeddings are given by a pre-trained LLM, and the topic and word embeddings are trained using text data. In these works, the model \eqref{traditional-TM} is retained, but the topic vectors $A_k$ are jointly determined by word and topic embeddings, and the weight vectors $w_i$ are jointly determined by topic and document embeddings. FASTopic \citep{wu2024fastopic} is a representative example, in which an optimal transport formulation is used to parameterize both $A_k$ and $w_i$.

However, the above approaches still do not allow for context-aware topic definitions. Regardless of how topics are parameterized by the topic embeddings, each word in the vocabulary is associated with a single, context-independent {\it topic relevance vector}. For example, the topic relevance vector for {\it bond} is a $K$-dimensional nonnegative vector whose entries sum to one. Although it may assign nonzero weights to both the {\it Finance} and {\it Entertainment} topics, it does not indicate the contexts in which {\it bond} is relevant to each topic. In contrast, our proposed model allows the topic relevance vector of a word to vary with its context, thereby enabling context-dependent topic definitions.

To the best of our knowledge, TopClus \citep{meng2022topic} is the only existing method that integrates contextualized word embeddings to enable context-aware topic definitions. 
Both this framework and ours model a topic as a continuous distribution over the word embedding space. However, their approach parameterizes these distributions using neural networks, whereas we adopt a fully nonparametric formulation. Another key distinction is that, in their model, document-level topic weights affect only the distributions of document embeddings but not those of word embeddings; equivalently, all word embeddings in the corpus are assumed to be identically distributed. In contrast, we allow the distributions of word embeddings to depend directly on the document-level topic weights.


Finally, we highlight two other advantages of our work over the existing NTM literature. First, our work provides a fully probabilistic framework with a solid theoretical foundation, whereas NTM approaches lack theoretical guarantees. Second, most existing methods rely on specialized estimation procedures, such as designing new neural network architectures or adapting EM algorithms with variational inference. In contrast, our method seamlessly integrates any traditional topic modeling algorithm as a plug-in module, without the need for modifications.

{\bf Paper Outline:} 
The remainder of this article is organized as follows: In \cref{sec:Model}, we explain the background and rationale of our model. In \cref{sec:Method}, we present our proposed method. \Cref{sec:Theory} establishes the theoretical properties. 
 \Cref{sec:Simu} contains simulation studies, and 
\cref{sec:RealData} analyzes two real datasets. Finally, \cref{sec:Discuss} concludes the paper with a discussion. Proofs and additional numerical results are provided in the supplemental material.

\vspace{-2em}
\section{PPTM: Background, Motivation, and Interpretation} \label{sec:Model} 
\vspace*{-5pt}

A key component of our framework is the assumption that the contextualized word embeddings---obtained by processing an input document through a pre-trained LLM---follow a Poisson process as in \eqref{PTM-2}. In this section, we review relevant background on word-embedding approaches and justify this assumption  by highlighting a structural property inherent to the architectures of modern LLMs. 

Word embeddings are vector representations of words that capture semantic relationships. Classical embedding methods \citep{mikolov2013distributed, pennington2014glove}, train neural networks to assign a single, fixed vector to each word in the vocabulary. However, these embeddings ignore the contextual variability of word meaning. 
In contrast, modern LLMs produce context-dependent word embeddings, enabled by the transformer architecture \citep{vaswani2017attention}. 
Unlike recurrent neural networks (RNNs), transformers process all words in parallel rather than sequentially. As we show below, this architectural design weakens order-induced dependencies and produces embeddings that are naturally suited to point-process modeling.

Consider an input document represented by a sequence $v_1,v_2,\ldots,v_N$, where each $v_j$ takes values in a vocabulary, and the sequence length $N$ can vary from occurrence to occurrence. An (encoder) transformer ${\cal T}$ maps $v_1,v_2,\ldots,v_N$ to another sequence $z_1, z_2, \ldots, z_N\in\mathbb{R}^d$, where $d$ is the dimensionality of the output embedding space. We denote it by $z_{1:N}={\cal T}(v_{1:N})$ for brevity. 
By the design of a transformer, ${\cal T}$ is a composition of multiple sequence-to-sequence mappings, each called a {\it layer}. 
The first two are an {\it Input Embedding} layer, ${\cal T}^{\text{inp}}$, and a {\it Positional Encoding} layer, ${\cal T}^{\text{pos}}$, and all remaining layers are {\it Self Attention} layers, ${\cal T}^{\text{att}}_1, {\cal T}^{\text{att}}_2, \cdots, {\cal T}^{\text{att}}_L$. 
\vspace{-.3em}
\beq \label{transformer}
{\cal T} = {\cal F} \circ {\cal E}, \qquad\mbox{where}\quad {\cal E} = {\cal T}^{\text{pos}}\circ {\cal T}^{\text{inp}}, \qquad {\cal F} = {\cal T}^{\text{att}}_L \circ {\cal T}^{\text{att}}_{L-1}\circ\cdots\circ {\cal T}^{\text{att}}_1. 
\vspace{-.2em}
\eeq
We also write $x_{1:N}={\cal E}(v_{1:N})$ and $z_{1:N}={\cal F}(x_{1:N})$. 
The two sequence-to-sequence mappings, ${\cal F}$ and ${\cal E}$, have a key difference: ${\cal E}$ is applied {\it position-wise}, meaning that $x_j$ only depends on $v_j$ but not any other $v_k$ with $k\neq j$, while ${\cal F}$ is {\it cross-positional}, where $z_j$ depends on all of $x_1, \ldots, x_N$. For example, in a toy version of ${\cal F}$, $z_j =\sum_{k=1}^N a(x_j, x_k) \zeta(x_k)$, where  $\zeta: \mathbb{R}^{d_0}\to\mathbb{R}^{d}$ is a mapping that is applied position-wise ($d_0$ is the dimensionality of $x_j$'s), and $a: \mathbb{R}^{d_0}\times \mathbb{R}^{d_0}\to \mathbb{R}$ is a bi-variate function that determines the ``attention weight" of the $k$th position on the $j$th position. 
This cross-positional mapping scheme is what enables the embeddings to encode contextual information.

There are many possible cross-positional mapping schemes. For instance, we may consider auto-regressive mappings where $z_j$ depends on $x_{1:j}$. One distinct feature of the self-attention mapping is its {\it permutation invariance}. In the above toy example, we clearly observe that permuting the input vectors $x_1, \ldots, x_N$ only results in a permutation of $z_1, \ldots, z_N$, with no other changes. In contrast, this property does not hold for autoregressive mappings, where re-ordering the inputs produces a different set of vectors. We formally state this property in the following lemma:
\begin{lemma} \label{lem:transformer}
Let ${\cal T}$ be a transformer as in \eqref{transformer}, where each ${\cal T}^{\text{att}}_\ell$ consists of a Multi-Head Attention sub-layer and a Feed Forward sub-layer, with two sub-layers connected by a Residual Connection. 
Then, ${\cal F}={\cal T}^{\text{att}}_L\circ {\cal T}^{\text{att}}_{L-1}\circ\cdots\circ {\cal T}^{\text{att}}_1$ satisfies the permutation invariance property: for any input sequence $x_{1:N}$ and any order permutation operation ${\cal P}_N[\cdot]$, it holds that ${\cal F}\bigl( {\cal P}_N[x_{1:N}]\bigr) = {\cal P}_N\bigl[ {\cal F}(x_{1:N})\bigr]$. 
\end{lemma}

As a consequence, the mapping ${\cal F}$ can be viewed as transforming a ``bag of vectors" $\{x_1, \ldots, x_N\}$ into another ``bag of vectors" $\{z_1, \ldots, z_N\}$, disregarding input order. Recall that ${\cal T}={\cal F}\circ {\cal E}$, where ${\cal E}$ contains only two layers and ${\cal F}$ contains all the remaining layers. 
In other words, the input order of words are not useful in most layers. 
This motivates us to model the output as a ``bag of vectors" rather than an ordered sequence.
Modeling a bag of vectors is equivalent to modeling its {\it empirical measure} $\hat{\lambda}(\cdot)$, where, for any measurable set $B \subset \mathbb{R}^d$, $\hat{\lambda}(B)$ is the number of vectors falling into $B$. In statistics, an empirical measure is modeled using a point process, with the Poisson process being the most commonly used. This explains our choice of the model in \eqref{PTM-2}.

{\bf Remark 1}: It is important to note that our framework doesn't model a natural language itself. Instead, it models the learnt representations of a natural language. In the modern machine learning practice, it is believed that the pre-training (i.e., representation learning) step can  find desirable representations of natural language so that they can be treated as exchangeable data in downstream tasks \citep{zhang2022analysis, chen2025exchangeability, chen2023learning}. We hold a similar perspective here. 


Next, we state an equivalent data generating process for the model in \eqref{PTM-2}, which follows well-established theory of Poisson processes \citep{last2017lectures}: 
\vspace{-.5em}
\begin{equation}\label{PTM-equivalence}
\textstyle N_i\sim \mathrm{Poisson}(N), \qquad z_{i1},z_{i2},\ldots,z_{iN_i}\bigm|N_i\;\mathrel{\overset{\scalebox{0.6}{ i.i.d.}}{\scalebox{1.2}[1]{$\sim$}}}\; \Omega_i(\cdot).
\end{equation}
This expression not only provides an alternative characterization of the PPTM but also suggests several natural extensions of the model. For instance, the Poisson distribution assumed for document lengths can be generalized to a mixture of Poisson distributions to capture potential heterogeneity and multimodality in document lengths (e.g., a news corpus may contain both short news flashes and long articles). We further note that the i.i.d. assumption on the $z_{ij}$'s arises from modeling the embeddings with a Poisson process rather than a more general point process. A natural extension is to consider a Cox process, which generalizes the Poisson process by allowing the intensity measure $\Omega_i(\cdot)$ itself to be stochastic, thereby inducing dependence among the $z_{ij}$'s. 


Finally, we discuss how to determine the topic relevance of a word under our framework. 
\begin{definition} \label{def:topic-relevance}
We call ${\cal B}_k(z) = {\cal A}_k(z)/[\sum_{\ell=1}^K {\cal A}_\ell(z)]$ the normalized topic densities. 
For any $z\in\mathbb{R}^d$, its topic relevance vector is defined as $\boldsymbol{\mathcal{B}}(z)=\bigl(\mathcal{B}_1(z), \mathcal{B}_2(z), \ldots, \mathcal{B}_K(z)\bigr)' $.  
\end{definition}

Under this definition, a large value of ${\cal B}_k(z)$ indicates that the embedding $z$ has a higher density under the $k$th topic than under the other topics. The $K$ entries of $\boldsymbol{\mathcal{B}}(z)$ sum to $1$. When ${\cal B}_k(z)=1$, it means that $z$ appears exclusively in the $k$th topic; we refer to such $z$ an {\it anchor embedding} of the $k$th topic. 
Because our notion of topic relevance is defined with respect to embedding values rather than words, it is inherently context-dependent. In Section~\ref{sec:RealData} (particularly,  Figure~\ref{fig:AP-context} and Table~\ref{tb:MADStat-context}), we demonstrate that this context-aware definition leads to improved performance and interpretability. 

{\bf Remark 2}: In practice, it is often desirable to produce a ``representative word list" for each topic to facilitate interpretation and labeling. This is feasible within our framework because the original text allows us to trace each observed embedding back to its corresponding word. Specifically, we identify the observed embeddings with largest values of ${\cal B}_k(z)$ and map them back to their associated words to form the representative list. See Section~\ref{sec:RealData} for details.

\vspace{-2em}
\section{Integrating Word Embeddings in Topic Modeling} \label{sec:Method}

\vspace*{-10pt}

We introduce our algorithm for estimating parameters of PPTM. An outline of our algorithm for estimating the topic desities has been given in Section~\ref{subsec:outline}, and we now provide details in   \cref{subsec:TM,subsec:KS}. 
In \cref{subsec:estW}, we describe how to estimate the topic weight vector for each document.

\vspace{-5pt}

\subsection{Topic Modeling on Hyperword Counts} \label{subsec:TM}

Recall that we define a partition ``net'' of the embedding space \( {\cal Z} = \bigsqcup_{m=1}^M \calR_m \), where \( \calR_1, \ldots, \calR_M \) are disjoint regions, termed ``hyperwords''. The count of a hyperword \( \calR_m \) in document \( i \) is defined as the number of embeddings \( z_{i1}, z_{i2}, \ldots, z_{iN_i} \) that fall into \( \calR_m \). This forms a hyperword count matrix \( X^{\text{net}} = [X_1^{\text{net}}, X_2^{\text{net}}, \ldots, X_n^{\text{net}}] \in \mathbb{R}^{M \times n} \), where:
\vspace{-.5em}
\begin{equation} \label{def:Xnet}
X_i^{\text{net}}(m) = \bigl|\{z_{i1}, z_{i2}, \ldots, z_{iN_i}\} \cap {\cal R}_m\bigr|, \qquad \text{for } 1 \leq m \leq M.
\vspace{-.5em}
\end{equation}
Using the properties of a Poisson point process \citep{last2017lectures}, 
we can prove the following lemma:
\vspace*{-5pt}
\begin{lemma} \label{lem:induced-model}
Under 
the model given by \eqref{PTM-1}-\eqref{PTM-2}, 
the entries $X_i^{\text{net}}$ are independent and satisfy $X_{i}^{\text{net}}(m)\sim \mathrm{Poisson}(\lambda_{im})$, where $\lambda_{im}=N\int_{i\in \calR_m}\Omega_i(z)dz$. Conditioning on $N_i$, we have
\vspace{-.5em}
\[
\textstyle X_i^{\text{net}}\sim \mathrm{Multinomial}\bigl(N_i,\, \Omega_i^{\text{net}}\bigr), \qquad\mbox{with}\quad\Omega_i^{\text{net}}=\sum_{k=1}^K w_i(k)A_k^{\text{net}},  
\vspace{-.5em}
\]
where $A_k^{\text{net}}\in\mathbb{R}^M$ is such that $A_{k}^{\text{net}}(m)=\int_{z\in\calR_m}A_k(z)dz$ for all $1\leq m\leq M$. 
\end{lemma} 

Comparing \cref{lem:induced-model} with \eqref{traditional-TM}, we observe that $X^{\text{net}}$ follows a traditional topic model, characterized by an effective vocabulary size $M$. Consequently, traditional topic modeling algorithms can be applied to $X^{\text{net}}$ to estimate the topic matrix $\widehat{A}^{\text{net}} = [\widehat{A}_1^{\text{net}}, \widehat{A}_2^{\text{net}}, \ldots, \widehat{A}_K^{\text{net}}] \in \mathbb{R}^{M \times K}$. Each column $\widehat{A}_k^{\text{net}}$ serves as an approximation of $A_k^{\text{net}}$, as defined in \cref{lem:induced-model}.

To implement this idea, we can conceptually consider ${\cal Z} = [0,1]^d$ as being partitioned into a hypercubic grid. For a small $\epsilon \in (0,1)$, ${\cal Z}$ is divided into equal-sized hypercubes with side length $\epsilon$. Since the total number of hypercubes is proportional to $1/\epsilon^d$, theoretical analysis can be performed by setting $\epsilon = O(M^{-1/d})$, where $M$ is the number of hypercubes, or the effective vocabulary size.

In practice, however, hypercubic grids are not efficient for grouping semantically related word embeddings. This inefficiency arises not only from the curse of dimensionality but also from the fact that word embeddings are not uniformly distributed on $\mathcal{Z}\subset\mathbb{R}^d$, reflecting the heterogeneity in word frequencies. Consequently, some hypercubes may remain empty while others may contain too many embeddings. To address this limitation, we use k-means clustering to group embeddings, which naturally induces a more appropriate partition. Specifically, let $\widehat{x}_1, \widehat{x}_2, \ldots, \widehat{x}_M$ denote the cluster centers. For $1 \leq m \leq M$, we define $
\textstyle \mathcal{R}_{m} = \bigl\{ z \in \mathbb{R}^{d} \bigm\vert \lVert z - \widehat{x}_m \rVert_{2} < \min_{j\neq m}\lVert z - \widehat{x}_{j} \rVert_{2} \bigr\}$. 
This formulation corresponds to a Voronoi diagram, 
which aligns with the cluster assignment mechanism in k-means. Intuitively, this data-driven approach generates smaller $\mathcal{R}_m$ in regions of higher embedding density, effectively capturing the intended grouping.

Next, we apply traditional topic modeling to the hyperword count matrix \( X^{\text{net}} \). Existing methods can be grouped into two main categories: Bayesian approaches and anchor-word-based approaches. Bayesian approaches, including the famous LDA algorithm \citep{blei2003latent} and its extensions, 
typically assume a Dirichlet prior on \( w_1, w_2, \ldots, w_n \) and estimate the topic vectors by maximizing marginal likelihoods using the EM algorithm or variational inference. In contrast, anchor-word-based approaches \citep{arora2013practical,bing2020fast,ke2022using} assume the existence of ``anchor words''---words uniquely associated with specific topics---and 
leverage the structure of the word count matrix under this assumption.
Importantly, our framework is flexible enough to accommodate any of these  topic modeling approaches. 

Throughout this paper, we mainly use Topic-SCORE \citep{ke2022using} as the plug-in topic modeling algorithm, for its computational efficiency and rigorous theoretical guarantees. For ease of presentation, we still use the notations in \eqref{traditional-TM}. Under this model, the $p\times n$ data matrix $X=[X_1, X_2, \ldots, X_n]$ is a low-rank matrix plus noise, which motivates us to apply singular value decomposition (SVD). 
The key discovery in \cite{ke2022using} is that the matrix formed by the first $K$ left singular vectors of $X$ has an explicit connection to the target quantity---the topic vectors---through a low-dimensional simplex geometry. This geometry is a consequence of both the inherent structure of the topic model and the use of the SCORE normalization \citep{SCOREreview} on singular vectors. Inspired by this discovery,  \cite{ke2022using} proposed a fast spectral algorithm, which reduces the topic estimation into a low-dimensional simplex vertex hunting problem. Further details on this algorithm are provided in the supplemental material.



\vspace*{-5pt}
\subsection{Estimation of Topic Densities} \label{subsec:KS}
\vspace*{-5pt}

By applying traditional topic modeling algorithms on \( X^{\text{net}} \), we obtain the topic vector \(\widehat{A}_k^{\text{net}}\) for each topic \(1 \leq k \leq K\) at the hyperword level. These topic vectors can be leveraged for density estimation of the corresponding topic measures \(\mathcal{A}_k(\cdot)\) for \(1 \leq k \leq K\).

To achieve this, we adopt the kernel smoothing approach commonly used in kernel density estimation.  
Let \(\mathcal{K}:\mathbb{R}^d \to \mathbb{R}\) be a kernel function satisfying \(\int_{\mathbb{R}^d} \mathcal{K}(u) \, du = 1\), and let \(h > 0\) be a bandwidth parameter. We define the rescaled kernel as \(\mathcal{K}_h(u) = \mathcal{K}(u/h)/h^d\). By a change of variable in the integral, it follows that \(\int_{\mathbb{R}^d} \mathcal{K}_h(u) \, du = 1\). Using this rescaled kernel, we propose the following estimator for \(\mathcal{A}_k(\cdot)\):
\begin{equation} \label{def:hatA}
\widehat{\mathcal{A}}_k(z_0) = \sum_{m=1}^M \zeta_m(z_0) \widehat{A}_k^{\text{net}}(m), \qquad \text{where} \quad \zeta_m(z_0) = \frac{1}{\mathrm{Vol}(\mathcal{R}_m)} \int_{\mathcal{R}_m} \mathcal{K}_h(z-z_{0}) \, dz.
\end{equation}
Here, \(\mathcal{K}(\cdot)\) and \(\widehat{\mathcal{A}}_k(\cdot)\) are defined on \(\mathbb{R}^d\) rather than the subset \(\mathcal{Z}\), for technical convenience.

To justify the choice of \(\zeta_{m}(\cdot)\), we consider a naive density estimator \(
  \widehat{\mathcal{A}}^{\text{naive}}_{k}(z_{0}) = \sum_{m=1}^{M} \frac{\mathbb{I}_{z_{0} \in \mathcal{R}_{m}} \widehat{A}_{k}^{\text{net}}(m)}{\mathrm{Vol}(\mathcal{R}_{m})}
\),  which uniformly assigns the probability mass \(\widehat{A}_{k}^{\text{net}}(m)\) over \(\mathcal{R}_{m}\). It is straightforward to verify that:
\begin{equation}\label{eq:kernelsmoothing}
  \widehat{\mathcal{A}}_{k}(z_{0}) = \sum_{m=1}^{M} \frac{\widehat{A}_{k}^{\text{net}}(m)}{\mathrm{Vol}(\mathcal{R}_{m})} \int_{z \in \mathcal{R}_{m}} \mathcal{K}_{h}(z-z_{0}) dz = \int_{z \in \mathbb{R}^{d}} \mathcal{K}_{h}(z-z_{0}) \widehat{\mathcal{A}}^{\text{naive}}_{k}(z) dz,
\end{equation}
which means that \(\widehat{\mathcal{A}}_{k}(\cdot)\) is the convolution (more precisely, cross-correlation) of \(\widehat{\mathcal{A}}^{\text{naive}}_{k}(\cdot)\) with the rescaled kernel function \(\mathcal{K}_{h}(\cdot)\). This motivates the term \emph{kernel smoothing} for this 
step (see \cref{fig:CTM}).

Equation \eqref{eq:kernelsmoothing} immediately implies that \(\int_{z_{0} \in \mathbb{R}^{d}} \widehat{\mathcal{A}}_{k}(z_{0}) dz_{0} = 1\). If the kernel \(\mathcal{K}\) is non-negative, then \(\widehat{\mathcal{A}}_{k}(\cdot)\) corresponds to a proper probability density on \(\mathbb{R}^{d}\). Although \(\widehat{\mathcal{A}}_{k}(z_{0})\) does not typically vanish for \(z_{0} \in \mathbb{R}^{d} \setminus \mathcal{Z}\), it remains close to zero and negligible for \(z_{0}\) such that \(\inf_{z \in \mathcal{Z}} \lVert z-z_{0} \rVert_{2}\geq \epsilon > 0\), provided the bandwidth \(h\) is sufficiently large.

Following the convention in the nonparametric literature, we further consider the use of \emph{higher-order kernels} to reduce bias in density estimation, particularly for our theoretical analysis.  
For any $\beta>0$, let $\lfloor\beta\rfloor$ denote the largest integer that is $<\beta$. When $d=1$, an order-$\beta$ kernel satisfies that
\resetspacing
\begin{equation}\label{eq:momentconditionhigh}
  \int \mathcal{K}(u) \, du = 1, \quad \int u^k \mathcal{K}(u) \, du = 0 \quad \forall 1\leq k\leq \lfloor \beta \rfloor, \quad  \int \lvert u \rvert^{\beta}\lvert\mathcal{K}(u)\rvert du<\infty.
\end{equation}
When $d\geq 2$, we extend the definition in \eqref{eq:momentconditionhigh} by taking a product of 1-dimensional kernels. For instance, the Gaussian kernel is an order-\(1\) kernel.  
For \(\beta \geq 2\), however, an order-\(\beta\) kernel must take negative values on a subset of \(\mathbb{R}\) with positive Lebesgue measure. As a result, the estimators \(\widehat{\mathcal{A}}_k(\cdot)\) 
can also take negative values. Nevertheless, as noted in Chapter 1.2.1 of \cite{tsybakov2009introduction}, this issue is of limited practical concern because the positive part of the estimator, 
\(\widehat{\mathcal{A}}_k^+(\cdot) = \max \{\widehat{\mathcal{A}}_k(\cdot), 0\}\), can be renormalized to yield a non-negative probability measure. This adjustment ensures the validity of the density estimator without significant loss of theoretical rigor.

By definition \eqref{def:hatA}, the calculation of \(\zeta_{m}(z_{0})\) involves integration over the kernel function \(\mathcal{K}_{h}(\cdot)\). For a general kernel function and partition net, obtaining the exact integral is not always feasible. To address this, we provide two numerical approximations for \(\zeta_{m}(z_{0})\) that are more practical:
\begin{equation} \label{def:weight-proxy}
\widehat\zeta^{(1)}_m(z_0) = \frac{\sum_{z_{ij} \in \mathcal{R}_m} \mathcal{K}_h(z_{ij}-z_{0})}{\#\{z_{ij} \mid z_{ij} \in \mathcal{R}_m\}}, \qquad \text{or} \qquad \widehat\zeta^{(2)}_m(z_0) = \mathcal{K}_h(x_m-z_{0}),
\end{equation}
where \(z_{ij}\) is defined in \eqref{def:Xnet}, and \(x_m\) is a single point chosen from \(\mathcal{R}_m\). Typically, \(x_m\) is set as \(\widehat{x}_m\), the k-means cluster center, if the partition net is based on the k-means-induced Voronoi diagram 
in \cref{subsec:TM}. For computational and memory efficiency, we adopt \(\widehat\zeta^{(2)}_m(z_0)\) in our experimental study.


%


\spacingset{1}
\begin{algorithm}[tb!]
\caption{The TRACE algorithm for estimating the topic densities in PPTM} \label{alg:CTM}
\medskip
{\bf Input}: \(n\) documents, a pre-trained LLM, a traditional topic modeling algorithm, \(K\) (number of topics), \(d\) (embedding dimension), a kernel \(\mathcal{K}(\cdot)\), parameters \(M\) and \(h\), and any \(z_0 \in \mathbb{R}^d\).

\bigskip
{\bf Word Embedding}: Process the input documents using the pre-trained LLM to generate raw word embeddings \(\{z^{\text{raw}}_{ij}\}_{1 \leq i \leq n, 1 \leq j \leq N_i}\). 
Fit a UMAP model on a held-out sample of embeddings to obtain $f^{\text{UMAP}}:\mathbb{R}^{d_{\text{raw}}}\to \mathbb{R}^d$. Let $z_{ij}=f^{\text{UMAP}}(z_{ij}^{\text{raw}})$, for $1\leq i\leq n$, $1\leq j\leq N_i$. 

\bigskip
{\bf Topic Modeling}:
\vspace{-5pt}
\begin{enumerate}
\item {\it Net-rounding}: Perform k-means clustering on \(\{z_{ij}\}_{1 \leq i \leq n, 1 \leq j \leq N_i}\). Denote the cluster centers as \(\widehat{x}_1, \ldots, \widehat{x}_M\), and define the hyperwords \(\calR_1,  \ldots, \calR_M\) as regions in the corresponding Voronoi diagram. Construct the hyperword count matrix \(X^{\text{net}}\) as in \eqref{def:Xnet}.
\item {\it Traditional Topic Modeling}: Apply a traditional topic modeling algorithm (e.g., Topic-SCORE) to \(X^{\text{net}}\). Let \(\widehat{A}_1^{\text{net}}, \ldots, \widehat{A}_K^{\text{net}}\) be the estimated topic vectors.
\item {\it Kernel Smoothing}: Compute \(\widehat\zeta_1(z_0), \ldots, \widehat\zeta_M(z_0)\) using \eqref{def:weight-proxy}. For each \(1 \leq k \leq K\), calculate \(\widehat{\cal A}_k(z_0)\) as defined in \eqref{def:hatA} using \(\widehat{A}_k^{\text{net}}\), weights \(\widehat\zeta_m(z_0)\), and bandwidth \(h\).
\end{enumerate}

{\bf Output}: \(\widehat{\cal A}_1(z_0),  \ldots, \widehat{\cal A}_K(z_0)\) for any \(z_0 \in \mathbb{R}^d\).

\end{algorithm}

\resetspacing


Our complete algorithm, TRACE, is presented in \cref{alg:CTM}. Notably, we include a dimension reduction step when obtaining the word embeddings to better control the dimension \(d\). The output embeddings of large language models (LLMs) are often high-dimensional (e.g., \(768\) for BERT and \(4096\) for LLaMA). To improve efficiency and performance in downstream tasks, reducing the dimensionality of these embeddings is commonly recommended \citep{raunak2017simple}. 
We suggest choosing \(d\) to be roughly comparable to the anticipated number of topics. To preserve both the local and global structure of the embedding point cloud, we employ UMAP \citep{mcinnes2018umap} as the default dimension reduction algorithm. For cases with a large number of embeddings and limited computational resources, we further recommend fitting the UMAP model using a random subsample of the embeddings and applying the learned projection to all points. Similarly, in the net-rounding step, the k-means algorithm can be replaced with mini-batch k-means to enhance computational efficiency.

{\bf Remark 3}: There is a theoretical justification of applying dimension reduction. Let $f: \mathbb{R}^d\to\mathbb{R}^{q}$ be a fixed mapping, with $q\leq d$. If $x_{ij}$'s follow the model in \eqref{PTM-1}-\eqref{PTM-2}, then $f(x_{ij})$'s satisfy a similar model, with the same weight vectors $w_1, w_2, \ldots, w_n$ and new topic measures $\widetilde{\cal A}_1, \ldots, \widetilde{\cal A}_K$, where each $\widetilde{\cal A}_k$ is the push-forward measure of ${\cal A}_k$ by $f$. This ensures that we can continue to assume the PPTM on projected data. The argument here requires that the mapping $f$ is independent of data, which can be satisfied if we hold out a small fraction of embeddings to train the UMAP model.

{\bf Selection of $K$ and tuning parameters}: TRACE has three tuning parameters $(M, h, d)$. 
The net size parameter \(M\) introduces an important bias-variance trade-off, and its optimal theoretical value will be given in Section~\ref{sec:Theory}. In practice, we recommend setting \(M\) between \(0.05\%\) and \(0.2\%\) of the total number of embeddings. 
For the bandwidth $h$, we propose a maximum entropy criterion for a data-driven selection of \(h\), which details are in the supplemental material. 
The latent dimension $d$ is fixed at $d=10$ in our real-data analysis. We also studied different choices of $d$ and found that choosing $d$ comparable with $K$ typically yields the most satisfying results. 
Finally, although the number of topics $K$ is not a tuning parameter, it is often unknown in real data. We used a two-step approach for selecting $K$. First, define a range for \(K\) based on prior knowledge of the text corpus and the desired level of topic granularity. Second, after constructing the hyperword count matrix, examine the scree plot of its singular values to specify \(K\) within the pre-defined range. Additionally, we analyze the rules of various hyperparameters via simulation in Section~\ref{sec:Simu},  and provide 
detailed guiding principles for choosing these hyperparameters in Appendix~\ref{supp:AlgAP} and Appendix~\ref{supp:Algmadstat}.

\vspace*{-5pt}
\subsection{Estimation of Topic Weight Vectors for Each Document} \label{subsec:estW}
\vspace*{-5pt}

Since the hyperword count matrix \(X^{\text{net}} = [X_1^{\text{net}}, \ldots, X_n^{\text{net}}] \in \mathbb{R}^{M \times n}\) follows a traditional topic model, the weight vector \(w_i\) for each document \(i\) (\(1 \leq i \leq n\)) can be estimated using well-established techniques in literature. Having already applied Topic-SCORE \citep{ke2022using} to obtain the topic matrix \(\widehat{A}^{\text{net}} = [\widehat{A}_1^{\text{net}}, \ldots, \widehat{A}^{\text{net}}_K] \in \mathbb{R}^{M \times K}\), we further adopt their regression-based approach to estimate \(w_i\): 
\vspace{-.7em}
\begin{equation} \label{def:hat-w}
\widehat{w}_i = \operatorname{\arg\min}_{b \in \mathbb{R}^K} \bigl\lVert (\mathbf{1}_M' X_i^{\text{net}})^{-1} X_i^{\text{net}} - \widehat{A}^{\text{net}} b \bigr\rVert_2^2.
\vspace{-.7em}
\end{equation}
By assumption, the weight vectors \(w_i\) are non-negative and sum to one. However, due to randomness in the model assumptions and potential mis-specification when applied to real data, the estimators \(\widehat{w}_i\) are not guaranteed to satisfy these constraints. To address this, we mention two approaches to enforce valid weights. First, following \cite{ke2022using}, we can set all negative entries of the solution to zero and renormalize the vector to have unit \(\ell^1\)-norm. Alternatively, we can directly impose constraints during the optimization process in \eqref{def:hat-w}, ensuring that \(\widehat{w}_i\) satisfies the properties of a weight vector. In practice, 
we use the first approach as the default for computational efficiency.

Furthermore, since the number of documents \(n\) can be very large in real-world datasets, we are estimating a high-dimensional matrix \(W \). To address potential overfitting and improve numerical stability, we recommend adding a ridge penalty to \eqref{def:hat-w}: $\widehat{w}_i = \operatorname{\arg\min}_{b \in \mathbb{R}^K} \bigl\lVert (\mathbf{1}_M' X_i^{\text{net}})^{-1} X_i^{\text{net}} - \widehat{A}^{\text{net}} b \bigr\rVert_2^2 + \lambda \lVert b \rVert_2^2$, 
where the penalty parameter \(\lambda\) is typically set to a small value below \(0.1\).

Finally, 
for any incoming new document, we input it into the same LLM used in \cref{alg:CTM} to obtain raw embeddings. These embeddings are then projected to \(d\)-dimensional space using the UMAP transformation learned during the algorithm. With the previously-obtained Voronoi diagram fixed, we construct the hyperword count vector \(x_{\ast}^{\text{net}}\) for the new document, and we estimate its weight vector in a similar way:
  $\widehat{w}^{\ast} = \operatorname{\arg\min}_{b \in \mathbb{R}^K} \bigl\lVert (\mathbf{1}_M' x_{\ast}^{\text{net}})^{-1} x_{\ast}^{\text{net}} - \widehat{A}^{\text{net}} b \bigr\rVert_2^2 + \lambda \lVert b \rVert_2^2$.

\vspace*{-2em}

\section{Theoretical Properties}\label{sec:Theory}
\vspace*{-5pt}

We fix the number of topics, $K$, and the embedding space dimensionality, $d$, and assume ${\cal Z}=[0,1]^d$ 
without loss of generality. For any $d$-variate, $p$-times continuously differentiable function $f(\cdot)$ and a multi-index $\gamma=(\gamma_1,\ldots,\gamma_d)'$ such that each $\gamma_j\in \{0, 1,\ldots,p \}$ and $\lVert \gamma \rVert:=\gamma_{1}+\gamma_{2}+\cdots+\gamma_{d}=p$, we use $D^{\gamma}f(x)=\frac{\partial^p f(x)}{\partial x_{1}^{\gamma_{1}}\partial x_{2}^{\gamma_{2}}\cdots\partial x_{d}^{\gamma_{d}}}$ to denote the order-$p$ derivative of $f(\cdot)$ associated with $\gamma$. 
Assume that ${\cal A}_1(\cdot), \ldots, {\cal A}_K(\cdot)$ are fixed densities on ${\cal Z}$ satisfying the H\"{o}lder smoothness assumption:

\begin{assumption}[$\beta$-H\"{o}lder smoothness]\label{assump1}
For $p=\lfloor\beta\rfloor$, the order-$p$ derivative of ${\cal A}_k(\cdot)$ satisfies that  $
|D^{\gamma}{\cal A}_k(z)-D^{\gamma}{\cal A}_k(\tilde{z})|\leq c_1\|z-\tilde{z}\|^{\beta-p}$, for all $(z,\tilde{z})\in {\cal Z}^2$ and all multi-indices $\gamma=(\gamma_1,\gamma_2,\ldots,\gamma_d)'$ of order $p$, where $c_1>0$ is a constant.  
\end{assumption}

In the traditional topic model, the anchor word condition \citep{arora2013practical} is often imposed to ensure that the parameters are identifiable. We introduce an analogous condition for the PPTM framework.
For each $1\leq k\leq K$, 
define the anchor region of topic $k$ by
\vspace{-.5em}
\beq \label{def-anchor-region}
{\cal S}_k=\bigl\{z\in {\cal Z}: {\cal A}_k(z)\neq 0,\; \sum_{j: j\neq k}{\cal A}_j(z)=0\bigr\}. 
\vspace{-.5em}
\eeq
\begin{assumption}[Anchor region]\label{assump2}
The volumes of ${\cal S}_1, {\cal S}_2,\ldots, {\cal S}_K$ are lower bounded by 
$c_2>0$. 
\end{assumption}

We also define a function  $h(\cdot)$ and two $K\times K$ matrices $\Sigma_W$ and  $\Sigma_{\cal A}$ by
\beq \label{def-h-SigmaA}
h(z)=\sum_{j=1}^K {\cal A}_j(z), \quad \Sigma_W= n^{-1}WW', \quad \mbox{and}\quad \Sigma_{\cal A}(k,\ell)=\int_{z\in{\cal Z}}\frac{{\cal A}_k(z){\cal A}_\ell(z)}{h(z)}dz. 
\eeq
Here, $h(z)$ characterizes the variation of word frequency over the embedding space, $\Sigma_W$ is the topic-topic covariance matrix \citep{arora2013practical,ke2022using}, and $\Sigma_{\cal A}$ is analogous to the topic-topic overlapping matrix \citep{ke2022using} in the traditional topic model. 
By the definitions, $\Sigma_W$ and $\Sigma_{\cal A}$ are properly scaled such that their operator norms are upper bounded by constants. We further impose some mild regularity conditions: 

\begin{assumption}\label{assump3}
$\inf_{z\in {\cal Z}}\{h(z)\}\geq c_3$, $\lambda_{\min}(\Sigma_W)\geq c_4$, $\lambda_{\min}(\Sigma_{\cal A})\geq c_5$, and $\min_{1\leq k,\ell\leq K}\{\Sigma_{\cal A}(k,\ell)\}\geq c_5$, where $c_3$-$c_5$ are positive constants. 
\end{assumption}

We use a toy example to demonstrate that Assumptions~\ref{assump1}-\ref{assump3} can hold  simultaneously.
Define $\Psi(x)=\exp(\frac{1}{x^2-1})\cdot 1\{0\leq x< 1\}$. This function is supported on $[0,1)$ and has continuous derivatives of all orders in $(0,\infty)$. We fix $K=2$ and $d=1$, and let ${\cal A}_1(z)= a_0\cdot \Psi(3z/2)$, $
{\cal A}_2(z)= a_0\cdot \Psi(3(1-z)/2)$, and $ 
w_1,\ldots,w_n \overset{iid}{\sim} \mathrm{Dirichlet}({\bf 1}_2)$, 
where $a_0$ is a normalizing constant such that the integrals of  ${\cal A}_1(\cdot)$ and ${\cal A}_2(\cdot)$ are equal to $1$. In this example, both ${\cal A}_1(\cdot)$ and ${\cal A}_2(\cdot)$ have continuous derivatives of all orders in $[0,1]$, so Assumption~\ref{assump1} holds. The support of ${\cal A}_1(\cdot)$ is $[0,2/3)$, and the support of ${\cal A}_2$ is $(1/3,1]$. As a result, the anchor regions are ${\cal S}_1=[0,1/3)$ and ${\cal S}_2=(2/3,1]$, which verifies
Assumption~\ref{assump2}. 
By elementary properties of Dirichlet distributions, $\Sigma_W$ converges to a fixed matrix $\frac{1}{6}(I_2-\frac{1}{8}{\bf 1}_2{\bf 1}_2')$ as $n\to\infty$, so that $\lambda_{\min}(\Sigma_W)\to \frac{1}{8}$. 
Furthermore, the function $h(z)$ and the matrix $\Sigma_{\cal A}$ can be explicitly calculated. It suggests that Assumption~\ref{assump3} is satisfied.
This example can be extended to arbitrary $(K, d)$. In the supplementary material (especially Figure~\ref{fig:LB-example}), 
we show that there exist many parameter instances such that Assumptions~\ref{assump1}-\ref{assump3} hold.

\begin{lemma} \label{lem:identifiability}
    Under Assumptions~\ref{assump1}-\ref{assump3}, all the parameters in the PPTM framework are identifiable.
\end{lemma}

In the remainder of this section, first, in Section~\ref{subsec:UB}, we study the error rate 
in estimating the topic densities. Next, in Section~\ref{subsec:LB}, we derive an information-theoretic lower bound and discuss the optimality of our method. Finally, in Section~\ref{subsec:W-rate}, we study the error rate of estimating $w_i$'s.

\subsection{Error Rate for Estimating Topic Densities} \label{subsec:UB}

Our estimator uses a net on ${\cal Z}$. As discussed in Section~\ref{subsec:TM}, we use a special net in theoretical analysis, where each ${\cal R}_m$ is a hypercube with side length $\epsilon$. With this net, the  number of hyperwords is linked to $\epsilon$ through $
M \sim \epsilon^{-d}$. From now on, we use $\epsilon$, instead of $M$, as the net parameter.

The error in $\widehat{\cal A}_k(\cdot)$ decomposes into a ``bias'' term and a ``variance'' term.  
By \cref{lem:induced-model}, the traditional topic modeling on hyperword counts yields an estimate of $A_{k}^{\text{net}}(m)=\int_{z\in\calR_m}\mathcal{A}_k(z)dz$, for $1\leq m\leq M$. Combining this insight with \eqref{def:hatA}, we define a population quantity:
\begin{equation} \label{estimator-oracle}
\widetilde{\mathcal{A}}_k(z_{0})=\sum_{m=1}^M \zeta_m(z_{0}) A_k^{\text{net}}(m) = \sum_{m=1}^M \zeta_m(z_{0}) \int_{z\in\calR_m}\mathcal{A}_k(z)dz. 
\end{equation}
Here,  $\widetilde{\mathcal{A}}_k(z_{0})-\mathcal{A}_k(z_{0})$ is the ``bias'' term, and $\widehat{\mathcal{A}}_k(z_{0})-\widetilde{\mathcal{A}}_k(z_{0})$ is the  ``variance'' term.

First, we study the ``bias'' term. 
For $h\in (0,1/2)$, let ${\cal Z}_h=\{ z\in \mathcal{Z}: \lVert z-z_{0} \rVert\geq h,\,\forall\, z_{0}\in \mathbb{R}^{d}\setminus \mathcal{Z} \}$ be the 
set of points 
in the interior of ${\cal Z}$ and having
a distance at least $h$ from the boundary of ${\cal Z}$. 

\begin{lemma}[Bias]\label{lem:bias-new}
Suppose Assumptions~\ref{assump1}-\ref{assump3} are satisfied,  and the kernel function ${\cal K}(\cdot)$ in \eqref{estimator-oracle} is Lipschitz, has a compact support,
\spacingset{1}\footnote{Compactly-supported higher-order kernels can be constructed explicitly, as noted in Chapter 1.2.2 of \cite{tsybakov2009introduction}. However, this compact-support assumption is mainly for theoretical convenience and can be relaxed to permit kernels with exponential tails, such as the Gaussian kernel.}
\resetspacing 
and satisfies the requirement of order-$\beta$ kernel in \eqref{eq:momentconditionhigh}. There exist constant $C_0>0$ and $c_0\in (0,1)$ such that for all $h\in (0,c_0]$ and $1\leq k\leq K$, 
\[
\max_{z_0\in {\cal Z}_{h}}\lvert \widetilde{\mathcal{A}}_k(z_{0})-\mathcal{A}_k(z_{0})\rvert \leq C_0\bigl( h^\beta +\min\bigl\{ \epsilon^{ \beta\wedge 1},\,  h^{-(d+1)}\epsilon^{1+\beta\wedge 1}\bigr\}\bigr).
\] 
\end{lemma}

In classical kernel density estimation (KDE), the bias depends solely on the bandwidth \( h \). In contrast, in our framework, the bias is influenced by both $h$ and the side length $ \epsilon$ of the net. This distinction arises because, instead of applying kernel smoothing to data points sampled from a density, we apply it to the entries of \( \widehat{A}_k^{\text{net}} \). This fundamental difference results in distinct theoretical bounds and proof techniques.

Lemma~\ref{lem:bias-new} concerns the ``bias'' within the interior of ${\cal Z}$.  
For $z_0$ near the boundary, a ``boundary effect'' arises,  similar to that observed in classical KDE. Various boundary correction methods have been proposed (e.g., see \cite{jones1993simple}). These techniques could be incorporated in \eqref{def:hatA} to mitigate the boundary effect. However, it would complicate both the methodology and theoretical analysis. To maintain the conciseness of our presentation, we adhere to the current formulation and focus on the error within the interior of ${\cal Z}$. 

\begin{corollary} \label{cor:bias}
    In the setting of Lemma~\ref{lem:bias-new}, let $\delta_n=1/\log(n)$. With a proper $h$, for all $1\leq k\leq K$, $
\max_{z_0\in {\cal Z}_{\delta_n}}\lvert \widetilde{\mathcal{A}}_k(z_{0})-\mathcal{A}_k(z_{0})\rvert \leq C\epsilon^{\theta_d(\beta)}$, where $\theta_d(\beta) = (\beta\wedge 1)\cdot 1_{\{\beta< d+1\}} + \frac{2\beta \cdot 1_{\{\beta\geq d+1\}}}{d+1+\beta}$. 
\end{corollary}

Next, we study the ``variance'' term. 
It arises from the error of traditional topic modeling on hyperword counts. 
Since Topic-SCORE is applied in this step, we adapt the analysis in \cite{ke2024entry} to obtain the following bound. Its proof includes checking that under Assumptions~\ref{assump1}-\ref{assump3},
all the regularity conditions in \cite{ke2024entry} are satisfied in the reduced model for hyperword counts, which is non-obvious and needs a careful proof.    
\begin{lemma}[Variance] \label{lem:var-new}
Suppose Assumptions~\ref{assump1}-\ref{assump3} are satisfied,  $N\geq \log^3(n)$, $\log(N)=O(\log(n))$, and     
$(Nn)^{-1}\log^2(n)\leq \epsilon^d \leq \log^{-3}(n)$. As $n\to\infty$,  
with probability $1-o(n^{-2})$, for all $1\leq k\leq K$,  
\[
\max_{z_0\in {\cal Z}}|\widehat{\mathcal{A}}_k(z_{0}) - \widetilde{\mathcal{A}}_k(z_{0})|\leq C\sqrt{\frac{\log(n)}{Nn\epsilon^{d}}}.
\] 
\end{lemma}


A key observation from Lemma~\ref{lem:var-new} is that, unlike in classical KDE, the bandwidth $h$ does not affect the rate of the ``variance'' term. In classical KDE, the effect of $h$ on variance stems from the independence of samples: a larger 
$h$ aggregates more independent samples, leading to lower variance. In contrast, our variance analysis is based on strong \textit{entry-wise} bounds for 
$\widehat{A}_k^{\text{net}}$,  	
without assuming independence across its entries. This fundamental difference explains why $h$ does not impact the rate. Thanks to this appealing property, we can determine $h$ sorely based on minimizing the ``bias'' term (as we did in Corollary~\ref{cor:bias}). 

Lemma~\ref{lem:var-new}
and Corollary~\ref{cor:bias} together lead to a trade-off in choosing $\epsilon$. With the optimal choice of $\epsilon$, we can deduce the final error rate as follows: 
\begin{theorem} \label{thm:UB}
Under the conditions of Lemma~\ref{lem:var-new}
and Lemma~\ref{lem:bias-new}, suppose we choose $\epsilon=\bigl(\frac{Nn}{\log(n)}\bigr)^{\frac{1}{ 2\theta_d (\beta) + d}}$. Let $\delta_n=1/\log(n)$. With probability $1-o(n^{-2})$, for all $1\leq k\leq K$, 
\[
\spacingset{1}
\max_{z_0\in {\cal Z}_{\delta_n}} |\widehat{\mathcal{A}}_k(z_{0})-\mathcal{A}_k(z_{0})| 
\leq C \biggl(\frac{\log(n)}{Nn}\biggr)^{\theta^*_d(\beta)}, \quad \mbox{where}\quad \theta^*_d(\beta)= 
\begin{cases}
\frac{\beta}{2\beta+d}, & \mbox{if }0<\beta < 1,\cr
\frac{1}{2+d}, & \mbox{if }1\leq \beta < d+1,\cr
\frac{2\beta}{(4+d)\beta+d(d+1)}, &\mbox{if }\beta \geq d+1. 
\end{cases}
\resetspacing
\]
\end{theorem}

\subsection{Minimax Lower Bound with Discussions of Rate Optimality} \label{subsec:LB}

To evaluate the optimality of the rate established in Theorem~\ref{thm:UB}, we provide an information-theoretic lower bound. When $K=1$, $\Omega_i(\cdot)={\cal A}_1(\cdot)$ for $1\leq i\leq n$. It follows from the equivalent form of PPTM in \eqref{PTM-equivalence} that all embeddings are sampled from ${\cal A}_1(\cdot)$. Hence, the problem reduces to the classical KDE, and the optimal rate is $(Nn)^{-\frac{\beta}{2d+\beta}}$ for the integrated $\ell^1$-loss (e.g., \citep{hasminskii1990density}). 
When $K\geq 2$, we need to establish a lower bound for ourselves.  
Let $\Phi_{n,N}^\star :=\Phi_{n,N}^\star(K,d,c_1,c_2,c_3,c_4,c_5)$ denote the collection of $(\mathcal{A}_{1:K},W)$ that satisfy Assumptions~\ref{assump1}-\ref{assump3}. 

\begin{theorem} \label{thm:LB}
As $n\to\infty$, suppose $N\geq \log^3(n)$ and  $\log(N)=O(\log(n))$. There exist constants $C_1>0$ and $\delta_0\in(0,1)$ such that for all sufficiently large $n$, 
    \[
        \underset{\widehat{\mathcal{A}}}{\inf} \underset{(\mathcal{A}_{1:K},W)\in \Phi_{n,N}^\star}{\sup} \mathbb{P}\Biggl(
        \sum_{k=1}^K \int_{{\cal Z}} |\widehat{\mathcal{A}}_k(z)-\mathcal{A}_k(z)|dz \geq C_1\Bigl(\frac{1}{Nn}\Bigr)^{\frac{\beta}{2\beta+d}}\Biggr) \geq \delta_0. 
    \]
\end{theorem}

The proof of Theorem~\ref{thm:LB} differs from the lower bound analysis in nonparametric density estimation (which is a special case of $K=1$). 
Extending the least-favorable configuration to \( K>1 \) while ensuring that \( {\cal A}_k \)'s are simultaneously smooth and satisfy Assumptions~\ref{assump1}--\ref{assump3} presents a challenge.

We compare the lower bound in Theorem~\ref{thm:LB} with the upper bound in Theorem~\ref{thm:UB}. By Theorem~\ref{thm:UB}, with probability $1-o(n^{-2})$, for $\delta_n=1/\log(n)$,  the TRACE estimator satisfies:
\[
\sum_{k=1}^K \int_{{\cal Z}_{\delta_n}} |\widehat{\mathcal{A}}_k(z)-\mathcal{A}_k(z)|dz\leq
C \biggl(\frac{\log(n)}{Nn}\biggr)^{\theta^*_d(\beta)}. 
\]
First, there is a subtle difference in the integral on the left-hand side. As explained in Section~\ref{subsec:UB}, this is due to a boundary effect and can be mitigated using boundary correction techniques. Second, there is an additional $\log(n)$-factor in the upper bound, which arises from the analysis of Topic-SCORE. When $n$ is sufficiently large, this logarithmic factor is dominated by the polynomial factor in the bounds.  
Finally, we check the polynomial factor, which is $\frac{\beta}{2\beta+d}$ in the lower bound and $\theta_d^*(\beta)$ in the upper bound. The two match when $\beta\leq 1$, giving optimality of our method in this regime. 




\vspace*{-5pt}
\subsection{Error Rate for Estimating Topic Weight Vectors} \label{subsec:W-rate}
\vspace*{-5pt}

In \eqref{def:hat-w}, we have introduced a regression estimator $\widehat{w}_i$. The next theorem presents 
its error rate: 
\begin{theorem} \label{thm:estW}
 Under the conditions of Lemma~\ref{lem:bias-new}, suppose $\epsilon$ satisfies the requirement in Lemma~\ref{lem:var-new}. With probability $1-o(n^{-2})$, $n^{-1}\sum_{i=1}^n \|\widehat w_i - w_i\|_1\leq    C\sqrt{\frac{\log(n)}{Nn\epsilon^d}} + C\sqrt{\frac{\log(n)}{N}}$. 
\end{theorem}

In this 
bound, the first term is due to the error of estimating $A_1^{\text{net}}, \ldots,A_K^{\text{net}}$, and the second term originates from the noise in the hyperword counts. As $\epsilon$ varies, the second term does not change. Hence, we can choose a properly large $\epsilon$ so that the first term is  dominated by the second term: 
\begin{corollary}\label{cor:estW}
    Under the conditions of Lemma~\ref{lem:bias-new}, if we choose $\epsilon$ such that  $n^{-1}\log(n)\leq \epsilon^d\leq \log^{-3}(n)$, then with probability $1-o(n^{-2})$, $n^{-1}\sum_{i=1}^n \|\widehat{w}_i - w_i\|_1\leq CN^{-1/2}\sqrt{\log(n)}$. 
\end{corollary}

We observe that the choices of $\epsilon$ for estimating topic densities and for estimating topic weight vectors could be different. When estimating ${\cal A}_k(\cdot)$, there is a bias-variance trade-off, so the optimal order for $\epsilon$ is unique. However, for estimating $w_i$, we don't need an estimate of ${\cal A}_k(\cdot)$. The error only depends on how well we estimate the induced model for hyperword counts. A cruder net can reduce the dimensionality of the induced model, hence preferred. 
We also note that the rate for estimating $W$ in our PPTM framework coincides with that in the traditional topic model   \citep{wu2023sparse}.

\vspace*{-1cm}
\section{Simulations} \label{sec:Simu}
\vspace*{-5pt}
We use simulations to study the performance of our method and compare it with the word-count-based method Topic-SCORE \citep{ke2022using}. The neural topic modeling approaches (see Section~\ref{subsec:literature}) are for different model frameworks, so the comparison with these approaches is deferred to Section~\ref{sec:RealData}. 

To enable a comparison with Topic-SCORE, we must generate words together with embeddings. We achieve this by letting each topic density be a Gaussian mixture: ${\cal A}_k=\sum_{j=1}^p a_{jk} {\cal N}(\mu_j, I_d)$, where $\mu_j\in\mathbb{R}^d$ and $a_{jk}$'s are from entries of a  standard topic matrix $A=[A_1, A_2,\ldots, A_K]$. It follows that the intensity measure for document $i$ becomes $\Omega_i = \sum_{j=1}^p \gamma_{ij} {\cal N}(\mu_j, I_d)$, with $\gamma_{ij}=\sum_{k=1}^K w_i(k)a_{jk}$. Equivalently, for each word in this document, we draw it from the vocabulary using the PMF $\gamma_i=(\gamma_{ik})_{k=1}^p$; given this word is the $j$-th word in the vocabulary, we draw a contextualized embedding using ${\cal N}(\mu_j, I_d)$. In this way, we obtain both words and embeddings. 
We evaluate the performances of TRACE and Topic-SCORE using the following empirical \(\ell^{1}\)-loss functions:
\begin{equation}\label{eq:simloss}
\widehat{\mathcal{L}}_{\text{TRACE}} = \frac{1}{\lambda}\sum_{k=1}^K \sum_{j=1}^p \bigl\lvert \widehat{\mathcal{A}}_{k}(z_j)-\mathcal{A}_{k}(z_j)\bigr\rvert,\quad  \widehat{\mathcal{L}}_{\text{TSCORE}}=\sum_{k=1}^{K}\bigl\lVert \widehat{A}_{k}-A_{k} \bigr\rVert _{1}=\sum_{k=1}^{K}\sum_{j=1}^{p}\bigl\lvert \widehat{a}_{k,j}-a_{k,j} \bigr\rvert.
\end{equation}

Given $(n, N, K, d, p)$ and two parameters $( \rho, \tau)$, first, we generate the topic weight matrix $W$: For each $1\leq k\leq K$, set \(w_i = e_k\) for \((k-1)\rho + 1 \leq i \leq k\rho\), where \(e_k\) is the $k$-th standard basis of $\mathbb{R}^K$; draw other $w_i$'s by sampling its entries from \(\text{Unif}(0, 1)\) and re-normalizing to have a unit $\ell^1$-norm. 
Next, we set the topic densities as ${\cal A}_k = \sum_{j=1}^p a_{jk}{\cal N}(\mu_j, I_d)$, for $1\leq k\leq K$, where $\mu_1, \ldots, \mu_p$ and $A$ are constructed as follows: 1) Obtain \(\mu^*_1, \ldots, \mu^*_p \overset{iid}{\sim} \mathcal{N}(0, I_d)\) and 
let $\mu_j=\mu_j^*$ if $\|\mu_j^*\|\leq \sqrt{d}$, and $\mu_j = \frac{\sqrt{d}}{\|\mu_j^*\|}\mu_j^*$ otherwise. 2) Obtain a matrix $\widetilde{B}$: Let \(b_{k,j} = \mu_j^2(k) / \sum_{\ell=1}^K \mu_j^2(\ell)\) and  $\kappa_{j}=\mathop{\arg\max}\limits_{1\leq k\leq K}b_{k,j}$. Given $\tau\in (0.5, 1)$, we modify $b_{k,j}$ to $\widetilde{b}_{k,j}$ by letting $\widetilde{b}_{k,j}=b_{k,j}$ if  $b_{\kappa_j, j}<\tau$ and $\widetilde{b}_{k,j}=1\{ k=\kappa_j\}$ if $b_{\kappa_j, j}\geq \tau$. 3) Solve \(f_{1}, \ldots, f_{p}\) from minimizing  $\sum_{k=1}^{K}(\sum_{j=1}^{p} f_{j} \widetilde{b}_{k,j} - 1 )^{2}$, 
subject to \( f_{j} \geq 0\). There exist infinitely many solutions that yield zero value in the objective, and we find one such solution by using a projected gradient descent algorithm. 4) Let $A_k(j) = f_j \widetilde{b}_{k,j}$ for $1\leq k\leq K$ and $1\leq j\leq p$. 
Finally, we generate words and embeddings as described above. 
In this data generating process, $\rho$ controls the fraction of pure documents in each topic, and $\tau$ controls the size of anchor region.

\spacingset{1}
\begin{figure}[tb!]
\begin{subfigure}[b]{.52\textwidth}
\includegraphics[width=1\textwidth, height = .75\textwidth]{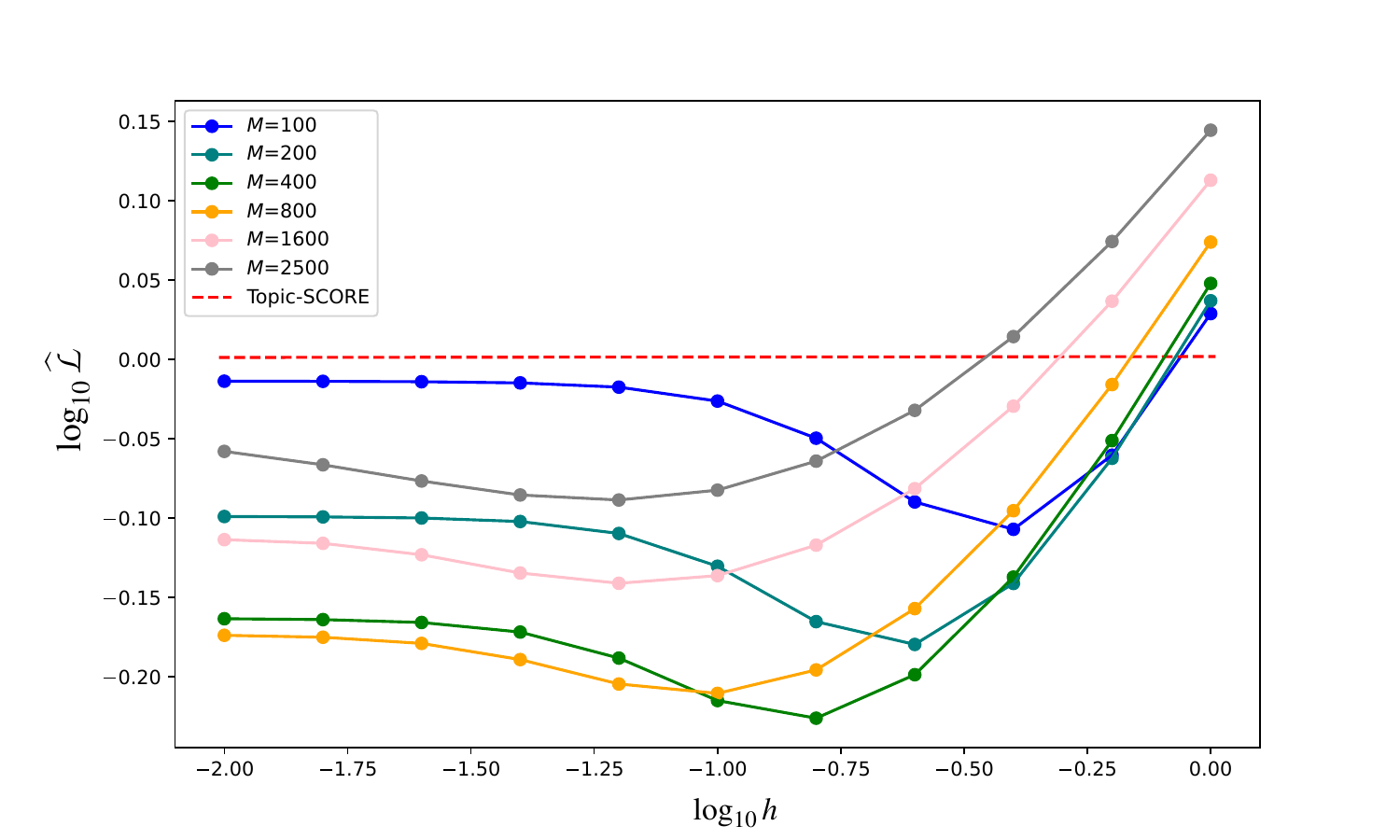}
\caption{\small Comparison of $\log_{10}\widehat{\mathcal{L}}$ versus $\log_{10}h$ for TRACE with varying $M$ and Topic-SCORE.} \label{fig:simulation-exp1}
\end{subfigure}
\hspace{5pt}
\begin{subfigure}[b]{.45\textwidth}
\includegraphics[width=.8\textwidth, height=0.42\textwidth, trim=0 0 880 0, clip=true]{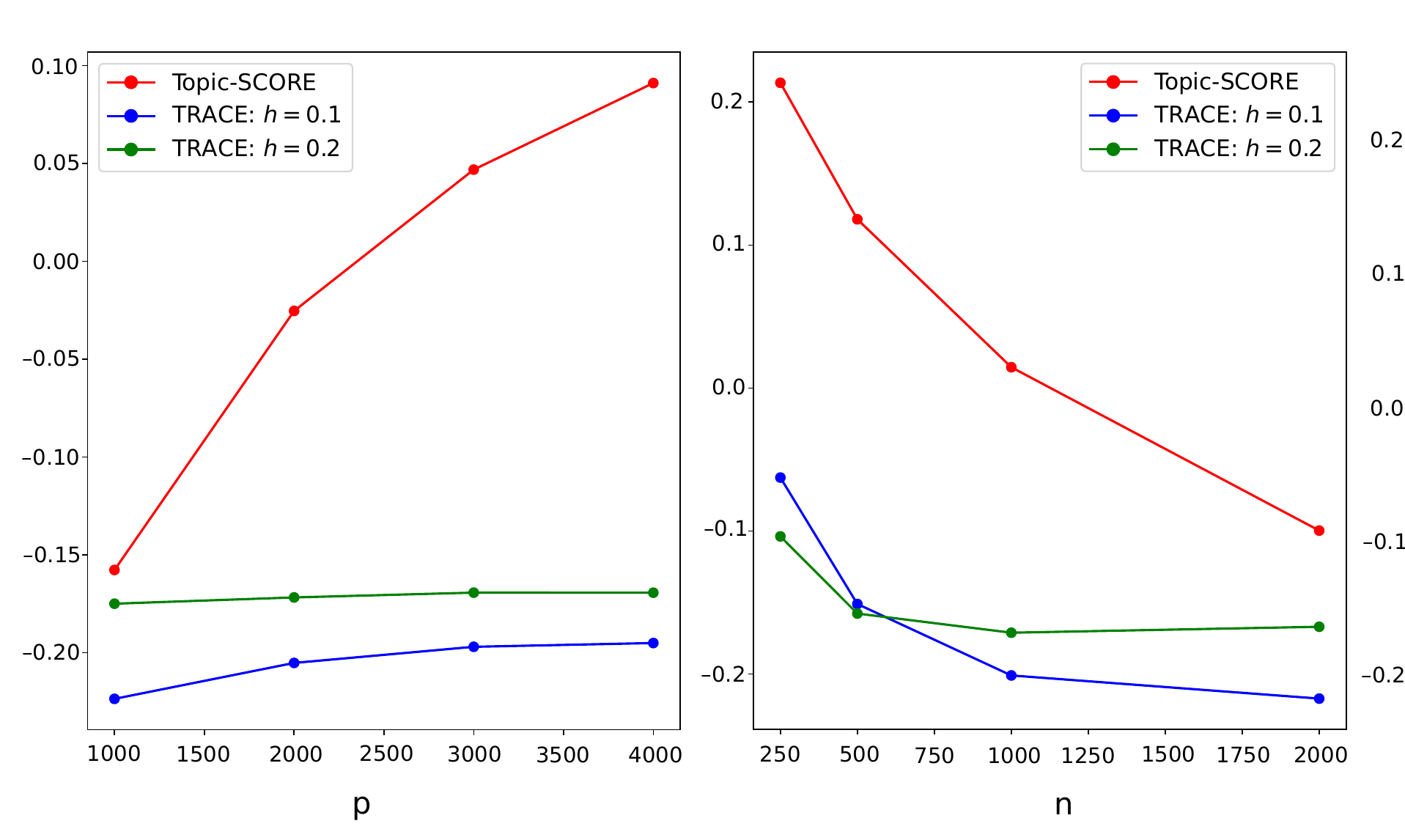}\\
\includegraphics[width=.8\textwidth, height=0.42\textwidth,  trim=890 0 0 0, clip=true]{PDF/Sim_1_p_n_N_K.pdf}
\caption{\small Comparison of $\log_{10}\widehat{\mathcal{L}}$ with varying $p$, $n$, $N$, or $K$ for TRACE and Topic-SCORE.} \label{fig:simulation-exp2}
\end{subfigure}
\caption{The simulation results.}
\end{figure}
\resetspacing

In the first scenario, we fix 
$(p, n, N, K, \tau, \rho)= (2500, 1000, 200, 3, 0.8, 5)$
and study the effects of varying the number of hyperwords \(M\) and the bandwidth \(h\). \cref{fig:simulation-exp1} presents the logarithm of the empirical loss \(\widehat{\mathcal{L}}_{\text{TSCORE}}\) and \(\widehat{\mathcal{L}}_{\text{TRACE}}\) against the logarithm of bandwidth \(h\) for Topic-SCORE and TRACE with \(M \in \{ 100, 200, 400, 800, 1600, 2500\}\). All results are averaged over 100 repetitions. We observe that for any fixed \(M\), the loss initially decreases and then increases as \(h\) grows, with the optimal bandwidth roughly between \(0.1\) and \(0.5\). A smaller \(h\) is generally less harmful than an overly large \(h\). Moreover, when \(h \leq 0.5\) is fixed, the loss first decreases and then increases as \(M\) grows from \(100\) to \(2500\), with the optimal performance achieved at \(M = 400\) or \(800\), aligning with the bias-variance trade-off suggested in Section~\ref{subsec:UB}. The loss for TRACE is generally smaller than that for  that of Topic-SCORE.

In the second experiment, we study how the performance changes with model parameters. 
In each of the four sub-experiments, one of the parameters \(p, n, N, K\) is varied while keeping all the other parameters the same as in Experiment~1. 
TRACE has two tuning parameters $(M, h)$. We fix $M=800$ and consider two choices of $h$: $0.1$ and $0.2$. See \cref{fig:simulation-exp2}. 
Consistent with both intuition and theory, the empirical loss increases with \(p\) and \(K\) but decreases with \(n\) and \(N\). Furthermore, TRACE with bandwidth \(h = 0.1\) or \(0.2\) consistently outperforms Topic-SCORE across all settings. Notably, the empirical loss for TRACE is less sensitive to the vocabulary size \(p\) compared to Topic-SCORE, showing only a \(10\%\) increase as \(p\) scales from \(1000\) to \(4000\).

\vspace*{-.5cm}
\section{Real-Data Applications} \label{sec:RealData}
\vspace*{-5pt}


\subsection{The Associated Press Corpus} \label{subsec:AP}
\vspace*{-5pt}

The Associated Press corpus, originally presented in \cite{harman1overview}, consists of $2246$ news articles from the Associated Press in the early 1990s. After removing articles with fewer than $50$ words, we retain \( n = 2106 \) articles. Preprocessing involves removing stop words (using the \texttt{nltk} Python library) and words with a total count of fewer than $10$ in the corpus. The cleaned articles are then input into the BERT model provided by the \texttt{flair} Python library to generate contextualized word embeddings. This process yields approximately $390,000$ raw embeddings, each of dimension $768$. We then utilize a random subset ($20\%$) of them to train a UMAP \citep{mcinnes2018umap} model to reduce the dimension to \( d = 10 \).\spacingset{0}\footnote{{More discussion on the choice of $d$ can be found in Appendix~\ref{supp:umap} and Appendix~\ref{supp:umap-d} of the supplemental material. }} \resetspacing 
We apply our algorithm to the projected embeddings with \( M=600 \) hyperwords. The number of topics, \( K=7 \), is determined by examining the scree plot of the singular values from this matrix.\spacingset{0}\footnote{{More discussion on how the topic structure evolves as the number of $K$ increases is provided in Appendix~\ref{supp:numberoftopics}.}} \resetspacing 
The plug-in traditional topic modeling algorithm is Topic-SCORE  \citep{ke2022using}. In the kernel smoothing step, we use a Gaussian kernel, with the bandwidth \( h \) selected based on a maximum entropy criterion. Further details on the implementation and parameter tuning are provided in Appendix \ref{supp:AlgAP}.

\begin{figure}[tb!]
\centering
\includegraphics[width=.6\textwidth]{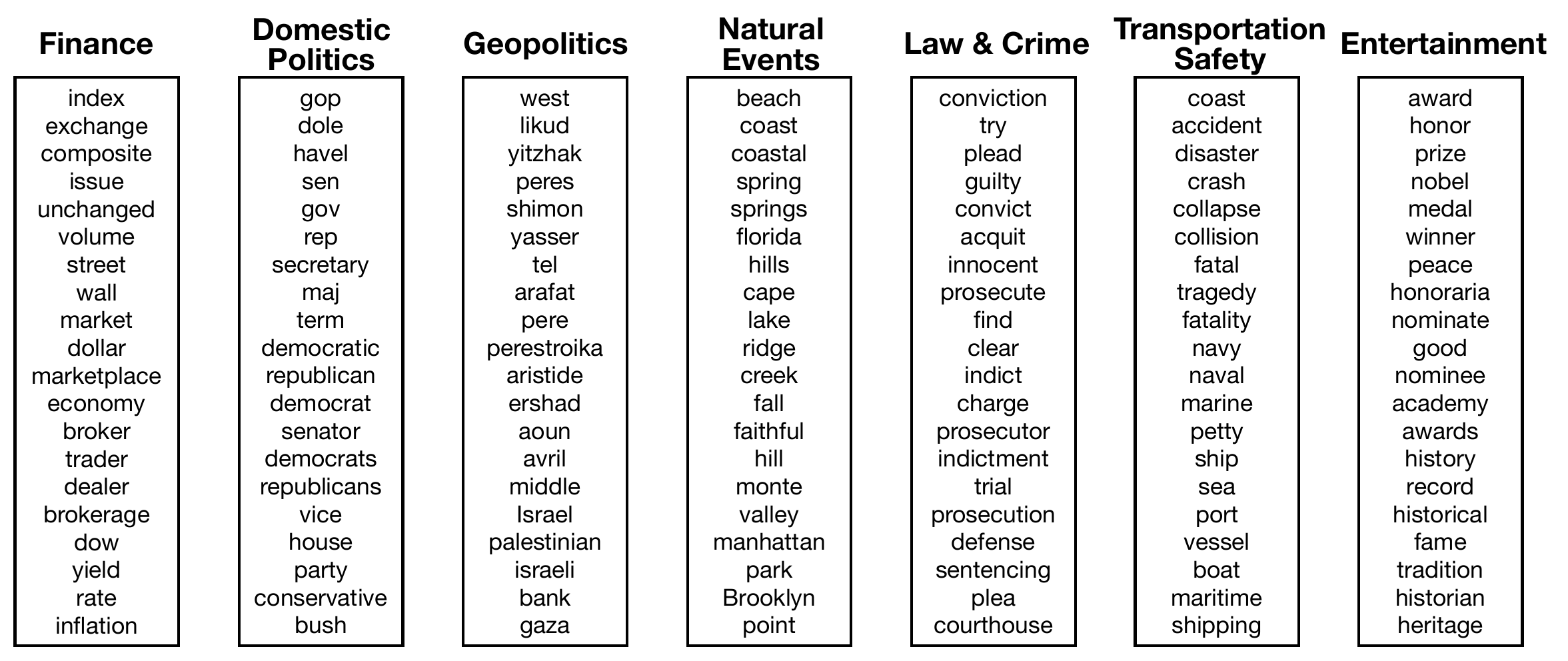}
\caption{\small The top 20 anchor words for each estimated topic measure on the AP dataset.} \label{fig:AP-anchor}
\end{figure}

For each \( \widehat{\mathcal{A}}_k(z) \), we compute its normalized version \( \widehat{\mathcal{B}}_k(z) = \widehat{\mathcal{A}}_k(z)\bigm/\sum_{\ell=1}^K \widehat{\mathcal{A}}_\ell(z) \), and then select \( z \)'s, i.e., the embeddings, with the largest \( \widehat{\mathcal{B}}_k(z) \) for each topic $k=1,\ldots,K$. The words associated with these embeddings are identified as the anchor words for topic \( k \). To enhance interpretability, we apply lemmatization using the \texttt{SpaCy} Python library to the anchor word list. We then review the anchor words and relevant documents for each topic and manually assign a descriptive name. The top 20 anchor words and assigned names for each topic are presented in \cref{fig:AP-anchor}.

\spacingset{1}
\resetspacing

\spacingset{1}
\begin{figure}[tb!]
\centering
\includegraphics[width=.6\textwidth, trim=80 100 40 48, clip=true]{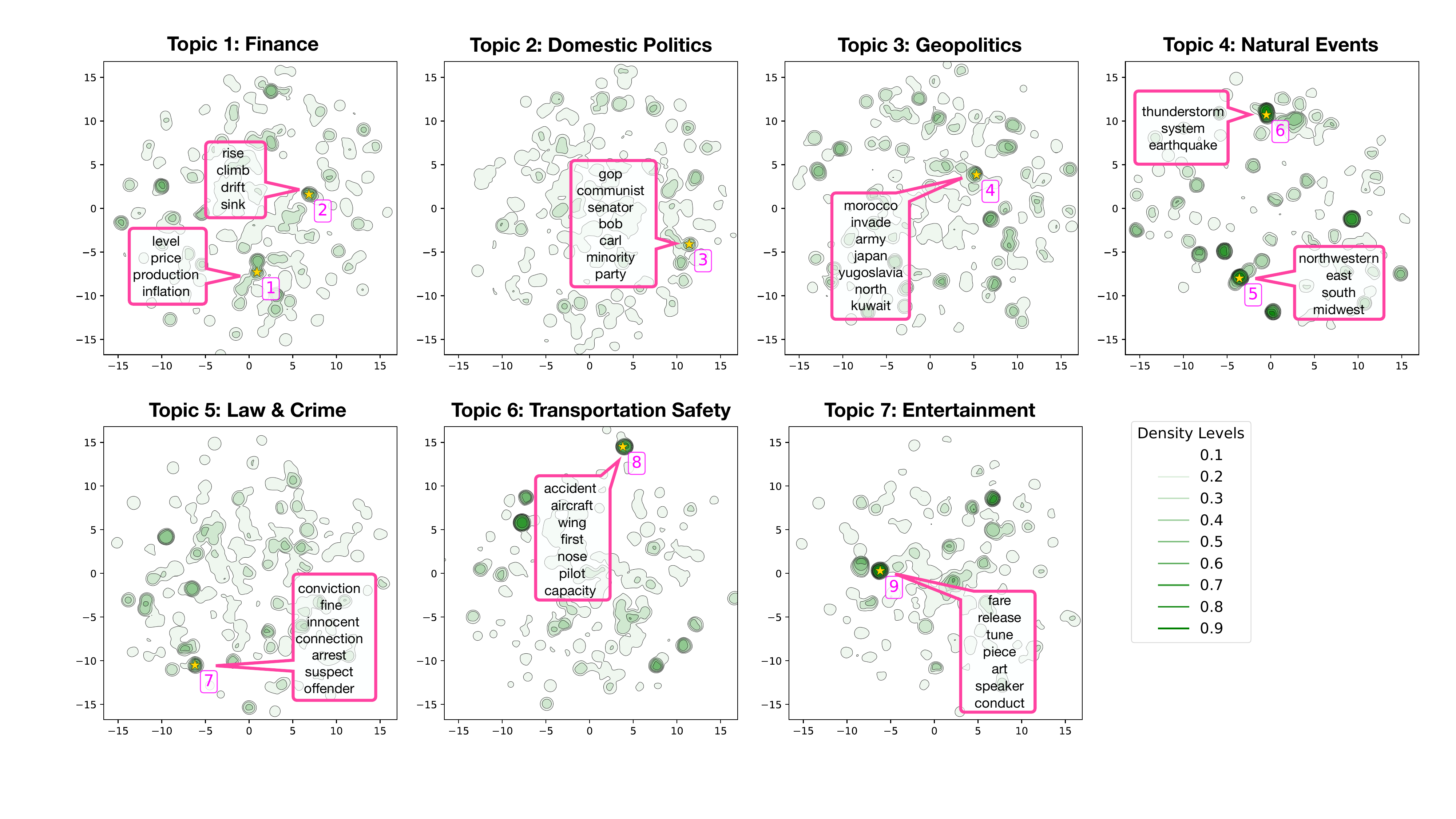}
\caption{\small Estimated topic measures on the AP dataset ($K=7$). For each topic, we compute $\widehat{\cal B}_k(z)=\widehat{\cal A}_k(z)/[\sum_{\ell=1}^K \widehat{\cal A}_\ell(z)]$ and plot its contour in a projected two-dimensional space (the projection is only made for visualization). Nine anchor regions are marked in the plots, each indicating a group of nearly-anchor words. Some representative words in each numbered region are given in the plots.} \label{fig:AP-topics}
\end{figure}
\resetspacing

Three of the topics (\textit{Finance}, \textit{Law \& Crime}, and \textit{Geopolitics}) align directly with those identified in \cite{ke2022using}, where Topic-SCORE was applied to the normalized word count matrix instead of the hyperword matrix. However, the results from Topic-SCORE become less interpretable as the number of topics increases for this dataset, while our method maintains clear topic separation even as \( K \) increases (see Appendix \ref{supp:apadditional} in the supplemental material). When \( K=7 \), as suggested by the scree plot, our method identifies four additional topics: \textit{Domestic Politics}, \textit{Natural Events}, \textit{Transportation Safety}, and \textit{Entertainment}. 
Notably, the topic \textit{Entertainment}, which is absent in Topic-SCORE, may be challenging to detect without external embedding information due to its diffuse word distribution. By comparison, the first three topics (\textit{Finance}, \textit{Law \& Crime}, and \textit{Geopolitics}) have more concentrated word distributions. For instance, the topic \textit{Law \& Crime} is dominated by terms such as \textit{conviction}, \textit{charge}, \textit{gun}, \textit{shot}, and \textit{police}. 
This analysis underscores the integration of word embeddings in our framework as a principled and effective way to handle low-frequency anchor words. By grouping multiple embeddings of low-frequency words into hyperwords, our method establishes a frequency structure that is easier to discern, facilitating the identification of more nuanced topics. Consequently, TRACE exhibits a stronger capacity to capture such ``weak'' topics, which are often overlooked by traditional word-count-based methods. Additional details on the comparison with Topic-SCORE can be found in Appendix \ref{supp:AlgAP} of the supplemental material.

\spacingset{1}
\begin{figure}[tb!]
\centering
\includegraphics[width=.88\textwidth]{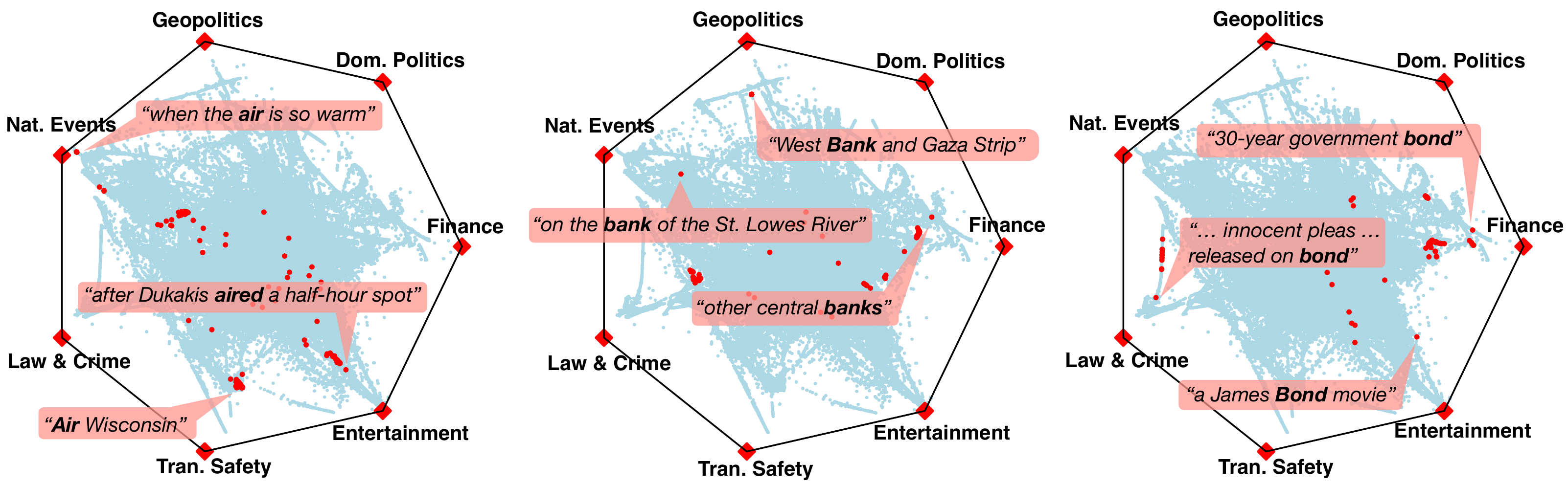}
\caption{\small Heptagon plots showing the embeddings of three words, \textit{air} (left), \textit{bank} (middle), and \textit{bond} (right), across topics of the AP dataset. Each vertex of the heptagon represents a topic, and red dots indicate embeddings associated with the respective word. Annotated examples highlight different contexts in which these words appear, demonstrating their semantic variability across topics.} \label{fig:AP-context}
\end{figure}
\resetspacing

In \cref{fig:AP-topics}, we visualize the contours of the re-normalized topic densities \( \widehat{\mathcal{B}}_1(\cdot), \ldots, \widehat{\mathcal{B}}_7(\cdot) \), projected onto two dimensions for easier interpretation. Each topic exhibits characteristic embeddings clustered around high-density regions, referred to as anchor regions, in the contour plot. 
E.g., the topic \textit{Natural Events} features multiple anchor regions: one associated with natural phenomena such as \textit{thunderstorm} and \textit{earthquake}, and another focused 
on directional terms like \textit{east} and \textit{south}.
 
Another key advantage of our framework is its ability to incorporate the context in which a word appears. We illustrate this using heptagon plots, defined as follows: Let \( \widehat{v}_1, \widehat{v}_2, \ldots, \widehat{v}_7 \) denote the vertices of a standard heptagon in \( \mathbb{R}^2 \). For a fixed word (e.g., \textit{bond}), each embedding \( z \) of this word is mapped to a point \( \xi(z) := \sum_{k=1}^7 \widehat{\mathcal{B}}_k(z) \widehat{v}_k \) within the standard heptagon, where \( \widehat{\mathcal{B}}_1(\cdot), \ldots, \widehat{\mathcal{B}}_7(\cdot) \) represent the re-normalized topic densities. The resulting point clouds \( \{\xi(z)\} \) for the embeddings of three words—\textit{air}, \textit{bank}, and \textit{bond}—are visualized in \cref{fig:AP-context}.
The position of \( \xi(z) \) within the heptagon reflects its associations with different topics. For instance, when \( \xi(z) \) is near a vertex, the embedding is primarily linked to a single topic. As the context of a word changes, its embedding \( z \) also shifts, leading to variations in its topic association. As shown in \cref{fig:AP-context}, the word \textit{bond} forms multiple embedding clusters. In the context of \textit{government bond}, it is predominantly associated with the topic \textit{Finance}. In contrast, when referring to \textit{James Bond}, the same word is more closely related to the topic \textit{Entertainment}.

We also compare our method with five other topic modeling methods that use word embeddings: BERTopic \citep{grootendorst2022bertopic}, ETM \citep{dieng2020topic}, TopClus \cite{meng2022topic}, FASTopic \citep{wu2024fastopic}, and ECRTM \citep{wu2023effective}. We set $K=7$ and use the BERT (word or document) embeddings for all methods (for ETM, the original method used Word2Vec embeddings; for a fair comparison, we replaced them by BERT embeddings). For each method, we used the code provided by authors or public libraries and adopted the default hyper-parameters. The top 20 representative words for each topic are presented in \cref{tb:AP-other-methods}.



We found that BERTopic strongly favors proper nouns as representative words, consistent with observations in \cite{meng2022topic}, making its topics harder to summarize; moreover, three out of seven topics (Topics 1, 3, 4) are heavily weighted toward geopolitical issues. ETM tends to select higher-frequency words, leading to significant overlap of representative words across topics. For example, the word ``national'' appears in multiple topics (Topics 1, 2, 4, 5, 7), a pattern that is also noted in \cite{meng2022topic}. TopClus seems to prioritize syntactic similarity among words over topic relevance and coherence (e.g., see Topics 2 and 4).  
In comparison,  FASTopic and ECRTM produce higher-quality topic clusters. However, FASTopic sometimes groups loosely related or adjacent themes into one topic, whereas ECRTM slices themes more narrowly and thus creates overlapping, near-duplicate topics (e.g., its 2nd and 3rd clusters both map to Domestic Politics).
Our method, however, achieves clear distinctions across topics while maintaining coherence within each topic, as shown in \cref{fig:AP-anchor}. 


\spacingset{.8}
\begin{table}[tb!]
\centering
\caption{\small Top 20 words for each topic from five benchmark methods.}
\label{tb:AP-other-methods}
\scriptsize
\scalebox{0.5}{
\begin{tabular}{c|p{6.2cm}|p{6.2cm}|p{6.2cm}|p{6.2cm}|p{6.2cm}}
\toprule
\textbf{Topic} & \textbf{BERTopic} & \textbf{ETM} & \textbf{TopClus} & \textbf{FASTopic} & \textbf{ECRTM} \\
\midrule
1 & \footnotesize demjanjuk, monet, havel, cubans, jackpot, czechoslovak, cdy, nauvoo, delvalle, hildreth, clr, mofford, gotti, sipc, wynberg, bono, cox, bosch, ames, blier 
  & \footnotesize president, new, soviet, state, national, two, south, chief, secretary, government, police, federal, congress, party, first, law, senate, bill, general, court 
  & \footnotesize appointment, elimination, adjustment, relax, bonus, regulate, rating, cancel, limit, adjusted, downward, improvement, abolished, ratio, cancellation, payment, rebound, rate, rebuild, payroll 
  & \footnotesize aids, disease, patients, researchers, abortion, virus, smoking, cancer, treatment, ban, tests, waste, blood, test, teachers, animals, testing, suit, epa, health 
  & \footnotesize mother, friends, movie, musical, love, actor, museum, music, actress, born, loved, husband, theater, wife, concert, art, ticketron, telecharge, teletron, band \\
\midrule
2 & \footnotesize epa, hubbert, pacs, uaw, pac, unleaded, darman, polaroid, abortions, chinn, shuster, skinner, duracell, shamrock, pesticide, batus, turtles, corroon, greenwald, edgemont 
  & \footnotesize official, police, government, first, court, military, state, national, office, war, service, people, report, company, possible, federal, judge, political, history, woman 
  & \footnotesize aired, mentioned, answered, noting, thanked, referring, described, impressed, commenting, interviewed, advised, emphasized, hailed, addressed, revealed, cited, consulted, instructed, citing, acknowledged 
  & \footnotesize dukakis, billion, budget, senate, tax, industry, percent, sen, vice, rep, plant, company, republican, spending, farmers, income, chairman, campaign, cost, jackson 
  & \footnotesize vargas, electoral, anc, votes, fujimori, students, elections, whites, party, cloudy, llosa, runoff, ballots, aristide, fair, blacks, voters, vote, mandela, voting \\
\midrule
3 & \footnotesize west, likud, yitzhak, peres, shimon, yasser, tel, arafat, pere, perestroika, aristide, ershad, aoun, avril, middle, israel, palestinian, israeli, bank, gaza
  & \footnotesize people, police, like, company, new, service, last, house, back, state, life, time, city, court, government, world, called, work, war, first
  & \footnotesize governmental, mayoral, vatican, political, politician, parliamentary, commandant, judicial, governing, commissioner, legislative, council, regulator, constitutional, electoral, presidency, inspector, courtroom, nationalist, democracy
  & \footnotesize yen, cents, index, futures, cent, stocks, trading, shares, ounce, dealers, dollar, unchanged, earnings, volume, stock, investors, dow, analysts, exchange, securities
  & \footnotesize budget, sen, bush, spending, deficit, gephardt, senate, cuts, dole, bill, gorbachev, farm, darman, bushs, administration, rep, trade, sasser, reagan, democrats \\
\midrule
4 & \footnotesize tass, republics, aoun, politburo, unification, ceausescu, walesa, yeltsin, hezbollah, syrian, militia, azerbaijan, kgb, perestroika, grigoryants, sakharov, kashmir, amal, sinhalese, kasparov
  & \footnotesize two, united, new, american, year, national, three, police, dukakis, state, end, march, university, help, gorbachev, force, war, good, say, british
  & \footnotesize chairs, exams, courses, dances, medals, performances, universities, lights, clubs, photos, genes, railroads, maps, classes, songs, musicians, rooms, churches, computers, buildings
  & \footnotesize accident, passengers, pilot, ship, shuttle, planes, degrees, river, crew, coast, rain, flight, pilots, snow, northwest, aboard, inches, mph, plane, feet
  & \footnotesize market, dollar, trading, index, yen, rates, rose, traders, unchanged, prices, stocks, analysts, ounce, stock, investors, volume, markets, futures, higher, share \\
\midrule
5 & \footnotesize aspirin, gesell, rowan, yates, ferret, zaccaro, gacy, warmus, steiger, getz, gisclair, smokers, bricklin, benedict, pekin, combe, menorah, trehan, ruffin, schneidman
  & \footnotesize million, thursday, two, tuesday, wednesday, official, year, percent, york, friday, billion, dukakis, monday, american, three, country, national, government, cent
  & \footnotesize dna, implant, nutrition, dental, electronic, medication, mechanical, nesting, slaughter, unconscious, rubber, sexuality, tissue, electrical, cereal, vegetable, revenge, drill, meat, massage
  & \footnotesize soviet, gorbachev, troops, communist, iraq, minister, israel, forces, peace, iran, africa, iraqi, moscow, israeli, summit, germany, rebels, military, saudi, soviets
  & \footnotesize water, engineers, venus, quake, accidents, mph, nasa, galileo, spacecraft, rain, feet, richter, savannah, shuttle, infected, nasas, winds, magellan, river, jupiter \\
\midrule
6 & \footnotesize fleisher, ticketron, leek, cosell, telecharge, wallenda, broyles, teletron, moonstruck, worlds, primetime, wednesdays, wed, elvis, presley, musicians, ballet, symphony, orchestra, archive
  & \footnotesize last, million, today, year, united, friday, wednesday, first, percent, week, day, thursday, group, monday, sunday, years, government, south, three, business
  & \footnotesize album, music, theater, comedy, musician, tournament, guitar, stretch, movie, actor, celebrity, championship, coach, poetry, literary, architect, script, choir, player, cartoon
  & \footnotesize film, art, magazine, music, nbc, love, movie, cbs, book, prize, editor, abc, actor, singer, band, museum, wine, sports, shows, network
  & \footnotesize troops, arab, israeli, palestinian, occupied, gaza, palestinians, israel, plo, diplomatic, war, strip, lebanon, soldiers, terrorists, arafat, diplomats, israelis, shultz, syria \\
\midrule
7 & \footnotesize lire, guilders, bullion, gainers, loser, yen, zurich, nikkei, midmorning, ounce, francs, troy, industrials, outnumbered, nyse, rsqb, lsqb, dollar, dealers, unchanged
  & \footnotesize percent, million, year, today, day, national, government, state, police, week, friday, sunday, month, south, billion, home, wednesday, west, city, first
  & \footnotesize avery, holland, alice, mandal, budapest, sony, monte, kat, willis, ontario, swedish, erie, johns, collins, dutch, nagoya, eddie, morton, hop, kong
  & \footnotesize police, trial, prison, hospital, charges, judge, attorney, court, died, arrested, death, investigation, shot, convicted, case, jury, murder, victims, charged, guilty
  & \footnotesize keating, case, gesell, walsh, documents, hearings, attorney, testimony, lawyers, indictment, regulators, judge, justice, criminal, courts, court, appeals, counsel, deconcini, fraud \\
\bottomrule
\end{tabular}
}
\end{table}
\resetspacing

\vspace*{-5pt}
\subsection{The MADStat Corpus} \label{subsec:MADStat}
\vspace*{-5pt}

The MADStat project \citep{ji2021co,ke2023recent} compiled data on \( 83,331 \) papers published in \( 36 \) statistics-related journals between 1975 and 2015. For our analysis, we focus on the text of the paper abstracts. Following \cite{ke2023recent}, we exclude extremely short abstracts (fewer than \( 38 \) words), reducing the dataset to \( n = 50,837 \). Consistent with the preprocessing steps for the AP dataset, we remove stop words and words with a total count of fewer than \( 10 \).
Additionally, in the original dataset, mathematical formulas in abstracts were converted into words representing Greek letters and LaTeX symbols (e.g., \textit{alpha}, \textit{tilde}). These words are often disproportionately frequent in some abstracts, creating an unintended isolated topic dominated by math-heavy abstracts. To mitigate this issue, we further remove \(24\) Greek letters and \(7\) math symbols. The preprocessed abstracts are then embedded using the BERT model, resulting in approximately $3,500,000$ raw embeddings. We apply our method to this dataset with \( K = 13 \), \( d = 10 \), \( M = 2400 \) and \( h = 0.1 \); details on the selection of these parameters can be found in Appendix \ref{supp:Algmadstat} of the supplemental material.


\cref{fig:MADStat-anchor+trend} (left) presents the 16 top ranked anchor words for each topic, selected in a manner similar to the AP dataset, alongside the topic names we manually assigned with assistance from \textit{GPT-4o}. These topics are well-interpreted as distinct research areas within statistics.


We compare the topics identified by our method with those reported in \cite{ke2023recent}, which were obtained by applying Topic-SCORE directly to normalized word counts with \( K = 11 \). A summary is in \cref{tb:MADStat-compare}. When a topic in our results has anchor words similar to a topic in \cite{ke2023recent}, we classify it as a (one-to-one) \textit{correspondence}. Six such pairs are observed, as shown in the first row of \cref{tb:MADStat-compare}.
%
The \textit{Time Series} topic from \cite{ke2023recent} splits into two topics in our results: \textit{Stochastic Process \& Time Series} and \textit{Survival Analysis}. This splitting arises from using contextualized word embeddings, which capture nuanced distinctions based on the context of shared anchor words. Conversely, four topics in \cite{ke2023recent} are recombined in our results. For instance, \textit{Nonparametric Statistics} and \textit{Variable/Model Selection} are distinctly identified in our analysis but were distributed across multiple topics in \cite{ke2023recent}.
Additionally, the \textit{Biomedical Statistics} topic reported in \cite{ke2023recent} is split into several topics in our results. Anchor words related to genetics (e.g., \textit{genome}, \textit{dna}, \textit{chromosome}) are now associated with \textit{Variable/Model Selection}, while those pertaining to epidemiology are aligned with \textit{Survival Analysis} and \textit{Clinical Trial}. Finally, our method identifies a completely new topic, \textit{Social \& Economic Studies}, which was not present in \cite{ke2023recent}.

\spacingset{1}
\begin{figure}[tb!]
\centering
\includegraphics[width=.48\textwidth]{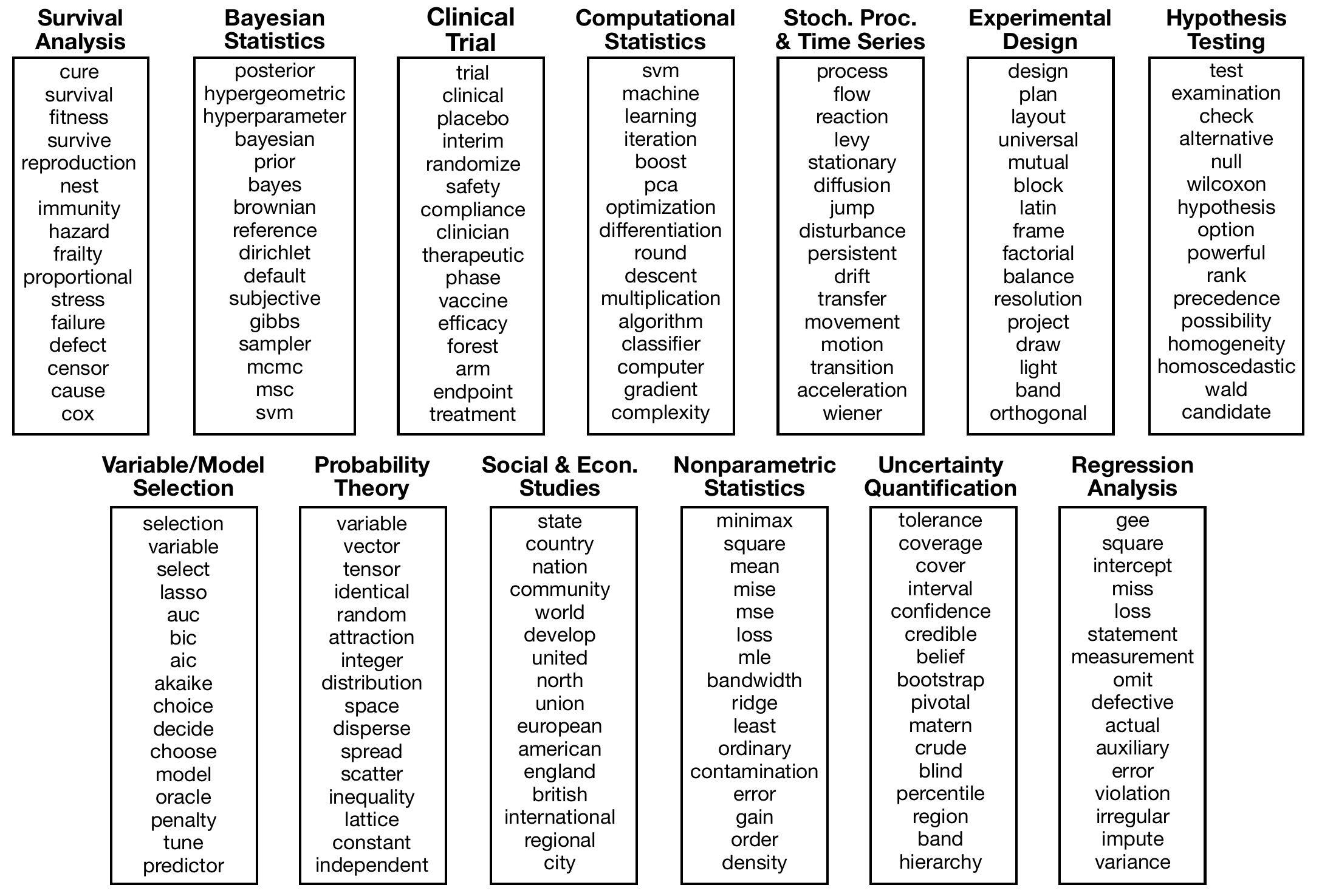}
\includegraphics[width=.5\textwidth, height=.3\textwidth]{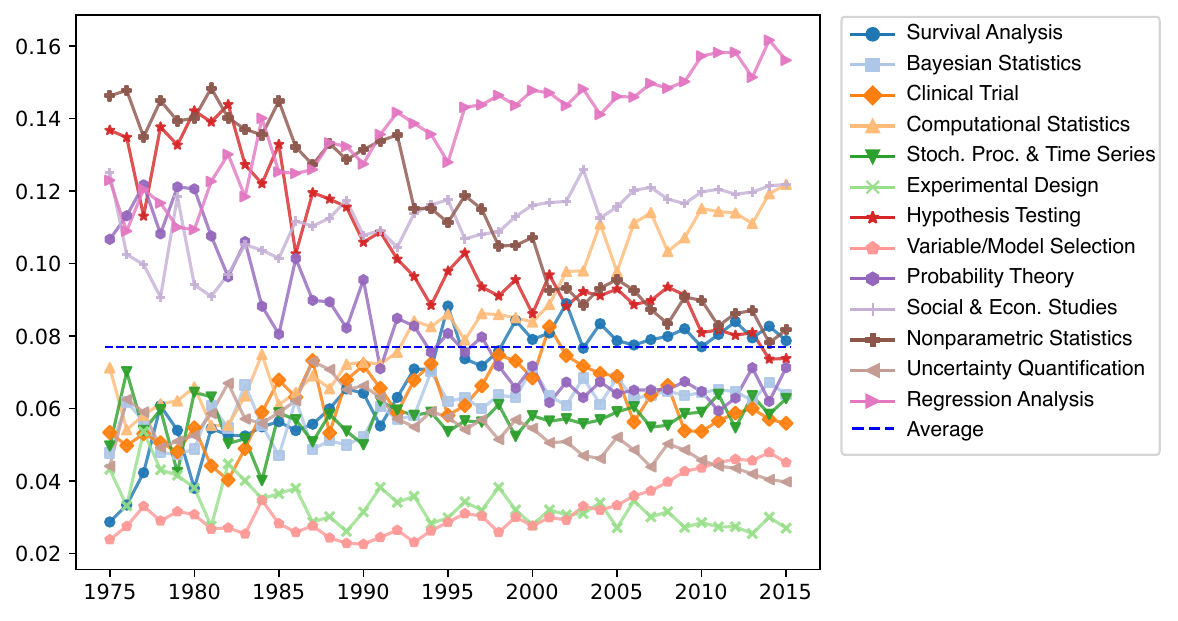}
\caption{\small The top 16 anchor words for each estimated topic measure (left) and the topic trend (right) on the MADStat dataset. In the topic trend, the plain average \( 1/13 \) shown as a dashed blue line for reference} \label{fig:MADStat-anchor+trend}
\end{figure}
\resetspacing

\spacingset{1}
\begin{table}[tb!]
\centering
\caption{\small Comparison between the topics in \cref{fig:MADStat-anchor+trend} and the topics reported in \cite{ke2023recent}.} \label{tb:MADStat-compare}
\scalebox{.6}{
\begin{tabular}{p{3cm} | p{18cm}}
\toprule
Correspondence & Bayesian Statistics, Clinical Trial, Computational Statistics ({\it Machine Learning} in \cite{ke2023recent}), Experimental Design, Hypothesis Testing, Probability Theory ({\it Mathematical Statistics} in \cite{ke2023recent}). \\
\midrule
Splitting \& &  {\it Time Series} in \cite{ke2023recent} $\Longrightarrow$ Stochastic Process \& Time Series (primary), and Survival Analysis.\\ 
Re-combination& {\it Regression} in \cite{ke2023recent} $\Longrightarrow$ Regression Analysis (primary), and Nonparametric Statistics. \\
& {\it Inference} in \cite{ke2023recent} $\Longrightarrow$ Uncertainty Quantification (primary), and Nonparametric Statistics.\\
& {\it Biomedical Statistics} in \cite{ke2023recent} $\Longrightarrow$ Variable/Model Selection, Survival Analysis, and Clinical Trial. \\
& {\it Latent Variable} in \cite{ke2023recent} $\Longrightarrow$ Variable Selection, Computational Statistics, and Regression Analysis.\\
\midrule
Emerging & Social \& Economic Studies.\\
\bottomrule
\end{tabular}}
\end{table}
\resetspacing

\spacingset{1}
\begin{table}[tb!]
\centering
\caption{\small The most relevant topics for two words {\it product} and {\it process} under different contexts.} \label{tb:MADStat-context}
\scalebox{.62}{
\begin{tabular}{p{12cm}l}
\toprule
\ \ Context & Most relevant topic\\
\midrule
{\it ``concerning the Gross Domestic \uline{Product} per capita''} &  Social \& Economic Studies (0.47)\\ 
{\it ``the pairwise likelihood is a  \uline{product} of bivariate likelihoods''} & Probability Theory (0.40)\\
{\it ``predation of chicks causing low  \uline{productivity}''} & Survival Analysis (0.80)\\
{\it ``for renewal \uline{processes} with infinite mean''} & Probability Theory (1.0)\\
{\it ``the \uline{processing} time required to solve ...''} & Computational Statistics (0.38)\\
\bottomrule
\end{tabular}}
\end{table}
\resetspacing


Following a similar approach to the AP dataset, we compute the values \( \widehat{\mathcal{B}}_1(z), \ldots, \widehat{\mathcal{B}}_{13}(z) \) for each embedding \( z \) of a given word. The search includes all inflectional and derivational forms of the word (e.g., \textit{productivity} is included when searching for embeddings of \textit{product}). For the AP dataset, these values are visualized in the plots as in \cref{fig:AP-context}. Here, for each embedding \( z \), we report the textual context along with the topic associated with the largest value among \( \widehat{\mathcal{B}}_1(z), \ldots, \widehat{\mathcal{B}}_{13}(z) \). \cref{tb:MADStat-context} highlights five example contexts for two words, \textit{product} and \textit{process}, with additional examples and the complete values of \( \widehat{\mathcal{B}}_1(z), \ldots, \widehat{\mathcal{B}}_{13}(z) \) provided in the supplemental material. 

Finally, we analyze the topic weights of these paper abstracts. Using the estimator in \eqref{def:hat-w}, we estimate the topic weights \( w_i \), incorporating ridge regularization with a parameter of $0.001$ for stability. For each year from 1975 to 2015, we compute the average of the estimated topic weights \( \widehat{w}_i \) for all papers published in that year, denoting the resulting vector as \( \bar{w}_t \in \mathbb{R}^{13} \). In \cref{fig:MADStat-anchor+trend} (right), we plot the trend of \( \{\bar{w}_t(k)\}_{t=1975}^{2015} \) for each topic. The analysis uncovers several notable trends: First, \emph{Regression Analysis} emerges as the most popular topic overall. Among applied areas, \emph{Social \& Economic Studies} stands out as the leading focus, encompassing nearly all non-biomedical applications. In contrast, biomedical applications are distributed across multiple distinct topics, with \emph{Survival Analysis} ranking as the second most popular applied area. Second, the popularity of several classical topics—such as \emph{Nonparametric Statistics}, \emph{Hypothesis Testing}, and \emph{Probability Theory}—declines gradually over time. Conversely, interest in \emph{Computational Statistics} shows a steady increase throughout the historical timeline, while \emph{Variable/Model Selection} experiences a sharp rise in popularity from 2005 to 2014.

\subsection{Quantitative Performance Metrics and Downstream Task Evaluation}

Topic modeling is an unsupervised problem with no ground truth available in real data. In the literature,   certain performance metrics, such as topic coherence and diversity, have been designed to enable evaluation and comparison. They are computed based on the top-$m$ representative words of learnt topics, for a pre-specified $m$ (e.g., we set $m=20$). Topic coherence (TC) measures the co-occurrence of top words of the same topic, and we use the $C_v$ defined in \cite{roder2015exploring}. Topic diversity (TD) quantifies the separation between topics in terms of non-overlapping top words, and we use the one in  \cite{dieng2020topic}. Both are computed via the \texttt{Gensim} Python package with default settings. 
Inspired by \citet{bianchi-etal-2021-pre,bianchi-etal-2021-cross}, we also define variants of TC and TD that account for the semantic relationship of words. Given a non-contextualized word embedding method (we use the default non-contextualized embeddings given by BERT), let $\phi_{ki}$ be the embedding of the $i$th top word in the $k$th estimated topic.
The Embedded TC is defined as $\frac{1}{K} \sum_{k=1}^K \sum_{1\leq i \neq j\leq m} \frac{\phi_{ki}^\top \phi_{kj}}{|\phi_{ki}| |\phi_{kj}|}$, measuring the semantic closeness of top words within topics, and the Embedded TD is $\frac{1}{K(K-1)} \sum_{k \neq \ell} \left| \frac{1}{m} \sum_{i=1}^m \phi_{ki} - \frac{1}{m} \sum_{j=1}^m \phi_{\ell j} \right|$, measuring the distinction of semantic meanings across topics. 
The results for TRACE and two other methods, ECRTM \citep{wu2023effective} and FASTopic \citep{wu2024fastopic}, are summarized in Table~\ref{tab:quantitative-comparison-merged-ap-sa}. 
Since $K$ is also unknown in real data, we also investigate different values of $K$ in this experiment. It shows that overall TRACE has the best topic coherence.  
For topic diversity, FASTopic is the best under the original TD criteria, and TRACE is the best under the Embedded TD criteria.

{\bf Remark 4}: These metrics are heuristic criteria designed to quantify what is a ``good" output. As a result, evaluation can be sensitive to how the metrics are defined, and in some cases, certain metrics may introduce bias. For example, both TD and Embedded TD penalize overlap of top words across topics. However, as illustrated in Figure~\ref{fig:AP-context}, the same word may be relevant to multiple topics depending on context, causing it to appear in the top word lists of several topics and leading to a low topic diversity score. This behavior is undesirable for context-aware topic definitions. For these reasons, we interpret these metrics with caution when evaluating performance.


\spacingset{1}
\begin{table}[t]
\centering
\caption{\small Topic coherence and diversity under various choices of \(K\) on AP and MADStat datasets. The mean and standard error (in brackets) on 20 independent runs are reported.}
\label{tab:quantitative-comparison-merged-ap-sa}
\renewcommand{\arraystretch}{0.9}
\setlength{\tabcolsep}{5pt}
\scalebox{.55}{
\begin{tabular}{l l
                c c c c
                c c c c}
\toprule
& & \multicolumn{4}{c}{AP} & \multicolumn{4}{c}{MADStat} \\
\cmidrule(lr){3-6} \cmidrule(lr){7-10}
Metric & Method
& \(K=5\) & \(K=10\) & \(K=20\) & \(K=50\)
& \(K=10\) & \(K=20\) & \(K=50\) & \(K=100\) \\
\midrule
\multirow{3}{*}{TC}
  & ECRTM
  & 0.477 ($\pm$0.009) & 0.486 ($\pm$0.003) & 0.552 ($\pm$0.001) & \textbf{0.630 ($\pm$0.002)}
  & 0.433 ($\pm$0.002) & 0.436 ($\pm$0.002) & 0.450 ($\pm$0.001) & 0.444 ($\pm$0.001) \\
& FASTopic
  & 0.411 ($\pm$0.004) & 0.483 ($\pm$0.006) & 0.509 ($\pm$0.004) & 0.532 ($\pm$0.002)
  & 0.480 ($\pm$0.006) & 0.502 ($\pm$0.004) & 0.508 ($\pm$0.002) & 0.499 ($\pm$0.001) \\
& TRACE
  & \textbf{0.528 ($\pm$0.004)} & \textbf{0.553 ($\pm$0.005)} & \textbf{0.559 ($\pm$0.002)} & 0.595 ($\pm$0.002)
  & \textbf{0.539 ($\pm$0.003)} & \textbf{0.541 ($\pm$0.002)} & \textbf{0.551 ($\pm$0.001)} & \textbf{0.561 ($\pm$0.001)} \\
\midrule
\multirow{3}{*}{Embedded TC}
  & ECRTM
  & 0.455 ($\pm$0.007) & 0.470 ($\pm$0.002) & 0.423 ($\pm$0.001) & 0.412 ($\pm$0.001)
  & 0.359 ($\pm$0.001) & 0.470 ($\pm$0.002) & 0.423 ($\pm$0.001) & 0.363 ($\pm$0.001) \\
& FASTopic
  & \textbf{0.561 ($\pm$0.001)} & \textbf{0.553 ($\pm$0.002)} & 0.530 ($\pm$0.002) & 0.535 ($\pm$0.001)
  & \textbf{0.461 ($\pm$0.002)} & 0.437 ($\pm$0.001) & 0.455 ($\pm$0.001) & 0.480 ($\pm$0.001) \\
& TRACE
  & 0.522 ($\pm$0.004) & 0.528 ($\pm$0.003) & \textbf{0.540 ($\pm$0.002)} & \textbf{0.540 ($\pm$0.001)
  }& 0.456 ($\pm$0.003) & \textbf{0.469 ($\pm$0.002)} & \textbf{0.479 ($\pm$0.001)} & \textbf{0.497 ($\pm$0.001)} \\
\midrule
\multirow{3}{*}{TD}
  & ECRTM
  & 0.992 ($\pm$0.002) & 0.985 ($\pm$0.002) & 0.920 ($\pm$0.003) & 0.713 ($\pm$0.002)
  & 0.862 ($\pm$0.002) & 0.721 ($\pm$0.002) & 0.466 ($\pm$0.002) & 0.289 ($\pm$0.002) \\
& FASTopic
  & \textbf{1.000 ($\pm$0.000)} & \textbf{1.000 ($\pm$0.000)} & \textbf{0.997 ($\pm$0.001)} & \textbf{0.933 ($\pm$0.001)}
  & \textbf{1.000 ($\pm$0.000)} & \textbf{0.998 ($\pm$0.001)} & \textbf{0.930 ($\pm$0.001)} & \textbf{0.754 ($\pm$0.001)} \\
& TRACE
  & 0.988 ($\pm$0.001) & 0.964 ($\pm$0.002) & 0.924 ($\pm$0.003) & 0.789 ($\pm$0.005)
  & 0.973 ($\pm$0.002) & 0.917 ($\pm$0.002) & 0.782 ($\pm$0.001) & 0.619 ($\pm$0.003) \\
\midrule
\multirow{3}{*}{Embedded TD}
  & ECRTM 
  & 3.93 ($\pm$0.06) & 4.12 ($\pm$0.03) & 3.99 ($\pm$0.02) & 3.57 ($\pm$0.01)
  & 2.77 ($\pm$0.02) & 2.71 ($\pm$0.02) & 2.68 ($\pm$0.01) & 2.66 ($\pm$0.01) \\
& FASTopic
  & 3.97 ($\pm$0.04) & 3.97 ($\pm$0.02) & 3.99 ($\pm$0.02) & 3.79 ($\pm$0.01)
  & 3.83 ($\pm$0.06) & 3.83 ($\pm$0.03) & {\bf 3.94 ($\pm$0.01)} & 3.65 ($\pm$0.01) \\
& TRACE
  & \textbf{4.44 ($\pm$0.08)} & \textbf{4.23 ($\pm$0.05)} & \textbf{4.52 ($\pm$0.05)} & \textbf{4.42 ($\pm$0.04)}
  & \textbf{4.04 ($\pm$0.07)} & \textbf{4.20 ($\pm$0.04)} & \textbf{3.94 ($\pm$0.03)} & \textbf{3.90 ($\pm$0.04)} \\
\bottomrule
\end{tabular}
}
\end{table}
\resetspacing

\spacingset{1}
\begin{table}[tb]
\setlength{\tabcolsep}{8pt}
\centering
\caption{\small Document clustering performance on MADStat. Results are evaluated from $20$ independent runs.} \label{tab:clustering}
\scalebox{.65}{
\begin{tabular}{l c c c c c c}
\toprule
& \multicolumn{3}{c}{Journal Clustering (10 clusters)} & \multicolumn{3}{c}{Pairwise Author Clustering} \\
\cmidrule(lr){2-4} \cmidrule(lr){5-7}
Method &  ECRTM & FASTopic & TRACE &  ECRTM & FASTopic & TRACE \\
\midrule
Accuracy & 0.256 (0.004) & 0.254 (0.004) & \textbf{0.275 (0.003)} & 0.589 (0.002)  & 0.600 (0.003) & \textbf{0.617 (0.003)}\\
\bottomrule
\end{tabular}}
\end{table}
\resetspacing

The output of topic modeling is frequently used in downstream tasks such as document clustering. The MADStat data set has multiple document-level attributes, such as journal and authors. We use these attributes as true cluster labels and evaluate the accuracy of applying k-means clustering to the estimated topic weight vectors $\hat{w}_i$ from different methods. We consider two settings. In the first setting, we take all abstracts from the 10 journals that have the highest impact factors and run k-means clustering on the $\hat{w}_i$'s assuming 10 clusters. The accuracy (up to a cluster permutation) is summarized in Table~\ref{tab:clustering}. In this scenario, a random guess has an accuracy of 0.1, so all methods significantly outperform the random guess. 
However, since these journals tend to have overlapping topic preference, the accuracy is relatively low, with TRACE being the best among the three. In the second setting, we focus on the top 15 authors who have the most papers in this data set. For each pair of authors, we take all of their text abstracts and run k-means clustering on the $\hat{w}_i$'s assuming two clusters. We then average the clustering accuracy over ${15\choose 2}=105$ author pairs. The results are also shown in Table~\ref{tab:clustering}. Unlike journal clustering, author clustering has higher accuracies, with an average of approximately $60\%$ over all author pairs, and around $80\%$ for some particular author pairs. Among the three methods, TRACE improves ECRTM and slightly improves FASTopic.

\vspace*{-.5cm}
\section{Discussion} \label{sec:Discuss}
\vspace*{-5pt}


We propose a framework for integrating contextualized word embeddings into topic modeling. Unlike traditional topic modeling approaches, our method captures semantic meaning and contextual information of words. Compared to other 
word-embedding-based topic modeling methods, our approach offers two 
benefits: (i) it provides flexibility by allowing the integration of any traditional topic modeling algorithm without modification, and (ii) it comes with a theoretical guarantee. 

The net-rounding step in our method reduces the problem to a traditional topic model with a vocabulary size of $M$. 
This can be viewed as a 
regularization, and our simulation results indeed show that the performance of our method improves when $M$ is chosen wisely and that as $M$ varies we observe a bias-variance trade-off. It is worth noting that this reduction in vocabulary size is only feasible due to the use of word embeddings. It would not be possible in the classical word-count setting, as there is no clear way to create $M$ representative hyperwords. Meanwhile, reducing $M$ inevitably yields information loss. In our method, this loss is  mitigated by the use of contextual embeddings, where the same word can correspond to multiple hyperwords depending on its context. 

The use of contextual embeddings significantly increases the computational cost: Embeddings 
from LLMs typically have dimensions in a few hundreds, and each word can correspond to multiple embeddings. To address this challenge, we propose solutions such as reducing the dimensionality of the embeddings and binning them into hyperwords. However, this approach is still computationally demanding, especially when the document length increases. A potential solution is to use sentence embeddings. By working with sentence embeddings, we could reduce the number of units to process in each document (a document has much 
less sentences than words). Leveraging the property that similar sentences are represented by similar embeddings, we may partition the sentence embedding space to create ``hyper-sentences''. Assuming the existence of anchor region in the sentence embedding space, we may adapt the traditional topic modeling algorithm to 
hyper-sentence counts.

The PPTM framework may also be extended to joint analysis of text data and other data. One example is a combination with network data analysis, especially networks with nodal covariates \citep{huang2024pcabm, zhang2022joint, wang2024variable}. For example, in coauthorsip and citation analysis, text abstracts are often used as nodal covariates. We may construct a joint model for network and text, assuming that the community membership of a node in the network and its topic weight vector $w_i$ are correlated.

%

\clearpage

 \makeatletter
\renewcommand{\section}{\@startsection{section}{1}{0pt}%
  {\baselineskip}{0.5\baselineskip}{\large\bfseries}}
\renewcommand{\subsection}{\@startsection{subsection}{2}{0pt}%
  {0.5\baselineskip}{0.3\baselineskip}{\normalsize\bfseries}}
\makeatother

\allowdisplaybreaks

\renewcommand\thetable{\thesection.\arabic{table}}
\renewcommand\thefigure{\thesection.\arabic{figure}}

\spacingset{1.72} 

\addcontentsline{toc}{section}{Appendix} 
\part{Appendix} 
\parttoc 

\appendix

\section{Self-Attention, Transformer, and Proof of Lemma~\ref{lem:transformer}}\label{supp:transformers}

Transformer 
\citep{vaswani2017attention} is the foundational architecture of modern large language models (LLMs), and self-attention is a building block of the transformer. 
In this section, we describe the self-attention mechanism and the transformer architecture using explicit mathematical forms, and then prove the permutation invariance property stated in Lemma~\ref{lem:transformer}. 

\begin{figure}[tb!]
\centering
\begin{subfigure}[b]{.65\textwidth}
\includegraphics[width=1\textwidth]{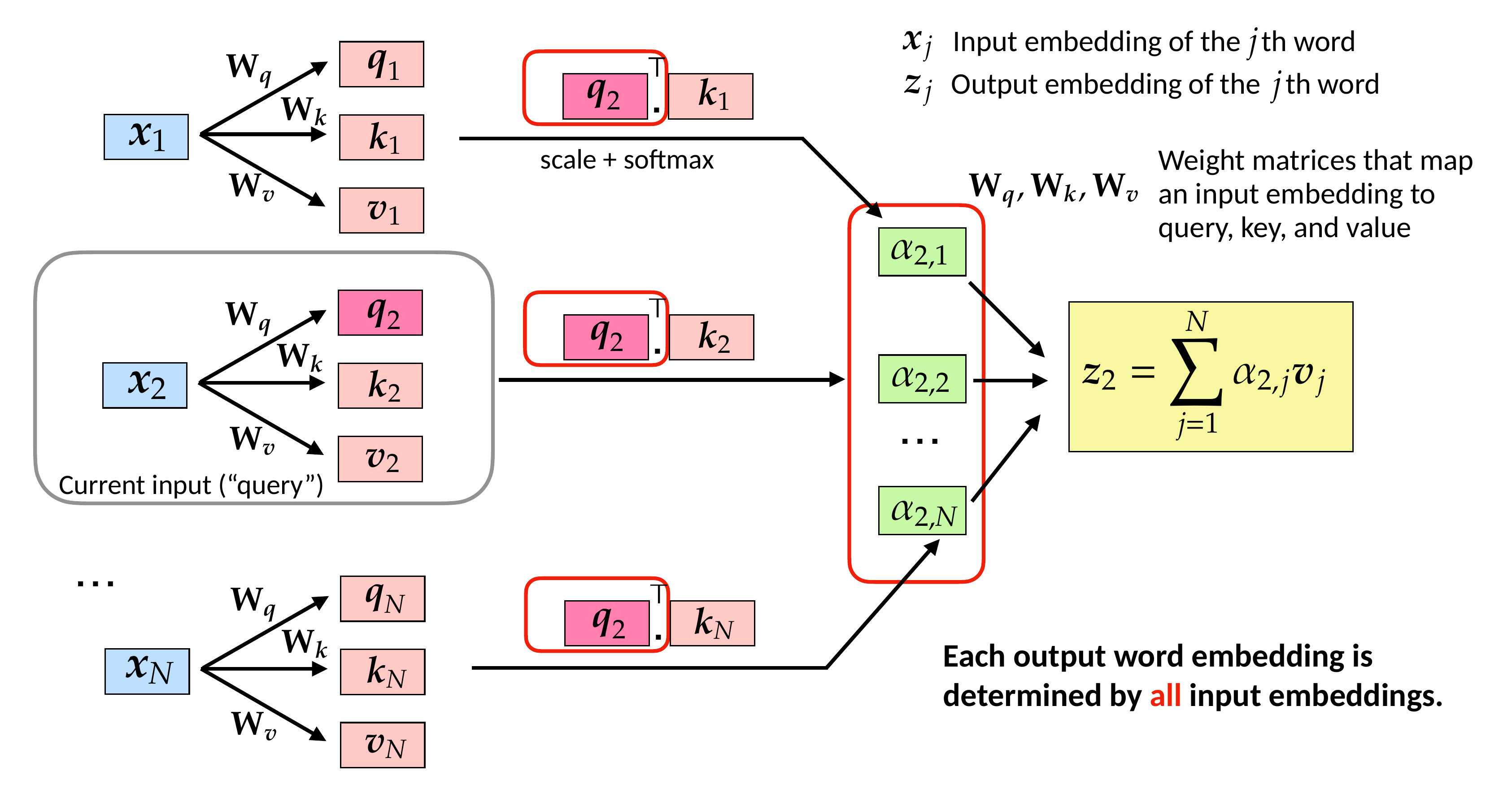} 
\caption{\small An illustration of a self-attention layer for creating contextualized word embeddings.} \label{fig:attention-supp}
\end{subfigure}
\hspace{5pt}
\begin{subfigure}[b]{.3\textwidth}
\includegraphics[width=.65\textwidth]{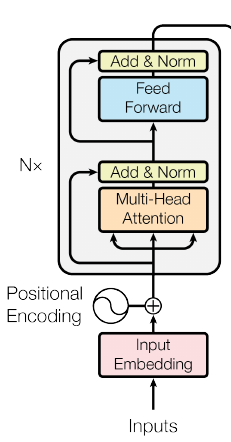}
\caption{\small The structure of an encoder transformer \citep{vaswani2017attention}.} \label{fig:transformer}
\end{subfigure}
\caption{}
\end{figure}

{\bf Self-attention}: Self-attention is a special class of sequence-to-sequence mappings. An example is illustrated in  
Figure~\ref{fig:attention-supp}. 
Let \( x_1, x_2, \ldots, x_N \in \mathbb{R}^{d_0} \) be an input sequence, where $x_j$ represents the input embedding of the $j$th word.  Each $x_j$ is first mapped into three vectors:
\beq \label{attention-1}
q_j = W_q x_j \in \mathbb{R}^{d_1}, \qquad  k_j = W_k x_j \in \mathbb{R}^{d_1}, \qquad v_j = W_v x_j \in \mathbb{R}^{d_2}.
\eeq
where \( W_q \in\mathbb{R}^{d_1\times d_0}\), \( W_k\in\mathbb{R}^{d_1\times d_0} \), and \( W_v \in\mathbb{R}^{d_2\times d_0}\) are parameters learnable in pre-training. The output is a new sequence $z_1, z_2,\ldots,z_N$ defined by
\beq \label{attention-2}
z_j =  \sum_{m=1}^N \alpha_{j,m} v_m, \qquad \mbox{where}\quad \alpha_{j,m}=\operatorname{Softmax}\left(q_j' k_m / \sqrt{d_1}\right). 
\eeq
Here, $q_j$, $k_j$, and $v_j$ are called the ``query", ``key", and ``value" at the $j$th token, respectively, and $\alpha_{j,m}$ is called the attention weight. Intuitively, the query encodes the contextual information each word is looking for, while the key represents the information each word can provide. When the key \( k_m \) aligns closely with the query \( q_j \), the \( m \)-th word is deemed more relevant to the context of the \( j \)-th word, so that  \( z_j \) places more ``attention" on the $m$-th word.

\spacingset{1}
\resetspacing

Multi-head attention refers to concatenating the outputs of multiple self-attention blocks. 
In detail, let $W_q^{(h)}, W_k^{(h)}, W_v^{(h)}$ be the parameter matrices of the $h$-th self-attention mapping, for $1\leq h\leq H$, and let $P$ be a $d^* \times d$ matrix, where $d$ is the output dimension, and $d^*=\sum_{h=1}^H d_h$, with $d_h$ being the dimension of the output in the $h$-th self-attention mapping. The output is
\beq \label{multi-head}
z_j^{\text{multi-head}} = P\begin{bmatrix}
\sum_{m=1}^N \operatorname{Softmax}\bigl(\frac{x_j' (W_q^{(1)})' (W_k^{(1)})'x_m}{ \sqrt{d_1}}\bigr) W_v^{(1)}x_m\\
\sum_{m=1}^N \operatorname{Softmax}\bigl(\frac{x_j' (W_q^{(2)})' (W_k^{(2)})'x_m}{ \sqrt{d_2}}\bigr) W_v^{(2)}x_m\\
\vdots\\
\sum_{m=1}^N \operatorname{Softmax}\bigl(\frac{x_j' (W_q^{(H)})' (W_k^{(H)})'x_m}{ \sqrt{d_H}}\bigr) W_v^{(H)}x_m
\end{bmatrix}, \qquad 1\leq j\leq N. 
\eeq


{\bf Transformer}: A standard transformer \citep{vaswani2017attention} contains an encoder and a decoder. 
For creating word embeddings (e.g., using BERT), only the encoder part is used. The encoder transformer is a sequence-to-sequence mapping, which takes as input a word (or token) sequence $v_1, v_2,\ldots, v_N$, where each $v_j$ takes values in the vocabulary ${\cal V}$. 

As illustrated in Figure~\ref{fig:transformer}, 
the first two layers are the Input Embedding layer and the Position Encoding layer. 
The Input Embedding layer creates $z_j^{\text{input}}=f(v_j)$, where $f: {\cal V}\to\mathbb{R}^{d_0}$ is a fixed mapping. The Positional Encoding layer creates $z_j^{\text{position}}\in\mathbb{R}^{d_0}$, where this vector is sorely determined by $(j, N, d_0)$ and doesn't depend on the value of $v_j$. The two embeddings are summed together:
\beq \label{input-embedding}
z^{(0)}_j := z_j^{\text{input}} +  z_j^{\text{position}} \;\;\in\;\; \mathbb{R}^{d_0}, \qquad \mbox{for $1\leq j\leq N$}. 
\eeq 

Next, these embeddings are updated in multiple attention layers. Each attention layer consists of a Multi-Head Attention sub-layer and a Feed Forward sub-layer. 
The Multi-Head attention mapping is as described in \eqref{multi-head}, and the Feed Forward mapping is a vector-to-vector mapping $g(\cdot)$ applied separately to each vector in the sequence. 

Finally, Residual Connection is used to connect two sub-layers within a multiple attention layer, as well as connecting different layers. Let $x_1, x_2,\ldots, x_N$ be the input of a sub-layer (or layer) ${\cal F}$, and let $z_{1:N}={\cal F}(x_{1:N})$ be the output sequence. Before $z_1, z_2,\ldots,z_N$ are sent to the next sub-layer (or layer), we apply a transformation as follows:
\beq \label{residual-connection}
z_j^* = \frac{1}{\sigma_j}(y_j - \mu_j {\bf 1}_d), \qquad\mbox{where}\quad y_j = x_j + z_j, \quad \mu_j = \frac{y_j'{\bf 1}_d}{d}, \quad \sigma_j = \frac{\|y_j - \mu_j{\bf 1}_d\|}{\sqrt{d}}.
\eeq 
In \eqref{residual-connection}, the input and output of ${\cal F}$ is first summed to obtain $y_j$, and then $y_j$ is normalized by a  layer normalization (a normalization with respect to features rather than samples). 

{\bf Proof of Lemma~\ref{lem:transformer}}: Once we write down the above mathematical forms, Lemma~\ref{lem:transformer} follows immediately: The Feed Forward layer and Residual Connection are {\it position-wise} mappings (i.e., the output at the $j$th position only depends on the input at the $j$th position), so they must satisfy the permutation invariance. The Multi-Head attention layer in \eqref{multi-head} is permutation-invariant because the attention weight between two positions $j$ and $m$ only depends on the input embedding values $x_j$ and $x_m$ but not the indices $(j, m)$.

{\bf Remark}: In modern LLMs, additional structures, such as masking layers, may be added into the transformer. In such cases, the permutation invariance property does not hold exactly. However, this property is only used to motivate our point process modeling rather than serving as a key assumption. Our main point is that, compared to recurrent neural networks (RNNs), the transformer architecture substantially weakens the role of input ordering of words, making it reasonable to model the output as a ``bag of vectors" rather than an ordered sequence.

\section{Detailed Steps of the TRACE Algorithm} \label{supp:Alg}

\subsection{The UMAP Projection of Word Embeddings}\label{supp:umap}

\textit{UMAP (Uniform Manifold Approximation and Projection)} is a widely used nonlinear dimension reduction technique designed for high-dimensional data. It is particularly effective for visualizing and analyzing data by projecting it into a lower-dimensional space while preserving both local and global manifold structures. Developed by \cite{mcinnes2018umap}, UMAP is based on mathematical principles of Riemannian geometry and algebraic topology, using a k-nearest neighbors (k-NN) graph to approximate the data’s manifold structure. In our implementation, the \texttt{umap-learn} Python library is adopted to fit a UMAP model. Three key parameters control its behavior:
\vspace*{-7pt}
\begin{itemize}
\setlength{\itemsep}{-6pt}
    {\item \textit{Number of reduced dimensions $d$}: Specifies the target dimensionality of the embedding space. This hyperparameter controls how much the data is compressed and directly affects the resolution at which structures in the data can be represented. Common choices are in the range of 2 to 50, depending on the complexity of the data and the downstream task. The detail discussion on the choice of $d$ can be find in Appendix~\ref{supp:umap-d}.}
    \item \textit{Number of Neighbors} (\texttt{n\_neighbors}): Determines the balance between local and global structure. Smaller values (e.g., 5--15) focus on preserving local relationships, while larger values (e.g., 50--100) emphasize global structure.
    \item \textit{Minimum Distance} (\texttt{min\_dist}): Governs the spacing between points in the embedding. Smaller values (e.g., 0.01) form tight clusters, while larger values (e.g., 0.5) create smoother transitions.
\end{itemize}
\vspace*{-7pt}

To address computational constraints, we train UMAP on a randomly selected subset of the data and apply the learned projection to transform all data points. A smaller subset, however, may increase the approximation error of the underlying manifold, lead to the loss of local patterns, or result in instability in the embedding. To assess the quality of the transformation, we visualize both the training points and the remaining points on the same plot after UMAP projection to ensure that external points are adequately represented by the training subset. Quantitatively, the ``trustworthiness'' score \citep{venna2001neighborhood,van2009learning} can be used to evaluate how well local structures are preserved, with a score above \( 0.5 \) recommended to avoid significant distortion.

{Additionally, we would like to discuss the influence of our model assumptions when applying UMAP. If this transformation is fixed or learned from external data, then applying it to a Poisson point process preserves the Poisson property in the transformed space. In our algorithm, however, for practical reasons, we learn the UMAP transformation using a small subset of the embeddings (e.g., 10\%-20\%) from the same dataset. As a result, the transformed points ${z_{ij}}$ no longer follow a perfect Poisson point process, since the transformation depends partially on the data it is applied to. That said, the deviation is minor when the proportion of embeddings used to learn the transformation is small or when the total number of embeddings is large. In the latter case, the learned transformation is expected to approximate a fixed map closely. If the dimension reduction method satisfies a consistency property, i.e., it converges to a deterministic transformation as the data size increases, then the transformed embeddings can be reasonably approximated by a Poisson point process. This approximation becomes increasingly accurate with larger datasets, which supports the use of the Poisson assumption in our theoretical analysis.}

\subsection{The Topic-SCORE algorithm}\label{supp:topicscore}
The \textit{Topic-SCORE algorithm} \citep{ke2022using} is a spectral method for topic modeling that leverages the geometric structure of word count matrices to efficiently identify and recover topics. Building on recent advances in text analysis, the Topic-SCORE algorithm offers a computationally efficient solution with strong theoretical guarantees, including provable convergence rates and robustness to noise. The integration of insights from entry-wise eigenvector analysis enhances the precision of spectral decomposition during SVD \citep{ke2024entry}, making the algorithm particularly well-suited for large datasets and short documents, as highlighted in \citep{ke2023recent}.

Specifically, let the word-topic matrix be \(A = [A_1, \ldots, A_K] \in \mathbb{R}^{p \times K}\) and the topic-document weight matrix be \(W = [w_1, \ldots, w_n] \in \mathbb{R}^{K \times n}\). Under the assumption of the traditional topic model \eqref{traditional-TM}, we have
\(
\mathbb{E}[X \operatorname{diag}(\mathbf{1}_p X)^{-1}] = AW
\),
where \(X\) is the word count matrix. The algorithm consists of the following five steps:

\vspace*{7pt}
\textit{Pre-SVD Normalization}: Normalize the word-document matrix to account for variations in document length and word frequencies. Two common approaches are:
\vspace*{-7pt}
\begin{itemize}
\setlength{\itemsep}{-6pt}
\item Using the word frequency matrix \(\widetilde{X} = X \operatorname{diag}(\mathbf{1}_{p} X)^{-1}\) instead of the word count matrix.
\item Normalizing the rows of the word count matrix as \(\widetilde{X} = \operatorname{diag}(n^{-1} X \mathbf{1}_{n})^{-1/2} X\).
\end{itemize}
\vspace*{-7pt}
We implement both approaches but default to the first approach for simplicity.

\textit{Singular Value Decomposition (SVD)}: Apply SVD to project the high-dimensional row space of \(\widetilde{X}\) into a \(K\)-dimensional subspace while preserving its simplicial cone structure. Specifically, this corresponds to the row space of the matrix \([\xi_{1}, \ldots, \xi_{K}]\), where \(\xi_{1}, \ldots, \xi_{K}\) are the first \(K\) left singular vectors of \(\widetilde{X}\).

\textit{SCORE Normalization}: The simplicial cone is transformed into a cross-sectional simplex using the SCORE normalization \citep{SCOREreview}. Define a matrix \(R \in \mathbb{R}^{p \times (K-1)}\), where \(R(j, k) = \xi_{k+1}(j)/\xi_1(j)\) for \(1 \leq j \leq p\) and \(1 \leq k \leq K-1\). Notably, by Perron's theorem, \(\xi_1\) is a strictly positive vector under certain conditions, ensuring that \(R\) is well-defined. The simplex structure in \(R = [r_{1}, \ldots, r_{p}]'\) is described as follows: All rows \(r_{1}, \ldots, r_{p}\) are contained in a low-dimensional simplex \({\cal S} \subset \mathbb{R}^{K-1}\), where \({\cal S}\) is spanned by \(K\) vertices \(v_1, v_2, \ldots, v_K\). If word \(j\) is an anchor word of topic \(k\) (i.e., the \(j\)-th row of \(A\) has only one nonzero entry located at \(k\)), then the \(j\)-th row of \(R\) lies exactly on the vertex \(v_k\).

\textit{Vertex Hunting}: Perform the vertex hunting step to identify the vertices \(v_{1}, \ldots, v_{K}\) of the convex hull formed by the rows \(r_{1}, \ldots, r_{p}\) of \(R\), corresponding to anchor words uniquely associated with each topic. Several algorithms can be used for this step, including the Successive Projections Algorithm (SPA) \citep{araujo2001successive}, N-FINDR \citep{winter1999n}, and Sketched Vertex Search (SVS) \citep{Mixed-SCORE}. Each of these algorithms delivers desirable performance with appropriate parameter tuning and outlier removal. In our implementation, we default to SPA due to its computational simplicity and practical effectiveness, while also providing implementations of N-FINDR and SVS for users seeking alternative approaches. 

\textit{Topic Matrix Estimation}: Finally, the topic matrix \(A = [A_1, \ldots, A_K]\) is reconstructed using the identified anchor words and the simplex geometry defined by their positions. Specifically, after obtaining \(v_1, v_2, \ldots, v_K\), the barycentric coordinate for each row of \(R\) is calculated. (Each point in \({\cal S}\) is expressed as a convex combination of the \(K\) vertices, and the vector of convex combination coefficients is referred to as the barycentric coordinate.) 
Denote by \(\pi_j \in \mathbb{R}^K\) the barycentric coordinate of the \(j\)-th row of \(R\), and write \(\Pi = [\pi_1, \pi_2, \ldots, \pi_p]' \in \mathbb{R}^{p \times K}\). \cite{ke2022using} established a connection between \(\Pi\) and the true topic vectors \(A_k \propto [\diag(\xi_1)]^{-1} \Pi e_k\) for \( 1 \leq k \leq K\). Hence, the topic vectors can be recovered from \(\Pi\).

As a final note, the original paper by \cite{ke2022using} provided an implementation of Topic-SCORE in R. In this work, we reimplement the algorithm in Python to improve accessibility and facilitate integration with modern data science workflows.

\section{Proofs of Theoretical Results} \label{supp:Proofs}

\subsection{Proof of Lemma~\ref{lem:induced-model}}
Fix $1\leq i\leq n$. We recall that \( {\cal Z} = \bigsqcup_{m=1}^M \calR_m \) is a partition of ${\cal Z}$ into disjoint regions, $X_i^{\text{net}}(m)$ represents the number of embeddings falling into ${\cal R}_m$, and the embeddings follow a Poisson point process with an intensity measure $N\Omega_i(\cdot)$. The definition of Poisson point processes \citep{last2017lectures} implies:
\beq \label{lem-induceTM-0}
\bigl\{X_i^{\text{net}}(m)\bigr\}_{m=1}^M \mbox{ are independent, $X_i^{\text{net}}(m)\sim \mathrm{Poisson}(\lambda_{im})$, and $\lambda_{im}= N\int_{z\in {\cal R}_m}\Omega_i(z)dz$}. 
\eeq
The next lemma presents a well-known connection between Poisson and multinomial  distributions, whose proof is elementary and omitted: 
\begin{lemma}\label{lem:poisson-to-multinomial}
    Suppose $Y=(Y_1, Y_2, \ldots, Y_M)'$ contain independent variables, with $Y_m\sim \mathrm{Poisson}(\lambda_m)$.   Let $q\in\mathbb{R}^M$ with $q_m=\frac{\lambda_m}{\sum_{j=1}^M \lambda_{j}}$. Then,  $Y|(Y_1+\ldots + Y_M=n) \sim \mathrm{Multinomial}(n, q)$, for all $n\geq 1$.  
\end{lemma}

\noindent
We apply Lemma~\ref{lem:poisson-to-multinomial} to $X_i^{\text{net}}$ and notice that  $N_i=\sum_{m=1}^M X_i^{\text{net}}(m)$. It yields:
\begin{equation} \label{lem-induceTM-1}
X_i^{\text{net}}\mid N_i \sim \mathrm{Multinomial}(N_i, \Omega_i^{\text{net}}), \quad\mbox{where}\;\; \Omega_i^{\text{net}}(m):=\frac{\lambda_{im}}{\sum_{m'=1}^M \lambda_{im'}}, \;\; \mbox{for $1\leq m'\leq M$}. 
\end{equation}
Comparing \eqref{lem-induceTM-1} with the desirable claim, it suffices to show that $\Omega_i^{\text{net}}$ satisfies
\beq \label{lem-induceTM-2}
\Omega_i^{\text{net}}=\sum_{k=1}^K w_i(k) A_k^{\text{net}}.
\eeq

We now show \eqref{lem-induceTM-2}. 
By our model in \eqref{PTM-1}, $\Omega_i(z)=\sum_{k=1}^K w_i(k){\cal A}_k(z)$. As a result, 
\[
\lambda_{im} = N\int_{z\in{\cal R}_m}\biggl[\sum_{k=1}^K w_i(k){\cal A}_k(z)\biggr]dz=N\sum_{k=1}^K w_i(k)A_k^{\text{net}}(m).  
\]
Hence, $\sum_{m'=1}^M \lambda_{im'}=N\sum_{k=1}^K w_i(k)\sum_{m'=1}^M A_k^{\text{net}}(m')=N$ (the last equality is due to that $A_k^{\text{net}}$ and $w_i$ both have unit entry sums). In the definition of $\Omega_i^{\text{net}}(m)$ in \eqref{lem-induceTM-1}, we substitute $\lambda_{im}$ and apply the equality of $\sum_{m'=1}^M \lambda_{im'}=N$. 
It follows that 
\begin{equation} \label{lem-induceTM-3}
\Omega_i^{\text{net}}(m)= \frac{N\sum_{k=1}^K w_i(k)A_k^{\text{net}}(m)}{N}=\sum_{k=1}^K w_i(k)A_k^{\text{net}}(m), \qquad\mbox{for $1\leq m\leq M$}.
\end{equation}
This proves \eqref{lem-induceTM-2}, and the claim follows immediately. \qed

\subsection{Proof of Lemma~\ref{lem:identifiability}}

We will prove the identifiability of PPTM by using the identifiability of the reduced models on hyperword counts for a sequence of nets. For each $\epsilon\in (0,1)$, consider a special net where each ${\cal R}_m$ is a hypercube with side length $\epsilon$. We consider a sequence: $\epsilon_s=1/s$, for $s=1,2,3,\ldots$. We use the notations $X^{\text{net},s}$ and $A_k^{\text{net},s}$ to indicate their dependence on $\epsilon_s$. 

By \cref{lem:induced-model}, the hyperword count matrix $X^{\text{net},s}$ follows a traditional topic model with parameters $A^{\text{net},s}=[A_1^{\text{net},s},\ldots, A_K^{\text{net},s}]$ and $W=[w_1,\ldots,w_n]$. It is known \citep{donoho2003does,arora2013practical} that these parameters are identifiable if the following requirements are satisfied: 
\begin{itemize}
    \item[(i)] Each $A_k^{\text{net},s}$ has at least one anchor hyperword.
    \item[(ii)] Both $A^{\text{net},s}$ and $W$ have a rank $K$. 
\end{itemize}

We first check (i). The $m$th hyperword is an anchor hyperword of topic $k$ if and only if  $A_k^{\text{net}}(m)\neq 0$ and $\sum_{\ell:\ell\neq k}A_\ell^{\text{net}}(m)=0$. Comparing it with the expression of $A_k^{\text{net}}(m)$ in \cref{lem:identifiability}, this requires:
\beq \label{lem-identifiability-1}
\int_{{\cal R}_m}{\cal A}_k(z)dz\neq 0, \qquad \mbox{and}\qquad \int_{{\cal R}_m}\sum_{\ell: \ell\neq k}{\cal A}_k(z)dz=0. 
\eeq
We notice that ${\cal A}_k(z) + \sum_{\ell:\ell\neq k}{\cal A}_\ell (z)=h(z)$, and $\int_{\calR_m} h(z)dz>0$ (because by our assumption, $h(z)$ is uniformly lower bounded by a constant in ${\cal Z}$). Consequently, if the second argument in \eqref{lem-identifiability-1} holds, then $\int_{\calR_m}{\cal A}_k(z)dz=\int_{\calR_m}h(z)dz>0$, which implies that the first argument holds. Therefore, we only need to check whether the first argument in \eqref{lem-identifiability-1} holds. This argument is equivalent to: 
\beq \label{lem-identifiability-2}
\calR_m\subset {\cal S}_k, \qquad\mbox{where ${\cal S}_k$ is the anchor region defined in \eqref{def-anchor-region}}. 
\eeq
By our assumption, the volume of ${\cal S}_k$ is lower bounded by a constant. Additionally, ${\cal A}_1(\cdot),\ldots,{\cal A}_K(\cdot)$ are all fixed densities that satisfy the $\beta$-Holder smoothness assumption. Hence, ${\cal S}_k$ must have a connected subset that contains a hypercube whose side length is $c'$, for a properly small constant $c'>0$. 
It follows that when $\epsilon_s$ is sufficiently small (i.e., $s$ is sufficiently large), there must exist a hypercube $\calR_m$ that is completely contained in ${\cal S}_k$. This proves \eqref{lem-identifiability-2} and verifies requirement (i). 

We then check (ii). The rank of $W$ follows immediately from our assumption on $\Sigma_W$. Let $h_m^{\textup{net}}=\sum_{k=1}^K A_k^{\textup{net}}(m)$,  and $H^{\textup{net}}=\diag(h_1^{\textup{net}}, h_2^{\textup{net}}, \ldots, h_M^{\textup{net}})$. Define a matrix $ \Sigma^{\textup{net}}_A = (A^{\textup{net}})'\bigl(H^{\textup{net}}\bigr)^{-1}A^{\textup{net}}$ (we have omitted the superscript $s$ in all places). 
This matrix will be studied carefully in \cref{lem:SigmaA}. Using the results there, $\|\Sigma_A^{\text{net}}-\Sigma_{\cal A}\|\leq C\epsilon_s^{\beta\wedge 1}$, 
where $\Sigma_{{\cal A}}$ is as defined in \eqref{def-h-SigmaA}. By Weyl's inequality, $\lambda_{\min}(\Sigma_A^{\text{net}})\geq \lambda_{\min}(\Sigma_{\cal A})-\|\Sigma_A^{\text{net}}-\Sigma_{\cal A}\|$; and by our assumption, $\lambda_{\min}(\Sigma_{{\cal A}})$ is lower bounded by a constant. Therefore, as long as $\epsilon_s$ is sufficiently small, $\lambda_{\min}(\Sigma_A^{\text{net}})$ must be nonzero. This has verified requirement (ii).

So far, we have shown that for this sequence of nets indexed by $s$, for all large enough $s$, requirements (i)-(ii) are satisfied, so $A^{\text{net},s}$ and $W$ are identifiable. 
The identifiability of $W$ follows directly. To obtain the identifiability of ${\cal A}_k(\cdot)$, suppose there exist $\widetilde{\cal A}_1(\cdot), \ldots, \widetilde{\cal A}_k(\cdot)$ that are distinct from ${\cal A}_1(\cdot), \ldots, {\cal A}_K(\cdot)$ but lead to the same distribution on data. As a result, the resultant $\widetilde{A}_k^{\text{net},s}$ must be equal to $A_k^{\text{net},s}$ for all sufficiently large $s$. It implies:
\beq \label{lem-identifiability-3}
\mbox{The integrals of ${\cal A}_k(\cdot)$ and $\widetilde{\cal A}_k(\cdot)$ are the same in all sufficiently small hypercubes}. 
\eeq
Combining \eqref{lem-identifiability-3} with the smoothness assumption, the two densities must be the same. Hence, ${\cal A}_k(\cdot)$ is identifiable. \qed

\subsection{Proofs of Lemma~\ref{lem:bias-new} and Corollary~\ref{cor:bias}}

Before we present the proofs of \cref{lem:bias-new} and \cref{cor:bias}, for convenience we introduce a formal definition of $\beta$-Hölder smoothness. Under this definition, the ``$\beta$-Hölder smoothness'' of $\mathcal{A}_{k}(\cdot)$ on $\mathcal{Z}$ is equivalent to \cref{assump1} provided that $\mathcal{Z}$ is compact in $\mathbb{R}^{d}$.

\begin{definition}[$\beta$-Hölder smoothness \cite{evans2022partial}]\label{def:holder}
 Fix $p\in \mathbb{Z}_{\geq 0}$ and $\alpha\in [0,1]$. Given $\Omega\subset \mathbb{R}^{d}$, suppose a function $f$ and its derivatives up to order $p$ are bounded on $\overline{\Omega}:=\{ s: \inf_{t\in \Omega}\lVert s-t \rVert =0\}$, the closure of $\Omega$. We define
 \begin{equation}
  \|f\|_{C^{p, \alpha}(\Omega)}=\|f\|_{C^p (\Omega)}+\max _{|\gamma|=p}\sup _{x, y \in \Omega, x \neq y} \frac{|D^{\gamma}f(x)-D^{\gamma}f(y)|}{\|x-y\|^\alpha},
 \end{equation}
 where $\gamma$ ranges over multi-indices and $\|f\|_{C^p(\Omega)}=\max _{|\gamma| \leq p} \sup _{x \in \Omega}\left|D^\gamma f(x)\right|$. We call $f$ \emph{Hölder smooth with exponent $\beta$} on $\Omega$, or simply \emph{$\beta$-Hölder smooth}, if $\lVert f \rVert_{C^{\lfloor \beta\rfloor,\beta-\lfloor\beta\rfloor}(\Omega)}<\infty$.
\end{definition}

\paragraph*{Remark} For any $p\in\mathbb{Z}_{\geq 0}$, $C^{p,1}(\Omega)$ and $C^{p+1}(\Omega)=C^{p+1,0}(\Omega)$ are different functional spaces. If $\Omega$ is compact, then it is well-known that $C^{p+1}(\Omega)\subset C^{p,1}(\Omega)$. Our definition of $\beta$-Hölder smoothness differs slightly from some standard references as for $\beta\in \mathbb{Z}_{\geq 1}$ we consider $C^{\beta}(\Omega)$ rather than $C^{\beta-1,1}(\Omega)$.

\begin{proof}[Proof of equivalence in assumption]
  The condition $\lVert f \rVert_{C^{p,\alpha}(\Omega)}<\infty$ of \cref{def:holder} clearly implies \cref{assump1} with $\beta=p+\alpha$. It suffices to prove that \cref{assump1} leads to $\lVert f \rVert_{C^{p,\alpha}(\mathcal{Z})}<\infty$ in our setup when $\mathcal{Z}$ is compact in $\mathbb{R}^{d}$. In fact, \cref{assump1} together with compactness of $\mathcal{Z}$ implies that for any multi-index $\gamma$ of order $p$, $\lVert D^{\gamma}\mathcal{A}_{k}(\cdot) \rVert$ is bounded. Thus, for any multi-index $\widetilde{\gamma}$ of order $p-1$, $D^{\widetilde{\gamma}}\mathcal{A}_{k}(\cdot)$ is continuously differentiable with bounded partial derivatives, meaning that $\lVert D^{\widetilde{\gamma}}\mathcal{A}_{k}(\cdot) \rVert$ is also bounded. By induction for any multi-index $\gamma$ of order up to $p$, we have $D^{\gamma}\mathcal{A}_{k}(\cdot)$ is bounded. Thus, $\lVert f \rVert_{C^{p}(\mathcal{Z})}<\infty$. Together with the assumption that $\lvert D^{\gamma}\mathcal{A}_{k}(z)-D^{\gamma}\mathcal{A}_{k}(\tilde{z}) \rvert\leq c_{1}\lVert z-\tilde{z} \rVert^{\alpha}$ for all $\gamma$ of order $p$, we conclude that $\lVert f \rVert_{C^{p,\alpha}(\mathcal{Z})}<\infty$.
\end{proof}

\begin{proof}[Proof of \cref{lem:bias-new}]
For convenience denote $p=\lfloor \beta\rfloor$ and $\alpha=\beta-p$. For any $\xi \in \mathcal{Z}_{h}=\{ z\in \mathcal{Z}: \lVert z-z_{0} \rVert\geq h,\,\forall\, z_{0}\in \mathbb{R}^{d}\setminus \mathcal{Z} \}$, we decompose $\widetilde{\mathcal{A}}_k(\xi)=\mathcal{A}_k^*(\xi) + \mathcal{A}_{k}^{\dag}(\xi)$, where
\begin{align} \label{bias-proof-1}
\mathcal{A}_{k}^*(\xi) &= \sum_{m=1}^M \int_{\calR_m}\frac{1}{h^{d}}\mathcal{K}\Bigl(\frac{z-\xi}{h}\Bigr)\mathcal{A}_{k}(z) dz =\int_{\mathcal{Z}}\frac{1}{h^{d}}\mathcal{K}\Bigl(\frac{z-\xi}{h}\Bigr)\mathcal{A}_{k}(z) dz, \cr
\mathcal{A}_{k}^{\dag}(\xi) &=  \sum_{m=1}^M \int_{\calR_m}\biggl[\zeta_{m}(\xi)- \frac{1}{h^{d}} \mathcal{K}\Bigl(\frac{z-\xi}{h}\Bigr)\biggr]\mathcal{A}_{k}(z)dz. 
\end{align}

Before we analyze these two quantities, we first clarify the choice of kernels here. To reduce bias, we have introduced higher-order kernels in \cref{subsec:KS} when $d=1$. Here in our proof, we consider for convenience higher-order kernels defined on $[0,1]\subset \mathbb{R}$ and let $\mathcal{K}(u)=0$ for $u\in \mathbb{R}\setminus [0,1]$. Note that such desirable higher-order kernels exist, with concrete examples given in \cite{fan1992bias,tsybakov2009introduction}. We further remark that similar analysis could potentially be generalized to higher-order kernels $\mathcal{K}(u)$ defined on $\mathbb{R}$ that decays exponentially fast in $u$. As a reminder, an order-$\beta$ kernel on $[0,1]\subset\mathbb{R}$ statisfies
\begin{equation}\label{eq:momentconditionhigh-remind}
  \int_{0}^{1}\mathcal{K}(u) \, du = 1, \quad \int_{0}^{1} u^k \mathcal{K}(u) \, du = 0 \quad \text{for } k = 1, \ldots, p, \quad \int_{0}^{1}\lvert u \rvert^{\beta}\lvert\mathcal{K}(u)\rvert du<\infty.
\end{equation}
We can generalize this concept to $d\geq 2$ by considering the product kernel:
\begin{equation*}
  \mathcal{K}(\cdot):\mathcal{Z}\subset \mathbb{R}^{d}\to\mathbb{R},\quad \mathcal{K}(u):=\mathcal{K}_{1}(u_{1})\mathcal{K}_{2}(u_{2})\cdots \mathcal{K}_{d}(u_{d}),
\end{equation*}
where $u=(u_{1},\ldots,u_{d})'$ and $\mathcal{K}_{1}(\cdot),\ldots,\mathcal{K}_{d}(\cdot): [0,1]\subset \mathbb{R}\to\mathbb{R}$ are order-$p$ kernels on $[0,1]$.


\subsection*{Upper Bounding $|\mathcal{A}_{k}^*(\xi)-\mathcal{A}_{k}(\xi)|$:} 

For $d=1$, given $z\in {\cal Z}$ we perform the Taylor expansion of $\mathcal{A}_{k}(z)$ around any point $\xi\in \mathcal{Z}_{h}$. Specifically, there exists a convex combination of $z$ and $\xi$ denoted by $t=t(z, \xi)\in{\cal Z}$ such that 
\begin{align*}
\mathcal{A}_{k}(z) & = \mathcal{A}_{k}(\xi) + \sum_{j=1}^{p -1}\frac{\mathcal{A}_{k}^{(j)}(\xi)}{j!}(z-\xi)^j + \frac{\mathcal{A}_{k}^{(p )}(t(z,\xi))}{p !}(z-\xi)^{p }\cr
&= \mathcal{A}_{k}(\xi) + \sum_{j=1}^{p }\frac{\mathcal{A}_{k}^{(j)}(\xi)}{j!}(z-\xi)^j + \frac{[\mathcal{A}_{k}^{(p )}(t)-\mathcal{A}_{k}^{(p )}(\xi)]}{p !}(z-\xi)^{p }.
\end{align*}
 Note that the summation in the second line is taken from $1$ to $p$ rather than $p-1$.
We substitute this in the expression of $\mathcal{A}_{k}^*(\xi)$, and it follows that $\mathcal{A}_{k}^*(\xi) =  \sum_{j=0}^{p +1}b_j(\xi)$, where 
\begin{equation} \label{bias-proof-2}
b_j(\xi)=
\begin{cases}
\mathcal{A}_{k}(\xi) \int_{\mathcal{Z}}\frac{1}{h}\mathcal{K}(\frac{z-\xi}{h})dz, & \mbox{for }j=0,\cr
\frac{\mathcal{A}_{k}^{(j)}(\xi)}{j!} \int_{\mathcal{Z}}\frac{(z-\xi)^j}{h}\mathcal{K}(\frac{z-\xi}{h})dz, & \mbox{for } 1\leq j\leq p , \cr
\frac{1}{p !} \int_{\mathcal{Z}}\bigl[\mathcal{A}_{k}^{(p )}(t)-\mathcal{A}_{k}^{(p )}(\xi)\bigr]\frac{(z-\xi)^{p }}{h}\mathcal{K}(\frac{z-\xi}{h})dz, & \mbox{for }j=p +1. 
\end{cases}
\end{equation}

For $d\geq 2$, we define a sequence of points $\{ \xi^{[\ell]} \}_{\ell=0}^{d}$ where $\xi^{[0]}=\xi=(\xi_{1},\ldots,\xi_{d})'$, $\xi^{[d]}=z=(z_{1},\ldots,z_{d})'$ and $\xi^{[\ell]}=(z_{1},\ldots, z_{\ell},\xi_{\ell+1},\ldots,\xi_{d})'$ for $1\leq \ell\leq d-1$. We perform the Taylor expansion of $\mathcal{A}_{k}(\xi^{[\ell]})$ around the point $\xi^{[\ell-1]}$ for $\ell=1,\ldots,d$. Specifically, there exists a convex combination of $\xi^{[\ell]}$ and $\xi^{[\ell-1]}$ denoted by $t_{\ell}=t_{\ell}(\xi^{[\ell]},\xi^{[\ell-1]})\in\mathcal{Z}$ such that
\begin{equation*}
  \mathcal{A}_{k}(\xi^{[\ell]})=\mathcal{A}_{k}(\xi^{[\ell-1]})+\sum_{j=1}^{p }\frac{\frac{\partial^{j}}{\partial z_{\ell}^{j}}\mathcal{A}_{k}(\xi^{[\ell-1]})}{j!}(z_{\ell}-\xi_{\ell})^j + \frac{[\frac{\partial^{p}}{\partial z_{\ell}^{p}}\mathcal{A}_{k}(t_{k})-\frac{\partial^{p}}{\partial z_{\ell}^{p}}\mathcal{A}_{k}(\xi^{[\ell-1]})]}{p !}(z_{\ell}-\xi_{\ell})^{p }.
\end{equation*}
Again we substitute this in the expression of $\mathcal{A}_{k}(\xi)$ and get $\mathcal{A}_{k}^{\ast}(\xi)=\sum_{j=1}^{p+1}b_{j}(\xi)$
\begin{equation} \label{bias-proof-ddayu1}
b_j(\xi)=\left\{
\begin{aligned}
&\mathcal{A}_{k}(\xi) \int_{\mathcal{Z}}\frac{1}{h^{d}}\mathcal{K}\Bigl(\frac{z-\xi}{h}\Bigr)dz, && \mbox{for }j=0,\cr
&\sum_{\ell=1}^{d}\frac{1}{j!}\int_{\mathcal{Z}} \frac{\partial^{j}}{\partial z_{\ell}^{j}}\mathcal{A}_{k}(\xi^{[\ell-1]})\frac{(z_{\ell}-\xi_{\ell})^j}{h^{d}}\mathcal{K}\Bigl(\frac{z-\xi}{h}\Bigr)dz, && \mbox{for } 1\leq j\leq p , \cr
&\sum_{\ell=1}^{d}\frac{1}{p !} \int_{\mathcal{Z}}\Bigl[\frac{\partial^{p}}{\partial z_{\ell}^{p}}\mathcal{A}_{k}(t_{\ell})-\frac{\partial^{p}}{\partial z_{\ell}^{p}}\mathcal{A}_{k}(\xi^{[\ell-1]})\Bigr]\frac{(z_{\ell}-\xi_{\ell})^{p }}{h^{d}}\mathcal{K}\Bigl(\frac{z-\xi}{h}\Bigr)dz, && \mbox{for }j=p +1. 
\end{aligned}\right.
\end{equation}

Since $\xi\in\mathcal{Z}_{h}$, the regions of integrations in \eqref{bias-proof-2} and \eqref{bias-proof-ddayu1} can be rewritten as $\mathbb{R}^{d}$. Noticing that $\int_{\mathbb{R}^{d}}\mathcal{K}(u)du=1$, by a change of variable we obtain $\int_{\mathbb{R}^{d}}\frac{1}{h^{d}}\mathcal{K}(\frac{z-\xi}{h})d\xi=1$. Combining this with the fact that ${\cal Z}=\bigsqcup_{m=1}^M\calR_m$, we deduce that
\[
b_0(\xi) = \mathcal{A}_{k}(\xi) \int_{\mathbb{R}^{d}}\frac{1}{h^{d}}\mathcal{K}\Bigl(\frac{z-\xi}{h}\Bigr)dz = \mathcal{A}_{k}(\xi).
\]
For each $1\leq j\leq p $ and $1\leq \ell\leq d$, we check that $\xi^{[\ell]}-\xi^{[\ell-1]}=(0,\ldots,0,z_{\ell}-\xi_{\ell},0\ldots,0)'$. Thus,
\begin{equation*}
\begin{aligned}
  &\int_{\mathbb{R}^{d}}\frac{\partial^{j}}{\partial z_{\ell}^{j}}\mathcal{A}_{k}(\xi^{[\ell-1]})\frac{(z_{\ell}-\xi_{\ell})^j}{h^{d}}\mathcal{K}\Bigl(\frac{z-\xi}{h}\Bigr)dz\\
  =&\int_{\mathbb{R}^{d}}\frac{\partial^{j}}{\partial z_{\ell}^{j}}\mathcal{A}_{k}((z_{1},\ldots,z_{\ell-1},\xi_{\ell},\ldots,\xi_{d})')\frac{(z_{\ell}-\xi_{\ell})^{j}}{h^{d}}\mathcal{K}_{1}\Bigl(\frac{z_{1}-\xi_{1}}{h}\Bigr)\cdots \mathcal{K}_{d}\Bigl(\frac{z_{d}-\xi_{d}}{h}\Bigr) dz_{1}\cdots d z_{d}\\
  =&\int_{\mathbb{R}^{\ell-1}}\frac{\partial^{j}}{\partial z_{\ell}^{j}}\mathcal{A}_{k}((z_{1},\ldots,z_{\ell-1},\xi_{\ell},\ldots,\xi_{d})')\frac{1}{h^{\ell-1}}\mathcal{K}_{1}\Bigl(\frac{z_{1}-\xi_{1}}{h}\Bigr)\cdots \mathcal{K}_{\ell-1}\Bigl(\frac{z_{\ell-1}-\xi_{\ell-1}}{h}\Bigr)dz_{1}\cdots d z_{\ell-1}\\
  &\quad  \int_{\mathbb{R}^{d-\ell}}\frac{1}{h^{d-\ell}}\mathcal{K}_{\ell+1}\Bigl(\frac{z_{\ell+1}-\xi_{\ell+1}}{h}\Bigr)\cdots \mathcal{K}_{d}\Bigl(\frac{z_{d}-\xi_{d}}{h}\Bigr) dz_{\ell+1}\cdots d z_{d}\int_{\mathbb{R}}\frac{(z_{\ell}-\xi_{\ell})^{j}}{h}\mathcal{K}_{\ell}\Bigl(\frac{z_{\ell}-\xi_{\ell}}{h}\Bigr)d z_{\ell}.
  \end{aligned}
\end{equation*}
Since $\int_{\mathbb{R}}u^j \mathcal{K}_{\ell}(u)du=0$, we have 
$$
\int_{\mathbb{R}}\frac{(z_{\ell}-\xi_{\ell})^{j}}{h}\mathcal{K}_{\ell}\Bigl(\frac{z_{\ell}-\xi_{\ell}}{h}\Bigr)d\xi_{\ell}= h^{j}\int_{\mathbb{R}}\frac{1}{h}\Bigl(\frac{z_{\ell}-\xi_{\ell}}{h}\Bigr)^{j}\mathcal{K}\Bigl(\frac{z_{\ell}-\xi_{\ell}}{h}\Bigr)d\xi_{\ell}=0.
$$ 
Thus, $\int_{\mathbb{R}^{d}}\frac{\partial^{j}}{\partial z_{\ell}^{j}}\mathcal{A}_{k}(\xi^{[\ell-1]})\frac{(z_{\ell}-\xi_{\ell})^j}{h^{d}}\mathcal{K}\Bigl(\frac{z-\xi}{h}\Bigr)dz=0$ and $b_{j}(\xi)=0$ for $1\leq j\leq p$.

Next, using the $\beta$-Hölder smoothness of $\mathcal{A}_{k}(z)$, we deduce that 
$$
\Bigl|\frac{\partial^{p}}{\partial z_{\ell}^{p}}\mathcal{A}_{k}(t_{\ell})-\frac{\partial^{p}}{\partial z_{\ell}^{p}}\mathcal{A}_{k}(\xi^{[\ell-1]})\Bigr|\leq C|t_{\ell}-\xi^{[\ell-1]}|^{\beta-p }\leq C|\xi^{[\ell]}-\xi^{[\ell-1]}|^{\beta-p }=C \lvert z_{\ell}-\xi_{\ell} \rvert^{\beta-p}. 
$$
As a result, we have
\begin{equation*}
\begin{aligned}
  &\biggl\lvert \int_{\mathbb{R}^{d}}\Bigl[\frac{\partial^{p}}{\partial z_{\ell}^{p}}\mathcal{A}_{k}(t_{\ell})-\frac{\partial^{p}}{\partial z_{\ell}^{p}}\mathcal{A}_{k}(\xi^{[\ell-1]})\Bigr]\frac{(z_{\ell}-\xi_{\ell})^j}{h^{d}}\mathcal{K}\Bigl(\frac{z-\xi}{h}\Bigr)dz\biggr\rvert\\
  =&\int_{\mathbb{R}^{\ell-1}}\Bigl\lvert \frac{\partial^{p}}{\partial z_{\ell}^{p}}\mathcal{A}_{k}(t_{\ell})-\frac{\partial^{p}}{\partial z_{\ell}^{p}}\mathcal{A}_{k}(\xi^{[\ell-1]})\Bigr\rvert\frac{1}{h^{\ell-1}}\biggl\lvert\mathcal{K}_{1}\Bigl(\frac{z_{1}-\xi_{1}}{h}\Bigr)\cdots \mathcal{K}_{\ell-1}\Bigl(\frac{z_{\ell-1}-\xi_{\ell-1}}{h}\Bigr)\biggr\rvert dz_{1}\cdots d z_{\ell-1}\\
  &\quad \int_{\mathbb{R}}\frac{\lvert z_{\ell}-\xi_{\ell}\rvert^{p}}{h}\biggl\lvert\mathcal{K}_{\ell}\Bigl(\frac{z_{\ell}-\xi_{\ell}}{h}\Bigr)\biggr\rvert d z_{\ell} \int_{\mathbb{R}^{d-\ell}}\frac{1}{h^{d-\ell}}\mathcal{K}_{\ell+1}\Bigl(\frac{z_{\ell+1}-\xi_{\ell+1}}{h}\Bigr)\cdots \mathcal{K}_{d}\Bigl(\frac{z_{d}-\xi_{d}}{h}\Bigr) dz_{\ell+1}\cdots d z_{d}\\
  \leq & C \int_{\mathbb{R}^{\ell}}\lvert z_{\ell}-\xi_{\ell} \rvert^{\beta}\frac{1}{h^{\ell}}\biggl\lvert \mathcal{K}_{1}\Bigl(\frac{z_{1}-\xi_{1}}{h}\Bigr)\cdots \mathcal{K}_{\ell}\Bigl(\frac{z_{\ell}-\xi_{\ell}}{h}\Bigr) \biggr\rvert d z_{1}\cdots d z_{\ell}\leq C' h^{\beta}.
  \end{aligned}
\end{equation*}
Thus, $b_{p+1}(\xi)$ can be bounded by $\lvert b_{p+1}(\xi) \rvert\leq C'' h^{\beta}$ for some constant $C''>0$.

Substituting the results above in $\mathcal{A}_{k}^*(\xi) =  \sum_{j=0}^{p +1}b_j(\xi)$, we get
\begin{equation}\label{bias-proof-3}
|\mathcal{A}_{k}^*(\xi)-\mathcal{A}_{k}(\xi)|\leq Ch^{\beta}. 
\end{equation}

\subsection*{Upper Bounding $\lvert \mathcal{A}_{k}^{\dag}(\xi)\rvert$:}

Recall that for $\xi\in\mathcal{Z}_{h}$, we define $\zeta_m(\xi) = \frac{1}{\mathrm{Vol}(\mathcal{R}_m)} \int_{\mathcal{R}_m} \frac{1}{h^{d}}\mathcal{K}(\frac{z-\xi}{h}) \, dz$ and 
\begin{equation*}
  \mathcal{A}_{k}^{\dag}(\xi) =  \sum_{m=1}^M \int_{\calR_m}\biggl[\zeta_{m}(\xi)- \frac{1}{h^{d}} \mathcal{K}\Bigl(\frac{z-\xi}{h}\Bigr)\biggr]\mathcal{A}_{k}(z)dz.
\end{equation*}
Since $\mathcal{A}_{k}(z)$ is a continuous function, for each $1\leq i\leq M$, there exists $x_m\in\calR_m$ such that 
\begin{equation}\label{bias-proof-add}
\mathcal{A}_{k}(x_m) = \frac{1}{\mathrm{Vol}(\calR_m)}\int_{\calR_m}\mathcal{A}_{k}(z)dz. 
\end{equation}

For any $z\in \mathcal{Z}$, $z\neq x_{m}$, we write $F_{\mathcal{A}_{k},z,x_{m}}(\lambda):=\mathcal{A}_{k}\bigl(x_{m}+\lambda \frac{z-x_{m}}{\lVert z-x_{m} \rVert}\bigr)$ with $\lambda\in \mathbb{R}$. Define
\begin{equation*}
  l:=\inf \biggl\{ \lambda: x_{m}+\lambda \frac{z-x_{m}}{\lVert z-x_{m} \rVert}\in \mathcal{Z} \biggr\},\quad u:=\sup \biggl\{ \lambda: x_{m}+\lambda \frac{z-x_{m}}{\lVert z-x_{m} \rVert} \in \mathcal{Z}\biggr\}.
\end{equation*}
First, we show that $\lVert F_{\mathcal{A}_{k},z,x_{m}}(\cdot) \rVert_{C^{p,\alpha}([l,u])}$ is bounded by some constant $c_{1}>0$.
In fact, if we define $\mathcal{L}: \lambda\mapsto x_{m}+\lambda \frac{z-x_{m}}{\lVert z-x_{m} \rVert}$, then
$\mathcal{L}$ is a bounded linear operator from the Banach space $C^{p,\alpha}(\mathcal{Z})$ to the Banach space $C^{p,\alpha}([l,u])$. Thus, we have $\lVert F_{\mathcal{A}_{k},z,x_{m}}(\cdot) \rVert_{C^{p,\alpha}([l,u])}\leq \lVert \mathcal{L} \rVert_{op}\lVert \mathcal{A}_{k}(\cdot) \rVert_{C^{p,\alpha}(\mathcal{Z})}$.

Now we perform the Taylor expansion for $F_{\mathcal{A}_{k},z,x_{m}}(\cdot)$ at $\lambda=0$. For each $z\in\mathcal{R}_{m}$, there exists $\lambda_{z}\in [0,\lVert z-x_{m} \rVert]$ such that
\[
\mathcal{A}_{k}(z) = \mathcal{A}_{k}(x_m) + \sum_{j=1}^{p }\frac{F_{\mathcal{A}_{j},z,x_{m}}^{(j)}(0)}{j!}\lVert z-x_m\rVert^j + \frac{[F_{\mathcal{A}_{k},z,x_{m}}^{(p )}(\lambda_{z})-F_{\mathcal{A}_{k},z,x_{m}}^{(p )}(0)]}{p !}\lVert z-x_m\rVert^{p }.
\]

As a result, we obtain that ${\cal A}_k^{\dag}(\xi)=\sum_{j=0}^{p+1}e_j(\xi)$ with
\begin{equation} \label{bias-proof-nextddayu1}
e_j(\xi)=\left\{
\begin{aligned}
 &\sum_{m=1}^M \mathcal{A}_{k}(x_m)\int_{\calR_m}\Bigl[ \zeta_{m}(\xi)- \frac{1}{h^{d}}\mathcal{K}(\frac{z-\xi}{h})\Bigr]dz, && \mbox{for }j=0,\cr
&\sum_{m=1}^M \frac{1}{j!} \int_{\calR_m} F_{\mathcal{A}_{k},z,x_{m}}^{(j)}(0)\lVert z-x_{m} \rVert^j\Bigl[ \zeta_{m}(\xi)- \frac{1}{h^{d}}\mathcal{K}(\frac{z-\xi}{h})\Bigr]dz, && \mbox{for } 1\leq j\leq p , \cr
&
\sum_{m=1}^M \frac{1}{p !}\int_{\calR_m} [F_{\mathcal{A}_{k},z,x_{m}}^{(p )}(\lambda_{z})-F_{\mathcal{A}_{k},z,x_{m}}^{(p )}(0)] \lVert z-x_{m} \rVert^{p }\Bigl[ \zeta_{m}(\xi)- \frac{1}{h^{d}}\mathcal{K}(\frac{z-\xi}{h})\Bigr]dz, && \mbox{for }j=p +1. 
\end{aligned}\right.
\end{equation}

The definition of $\zeta_{m}(\xi)$ implies that $\int_{\calR_m}\frac{1}{h}\mathcal{K}(\frac{x-\xi}{h})dx = \zeta_{m}(\xi)\cdot \mathrm{Vol}(\calR_m)$. As a result, 
\begin{equation} \label{bias-proof-5}
e_0 (\xi) = \sum_{m=1}^M \mathcal{A}_{k}(x_m)\left[ \zeta_{m}(\xi) \cdot \mathrm{Vol}(\calR_m) - \int_{\calR_m} \frac{1}{h^{d}}\mathcal{K}\Bigl(\frac{z-\xi}{h}\Bigr)dz\right]=0. 
\end{equation}

Next we analyze $e_j(\xi)$ for $1\leq j\leq p $, we first note that $\zeta_{m}(\xi)=\frac{1}{\mathrm{Vol}(\calR_m)}\int_{x\in\calR_m}\frac{1}{h^{d}}\mathcal{K}(\frac{x-\xi}{h})dx$. Additionally, we can write $\frac{1}{h^{d}}\mathcal{K}(\frac{z-\xi}{h})=\frac{1}{\mathrm{Vol}(\calR_m)} \int_{x\in\calR_m}\frac{1}{h^{d}}\mathcal{K}(\frac{z-\xi}{h})dx$. It follows that
\begin{equation} \label{bias-proof-add2}
\zeta_{m}(\xi) - \frac{1}{h^{d}}\mathcal{K}\Bigl(\frac{z-\xi}{h}\Bigr) = \frac{1}{\mathrm{Vol}(\calR_m)h^{d}}\int_{x\in\calR_m}\Bigl[ \mathcal{K}\Bigl(\frac{x-\xi}{h}\Bigr)- \mathcal{K}\Bigl(\frac{z-\xi}{h}\Bigr)\Bigr]dx
\end{equation}
Using \eqref{bias-proof-add2}, for each $1\leq j\leq p $, we can write:
\[
e_j(\xi) = \sum_{m=1}^M \frac{F_{\mathcal{A}_{k},z,x_{m}}^{(j)}(0)}{j!} \frac{1}{\mathrm{Vol}(\calR_m)}\int_{(x,z)\in\calR^2_m} \frac{\lVert z-x_m\rVert^j}{h^{d}}\Bigl[ \mathcal{K}\Bigl(\frac{x-\xi}{h}\Bigr)- \mathcal{K}\Bigl(\frac{z-\xi}{h}\Bigr)\Bigr]dxdz. 
\]

On one hand, we upper bound $\lvert e_{j}(\xi)\rvert$ using the Lipschitz continuity of the kernel $\mathcal{K}(\cdot)$, i.e., $|\mathcal{K}(u)-\mathcal{K}(u')|\leq C_{\mathcal{K}}\|u-u'\|$ for all $u,u'\in [0,1]$ and a constant $C_{\mathcal{K}}>0$. But note that such continuity assumption is only defined on $[0,1]$ rather than $\mathbb{R}$ while the input of $\mathcal{K}(\cdot)$ is extended to $\mathbb{R}$ by setting the value to be $0$ on $\mathbb{R}\setminus [0,1]$. To tackle this subtle issue, we need to identify three different cases for $\mathcal{R}_{m}$. Specifically, let ${\cal U}_h(\xi)=\{x: \|x-\xi\|\leq h\}$ and we can check that ${\cal K}(\frac{x-\xi}{h})=0$ for all $x\notin {\cal U}_h(\xi)$. Then the integration over $\mathcal{R}_{m}^{2}$ is treated differently according to the overlap between $\mathcal{R}_{m}$ and $\mathcal{U}_{h}(\xi)$:
\begin{itemize}\itemsep 0pt
    \item Case 1: Define the index set as $J_{1}=\bigl\{m: {\cal R}_m\subset {\cal U}_h(\xi)\bigr\}$. We apply the Lipschitz continuity of the kernel $\mathcal{K}(\cdot)$. 
    \item Case 2: Let $J_{2}=\bigl\{m:{\cal R}_m\subset {\cal Z}\setminus {\cal U}_h(\xi)\bigr\}$. The integral vanishes because $\mathcal{K}(\cdot)=0$. 
    \item Case 3: Let $J_{3}=\bigl\{m:\calR_m\cap {\cal U}_h(\xi)\neq \emptyset$, \text{ and } $\calR_m\cap ({\cal Z}\setminus {\cal U}_h(\xi))\neq\emptyset\bigr\}$. We apply the naive upper bound $|{\cal K}(z)|\leq C$, accounting for the fact that the total volume of such $\mathcal{R}_{m}$'s is small. 
\end{itemize}
For convenience, we denote
\[
e_{j,m}(\xi) = \frac{F_{\mathcal{A}_{k},z,x_{m}}^{(j)}(0)}{j!} \frac{1}{\mathrm{Vol}(\calR_m)}\int_{(x,z)\in\calR^2_m} \frac{\lVert z-x_m\rVert^j}{h^{d}}\Bigl[ \mathcal{K}\Bigl(\frac{x-\xi}{h}\Bigr)- \mathcal{K}\Bigl(\frac{z-\xi}{h}\Bigr)\Bigr]dxdz. 
\]
For Case 1, it follows that
\begin{align*} 
\sum_{m\in J_{1}}|e_{j,m} (\xi)| & \leq C_{\mathcal{K}}\sum_{m\in J_{1}}\lVert F_{\mathcal{A}_{k},z,x_{m}}^{(j)}(0)\rVert\cdot \frac{1}{\mathrm{Vol}(\calR_m)}\int_{(x,z)\in\calR^2_m} \frac{\lVert z-x_m\rVert^j \lVert x-z\rVert}{h^{d+1}}dxdz \cr
&\leq \frac{C_{\mathcal{K}}}{h^{d+1}}\sum_{m\in J_{1}}\lVert F_{\mathcal{A}_{k},z,x_{m}}^{(j)}(0)\rVert \cdot\epsilon^{j+1}\mathrm{Vol}(\calR_m)
\leq \frac{c_{1}C_{\mathcal{K}}\epsilon^{j+1}}{h^{d+1}} \mathrm{Vol}(\mathcal{Z}). 
\end{align*} 
For Case 3, $\calR_m$ must be within a distance $\epsilon$ to the boundary of ${\cal U}_h(\xi)$. Combining this with the fact that $\calR_m$'s are non-overlapping, we obtain
\begin{equation*}
  \sum_{m\in J_{3}}\operatorname{Vol}(\mathcal{R}_{m})\asymp h^{d-1}\epsilon.
\end{equation*}
Thus, we derive that
\begin{align*}
  &\sum_{m\in J_{3}}\lvert e_{j,m}(\xi) \rvert
  \leq \sum_{m\in J_{3}}\frac{\lVert F_{\mathcal{A}_{k},z,x_{m}}^{(j)}(0)\rVert}{j!} \frac{1}{\mathrm{Vol}(\calR_m)} \int_{(x,z)\in \mathcal{R}^{2}_{m}}\frac{2C\epsilon^{j}}{h^{d}} dxdz\\
  \leq & \sum_{m\in J_{3}} \lVert F_{\mathcal{A}_{k},z,x_{m}}^{(j)}(0)\rVert \frac{2 C\epsilon^{j}\mathrm{Vol}(\mathcal{R}_{m})}{h^{d}}\asymp \frac{\epsilon^{j+1}}{h}=o\Bigl(\frac{\epsilon^{j+1}}{h^{d+1}}\Bigr).
\end{align*}
Combining Case 1--3, we get
\[
\lvert e_{j}(\xi)\rvert \leq \sum_{m\in J_{1}}\lvert e_{j,m}(\xi) \rvert+\sum_{m\in J_{2}}\lvert e_{j,m}(\xi) \rvert+\sum_{m\in J_{3}}\lvert e_{j,m}(\xi) \rvert =O\Bigl(\frac{\epsilon^{j+1}}{h^{d+1}}\Bigr).
\]



On the other hand, we can always bound the difference in kernels by sum of absolute values like what we have done in the previous Case 3.
\begin{align*}
  &\lvert e_{j}(\xi) \rvert
  \leq \sum_{m=1}^M \frac{\lVert F_{\mathcal{A}_{k},z,x_{m}}^{(j)}(0)\rVert}{j!} \frac{1}{\mathrm{Vol}(\calR_m)} \int_{(x,z)\in\calR^2_m} \frac{\lVert z-x_m\rVert^j}{h^{d}}\Bigl\lvert \mathcal{K}\Bigl(\frac{x-\xi}{h}\Bigr)- \mathcal{K}\Bigl(\frac{z-\xi}{h}\Bigr)\Bigr\rvert dxdz\\
  \leq &\sum_{m=1}^M \frac{\lVert F_{\mathcal{A}_{k},z,x_{m}}^{(j)}(0)\rVert}{j!} \frac{1}{\mathrm{Vol}(\calR_m)} \int_{(x,z)\in \mathcal{R}^{2}_{m}}\frac{\epsilon^{j}}{h^{d}}\Bigl[\Bigl\lvert \mathcal{K}\Bigl(\frac{x-\xi}{h}\Bigr)\Bigr\rvert+\Bigl\lvert \mathcal{K}\Bigl(\frac{z-\xi}{h}\Bigr)\Bigr\rvert\Bigr] dxdz\\
  \leq & \sum_{m=1}^M \lVert F_{\mathcal{A}_{k},z,x_{m}}^{(j)}(0)\rVert\frac{1}{\mathrm{Vol}(\calR_m)} 2\epsilon^{j}\mathrm{Vol}(\mathcal{R}_{m})\int_{\mathcal{R}_{m}}\frac{1}{h^{d}}\Bigl\lvert \mathcal{K}\Bigl(\frac{u-\xi}{h}\Bigr)\Bigr\rvert du\\
  = & 2\lVert F_{\mathcal{A}_{k},z,x_{m}}^{(j)}(0)\rVert \epsilon^{j}\int_{\mathcal{Z}}\frac{1}{h^{d}}\Bigl\lvert \mathcal{K}\Bigl(\frac{u-\xi}{h}\Bigr)\Bigr\rvert du
  \leq c_{1}C_{\mathcal{K}}' \epsilon^{j},
\end{align*}
where $C_{\mathcal{K}}'=\int \lvert \mathcal{K}(u)\rvert du$ is a constant associated to the kernel $\mathcal{K}$. 
Therefore, we conclude that 
\begin{equation}\label{bias-proof-6}
\lvert e_{j} (\xi)\rvert=O(\epsilon^{j}\wedge h^{-(d+1)}\epsilon^{j+1}).
\end{equation}

Now we proceed to bound $\lvert e_{p+1}(\xi) \rvert$.
Again by $\beta$-Hölder smoothness of $F_{\mathcal{A}_{k},z,x_{m}}$, we have: 
$$
|F_{\mathcal{A}_k,z,x_{m}}^{(p )}(\lambda_{z})-F_{\mathcal{A}_k,z,x_{m}}^{(p )}(0)|\leq C\lambda_{z}^{\beta-p}\leq C\lVert z-x_m\rVert^{\beta-p }\leq C\epsilon^{\beta-p }.
$$ 
Combining this result with \eqref{bias-proof-add2}, we can deduce that
\begin{align*}
|e_{p +1}(\xi)|& \leq C\sum_{m=1}^M \epsilon^{\beta}\int_{z\in\calR_m}\Bigl|\zeta_{m}(\xi)-\frac{1}{h^{d}}\mathcal{K}\Bigl(\frac{z-\xi}{h}\Bigr)\Bigr|dz\cr
&\leq C\sum_{m=1}^M \epsilon^{\beta}\frac{1}{\mathrm{Vol}(\calR_m)h^{d}}\int_{(x,z)\in\calR^2_m}\Bigl| \mathcal{K}\Bigl(\frac{x-\xi}{h}\Bigr)- \mathcal{K}\Bigl(\frac{z-\xi}{h}\Bigr)\Bigr|dxdz.
\end{align*}
On one hand, following a similar analysis of utilizing the Lipschitz continuity of $\mathcal{K}$, we obtain that
\begin{align*}
  |e_{p +1}(\xi)|
  \leq C\sum_{m=1}^M \epsilon^{\beta}\frac{1}{\mathrm{Vol}(\calR_m)h^{d}}\cdot\frac{\epsilon}{h}\mathrm{Vol}^2(\calR_m)
= O\bigl(\epsilon^{\beta+1}/h^{d+1}\bigr). 
\end{align*}
On the other hand, we derive that
\begin{align*}
 \lvert e_{p +1}(\xi) \rvert &\leq 
 C\sum_{m=1}^M \epsilon^{\beta}\frac{1}{\mathrm{Vol}(\calR_m)}\int_{(x,z)\in\calR^2_m}\frac{1}{h^{d}}\Bigl[\Bigl| \mathcal{K}\Bigl(\frac{x-\xi}{h}\Bigr)\Bigr|+ \Bigl|\mathcal{K}\Bigl(\frac{z-\xi}{h}\Bigr)\Bigr|\Bigr]dxdz \cr
 &\leq 2C C_{\mathcal{K}} \epsilon^{\beta}= O(\epsilon^{\beta}).
\end{align*}
Thus, we conclude that
\begin{equation}\label{bias-proof-7}
  \lvert e_{p +1} (\xi)\rvert= O(\epsilon^{\beta}\wedge h^{-(d+1)}\epsilon^{\beta+1}).
\end{equation}
Recall that $\mathcal{A}_{k}^\dag(\xi) =  \sum_{j=0}^{p +1}e_j(\xi)$. We combine \eqref{bias-proof-5}, \eqref{bias-proof-6}, and \eqref{bias-proof-7} to obtain:
\begin{equation} \label{bias-proof-8}
\lvert \mathcal{A}_{k}^{\dag}(\xi)\rvert\leq C \min \{\epsilon^{\beta\wedge 1}, h^{-(d+1)}\epsilon^{1+\beta\wedge 1}\}. 
\end{equation}
\eqref{bias-proof-3} and \eqref{bias-proof-8} together imply 
\begin{equation*}
  \lvert \widetilde{\mathcal{A}}_k(\xi)-\mathcal{A}_k(\xi)\rvert\leq C(h^\beta +\min\{\epsilon^{ \beta\wedge 1}, h^{-(d+1)}\epsilon^{1+\beta\wedge 1}\}).
\end{equation*}
\end{proof}

\begin{proof}[Proof of \cref{cor:bias}]
  Note that for $\gamma \in [0,1]$, the quantity $\bigl(\epsilon^{\beta\wedge 1}\bigr)^{\gamma}\bigl(h^{-(d+1)}\epsilon^{1+\beta\wedge 1}\bigr)^{1-\gamma}$ interpolates between $\epsilon^{\beta\wedge 1}$ and $h^{-(d+1)}\epsilon^{1+\beta\wedge 1}$. Hence the upper bound in \cref{lem:bias-new} can be reformulated as 
\[
\lvert \widetilde{\mathcal{A}}_k(\xi)-\mathcal{A}_k(\xi)\rvert \leq Ch^\beta +C\min_{\gamma\in [0,1]} h^{(\gamma-1)(d+1)}\epsilon^{1-\gamma+\beta\wedge 1}.
\] 
By choosing $h=\epsilon^{\frac{1-\gamma+\beta\wedge 1}{(1-\gamma)(d+1)+\beta}}$, we get $|\widetilde{\mathcal{A}}_k(\xi)-\mathcal{A}_k(\xi)| \leq C\epsilon^{\max\limits_{\gamma \in [0,1]}\theta(\beta, \gamma)}$, where $\theta(\beta, \gamma)=\frac{(1-\gamma+\beta\wedge 1)\beta}{(1-\gamma)(d+1)+\beta}$. Notably $\theta (\beta, \gamma)$ is always maximized at either $\gamma=0$ or $\gamma=1$. Specifically if $\beta<d+1$, $\theta (\beta,\gamma)$ is maximized at $\gamma=1$ and $\max \theta(\beta,\gamma)=\theta(\beta,1)=\beta\wedge 1$. If $\beta\geq d+1$, $\theta(\beta,\gamma)$ is maximized at $\gamma=0$ and $\max \theta (\beta,\gamma)=\theta(\beta,0)=\frac{2\beta}{d+1+\beta}$.

\end{proof}

\subsection{Proof of \cref{lem:var-new}}
Note that $\widehat{\cal A}_k$ is obtained from the estimated topic vector $\widehat{A}_k^{\text{net}}$ at the hyperword level, and $\widetilde{\cal A}_k$ is constructed from $A_k^{\text{net}}$ in a similar way. Therefore, the key of the proof is to investigate the difference between $\widehat{A}_k^{\text{net}}$ and $A_k^{\text{net}}$. 
It depends on the accuracy of traditional topic modeling on hyperword counts. Since this step is conducted by Topic-SCORE \citep{ke2022using}, we quote an existing result: 

\begin{theorem}[Error rate of Topic-SCORE \citep{ke2024entry}] \label{thm:TSCORE}
Fix $K\geq 2$. Let $h_m^{\text{net}}=\sum_{j=1}^K A_k^{\text{net}}(m)$, for $1\leq m\leq M$, and $H^{\text{net}}=\diag(h_1^{\text{net}}, h_2^{\text{net}}, \ldots, h_M^{\text{net}})$. Define two matrices: $ \Sigma^{\text{net}}_A = (A^{\text{net}})'\bigl(H^{\text{net}}\bigr)^{-1}A^{\text{net}}$ and $\Sigma^*_W= \frac{1}{n} \sum_{i=1}^n (1-N_i^{-1})w_iw_i'$. We assume that there exist positive constants $\tilde{c}_1$-$\tilde{c}_4$ such that 
\begin{itemize} \itemsep 0pt
    \item[(a)] $\min_{1\leq m\leq M}\{ h_m^{\text{net}}\}\geq \tilde{c}_1K/M$. 
    \item[(b)] $\lambda_{\min}(\Sigma^*_W)\geq \tilde{c}_2$, $\lambda_{\min}(\Sigma^{\text{net}}_A)\geq \tilde{c}_2$, and $\min_{1\leq k,\ell\leq K}\Sigma^{\text{net}}_A(k,\ell)\geq \tilde{c}_2$. 
    \item[(c)] $\tilde{c}_3^{-1}\bar{N}\leq \min_{1\leq i\leq n}\{N_i\}\leq \max_{1\leq i\leq n}\{N_i\}\leq  \tilde{c}_3\bar{N}$,  where $\bar{N}=n^{-1}\sum_{i=1}^n N_i$. 
    \item[(d)] $\min\{M, \bar{N}\}\geq \log^3(n)$, $\max\{\log(M), \log( \bar{N})\} \leq \tilde{c}_4\log(n)$, and $M\log^2(n)\leq n\bar{N}$.
    \item[(e)] Each $A_k^{\text{net}}$ has at least one anchor hyperword (we say that $m$ is an anchor hyperword of $A_k^{\text{net}}$ if $A_k^{\text{net}}(m)\neq 0$ and $A_\ell^{\text{net}}(m)=0$ for all $\ell \neq k$). 
\end{itemize}
Let $\widehat{A}_1^{\text{net}}, \ldots, \widehat{A}_K^{\text{net}}$ be the estimates obtained by applying Topic-SCORE to $X^{\text{net}}$. 
There exists a constant $C^*>0$ such that with probability $1-O(n^{-3})$, 
\[
\sum_{k=1}^K |\widehat{A}_k^{\text{net}}(m)-A_k^{\text{net}}(m)|\leq C^*h_m^{\text{net}}\sqrt{\frac{M\log(n)}{n\bar{N}}}, \quad\mbox{simultaneously for }1\leq m\leq M. 
\]
\end{theorem}

To apply Theorem~\ref{thm:TSCORE}, we must verify that the induced topic model on hyperword counts satisfies the conditions (a)-(e). For convenience, we use $|{\cal R}_m|$, instead of $\mathrm{Vol}({\cal R}_m)$, to denote the volume of ${\cal R}_m$ throughout this proof; similar notations apply to other sets.  
First, we check Condition (a). In Lemma~\ref{lem:induced-model}, we have seen that $A_k^{\text{net}}(m)=\int_{{\cal R}_m}{\cal A}_k(z)dz$. 
It follows that $h^{\text{net}}_m =\int_{ {\cal R}_m}h(z)dz$, where $h(z)=\sum_{k=1}^K {\cal A}_k(z)$.
By our assumptions, $h(z)$ is lower bounded by a constant, and $|{\cal R}_m|$ is lower bounded by a constant times $1/M$. We immediately obtain:
\beq \label{lem-var-0}
h^{\text{net}}_m \geq C^{-1}M^{-1}, \qquad 1\leq m\leq M. 
\eeq

Next, we check Condition (b). 
The first inequality is about $\Sigma^*_W$. By Assumption~\ref{assump3},  $\lambda_{\min}(\Sigma_W)\geq c_4$, for a constant $c_4>0$. It suffices to bound $\|\Sigma_W-\Sigma_W^*\|$. Comparing the definitions, we find that
\beq \label{lem-var-1}
\|\Sigma_W-\Sigma^*_{W}\|=\Bigl\|\frac{1}{n}\sum_{i=1}^n N_i^{-1}w_iw_i'\Bigr\|\leq \frac{C}{n}\sum_{i=1}^n \frac{1}{N_i}. 
\eeq
Since each $N_i$ has a $\mathrm{Poisson}(N)$ distribution, we use the following lemma about Poisson tail probabilities, whose proof can be found in \cite{canonne2017short}:
\begin{lemma} \label{lem:Poisson-tail}
    Suppose $X\sim \mathrm{Poisson}(\lambda)$. For any $x>0$, $\mathbb{P}(|x-\lambda|>x)\leq 2\exp(-\frac{x^2}{2(\lambda + x)})$. 
\end{lemma}
We apply this result with $\lambda=N$ and $x=N/2$. It follows that 
\[
\mathbb{P}\bigl(|N_i-N|\leq N/2\bigr)\leq 2\exp(-N/3).  
\]
Since $N\geq \log^3(n)$, the probability on the right hand side is much smaller than $n^{-c}$, for any constant $c>0$. In particular, this probability is $o(n^{-3})$. Combining this with the probability union bound, we conclude that with probability $1-o(n^{-2})$,
\beq \label{lem-var-2}
N/2\leq N_i\leq  3N/2, \qquad\mbox{simultaneously for all }1\leq i\leq n.  
\eeq
We combine \eqref{lem-var-2} and \eqref{lem-var-1}. It yields that $\|\Sigma_W-\Sigma_W^*\|\leq CN^{-1}\leq C\log^{-3}(n)$, with probability $1-o(n^{-2})$. 
This claim, together with our assumption of $\lambda_{\min}(\Sigma_W)\geq c_4$, implies that $\lambda_{\min}(\Sigma_W^*)$ is lower bounded by a constant.

The remaining two inequalities in Condition~(b) are about $\Sigma_A^{\text{net}}$. Let $\|\cdot\|_{\max}$ denote the element-wise maximum norm of a matrix, and let $\mathrm{diam}(\cdot)$ denote the diameter of a set. The following lemma is proved in Section~\ref{proof:lem-SigmaA}: 

\begin{lemma} \label{lem:SigmaA}
Under the assumptions of Lemma~\ref{lem:var-new}, there exists a constant $C>0$ such that $\|\Sigma_A^{\text{net}}-\Sigma_{\cal A}\|\leq C\max_{1\leq m\leq M} [\operatorname{diam}({\cal R}_m)]^{\beta\wedge 1}$, with $\operatorname{diam}(\calR_m)= \sup_{x,z\in \calR_m}\lVert x-z \rVert$.
\end{lemma}

By basic linear algebra, $\lambda_{\min}(\Sigma^{\text{net}}_A)\geq \lambda_{\min}(\Sigma_{\cal A})-\|\Sigma_A^{\text{net}}-\Sigma_{\cal A}\|$, and $\min_{k,\ell}\Sigma^{\text{net}}_{A}\geq  \min_{k,\ell}\Sigma_{\cal A}-\|\Sigma_A^{\text{net}}-\Sigma_{\cal A}\|$. 
Our assumptions state that $\lambda_{\min}(\Sigma_{\cal A})$ and $\min_{k,\ell}\Sigma_{\cal A}(k,\ell)$ are both lower bounded by constants. Additionally,  by Lemma~\ref{lem:SigmaA} and our choice of the net, $\|\Sigma_A^{\text{net}}-\Sigma_{\cal A}\|$ is a sufficiently small constant. Combining these results, we conclude that $\lambda_{\min}(\Sigma_{A}^{\text{net}})$ and $\min_{k,\ell}\Sigma_A^{\text{net}}(k,\ell)$ are both lower bounded by constants. This has verified the last two inequalities of Condition (b). 

Finally, we verify Conditions (c)-(e). Condition (c) follows directly from \eqref{lem-var-2}. Since $M\asymp \epsilon^{-d}$, Condition~(d) is a consequence of the assumptions on $\epsilon$. Condition (e) is implied by \eqref{lem-identifiability-2} and the fact that the volume of $S_k$ is lower bounded (this is argued in the paragraph below \eqref{lem-identifiability-2}).

Now, we have shown that all the conditions in Theorem~\ref{thm:TSCORE} hold with probability $1-o(n^{-2})$. We apply this theorem and notice that $\bar{N}\leq N/2$ by \eqref{lem-var-2}. 
As a result, with probability $1-o(n^{-2})$, simultaneously for $1\leq m\leq M$, 
\beq \label{lem-var-3}
\sum_{k=1}^K |\widehat{A}_k^{\text{net}}(m)-A_k^{\text{net}}(m)|\leq C^* h_m^{\text{net}}\sqrt{\frac{M\log(n)}{\bar{N}n}} \leq Ch_m^{\text{net}}\sqrt{\frac{M\log(n)}{Nn}}. 
\eeq
By definitions, $
\widetilde{\cal A}_k(z_0) = \sum_{m=1}^M \zeta_m(z_0)A_{k}^{\text{net}}(m)$ and $\widehat{\cal A}_k(z_0) = \sum_{m=1}^M \zeta_m(z_0)\widehat{A}_{k}^{\text{net}}(m)$, 
with weights $\zeta_m(z_0)=\frac{1}{|\calR_m|}\int_{z\in\calR_m}{\cal K}_h(z-z_0)dz$.
Additionally, $h_m^{\text{net}}=\int_{\calR_m}h(z)dz$. 
Combining these observations with \eqref{lem-var-3}, we find that 
\begin{align} \label{lem-var-4}
\sum_{k=1}^K |\widehat{\cal A}_k(z_0)- &\widetilde{\cal A}_k(z_0)| \leq \sum_{m=1}^M \zeta_m(z_0)\cdot \sum_{k=1}^K |\widehat{A}_k^{\text{net}}(m)-A_k^{\text{net}}(m)|\cr
&\leq C\sum_{m=1}^M \frac{\int_{z\in \calR_m}|{\cal K}_h(z-z_0)|dz }{|\calR_m|}\cdot h_m^{\text{net}}\sqrt{\frac{M\log(n)}{Nn}} \cr
&\leq C\sqrt{\frac{M\log(n)}{Nn}} \sum_{m=1}^M \frac{\int_{\calR_m}h(z)dz}{|\calR_m|} \int_{\calR_m}|{\cal K}_h(z-z_0)|dz\cr
&\leq C\sqrt{\frac{M\log(n)}{Nn}} \sum_{m=1}^M  \int_{z\in \calR_m}|{\cal K}_h(z-z_0)|dz\cr
&\leq C\sqrt{\frac{M\log(n)}{Nn}}  \int_{z\in \mathbb{R}^d}|{\cal K}_h(z-z_0)|dz \cr
&\leq  C\sqrt{\frac{M\log(n)}{Nn}},
\end{align}
where the fourth line is because $\int_{\calR_m}h(z)dz\leq C|\calR_m|$ (by the continuity of $h(z)$), and the last line is due to that $\int_{\mathbb{R}^d}|{\cal K}(u)|du \leq C$. 
The desirable claim follows by noting that $M\asymp \epsilon^{-d}$. \qed

\paragraph*{Remark} The precise claim in Theorem~\ref{thm:TSCORE} involves an arbitrary  permutation of $\widehat{A}_1^{\text{net}}, \widehat{A}_2^{\text{net}}, \ldots, \widehat{A}_K^{\text{net}}$ (this arises from the user's choice of picking  which estimated topic vector to label as the ``first'' topic, and which  the ``second''; this labeling choice is arbitrary, but it has no impact on the algorithm's performance). 
Since each $\widehat{\cal A}_k(\cdot)$ is constructed sorely from $\widehat{A}_k^{\text{net}}$, the same permutation is inherited by $\widehat{\cal A}_1(\cdot), \ldots, \widehat{\cal A}_K(\cdot)$. As a result, our results hold subject to a permutation of $\widehat{\cal A}_1(\cdot), \ldots, \widehat{\cal A}_K(\cdot)$. In this paper, we follow \cite{ke2022using} in omitting this permutation throughout, as it has minimal impact on the proof while significantly simplifying the notation.

\subsubsection{Proof of \cref{lem:SigmaA}} \label{proof:lem-SigmaA}

Since $K$ is finite, $\lVert \Sigma_{A}^{\textup{net}}-\Sigma_{\mathcal{A}} \rVert\leq C\lVert \Sigma_{A}^{\textup{net}}-\Sigma_{\mathcal{A}} \rVert_{\max}$. It suffices to bound the max norm 
\[
\lVert \Sigma_{A}^{\textup{net}}-\Sigma_{\mathcal{A}} \rVert_{\max}:=\max_{1\leq k,\ell\leq K}\lvert \Sigma_{\mathcal{A}}^{\textup{net}}(k,\ell)-\Sigma_{\mathcal{A}}(k,\ell) \rvert.
\]
As a reminder, $\Sigma_{\mathcal{A}}^{\textup{net}}$ and $\Sigma_{\mathcal{A}}$ are defined by
\begin{equation*}
    \Sigma_{A}^{\text{net}}(k,\ell)=\sum_{m=1}^{M}\frac{\int_{\mathcal{R}_{m}}\mathcal{A}_{k}(z)dz\ \int_{\mathcal{R}_{m}}\mathcal{A}_{\ell}(z)dz}{\int_{\mathcal{R}_{m}}h(z)dz},\quad \Sigma_{\mathcal{A}}(k,\ell)=\sum_{m=1}^{M}\int_{\mathcal{R}_{m}}\frac{\mathcal{A}_{k}(z)\mathcal{A}_{\ell}(z)}{h(z)}dz,
\end{equation*}
where $h(z)=\sum_{k=1}^{K}\mathcal{A}_{k}(z)$. We derive that
\begin{align} \label{SigmaA-0}
    &\bigl\lvert \Sigma_{A}^{\text{net}}(k,\ell)-\Sigma_{\mathcal{A}}(k,\ell) \bigr\rvert
    = \sum_{m=1}^{M}\int_{\mathcal{R}_{m}}\mathcal{A}_{\ell}(z)\biggl\lvert\frac{\int_{\mathcal{R}_{m}}\mathcal{A}_{k}(\zeta)d\zeta}{\int_{\mathcal{R}_{m}}h(\zeta)d\zeta}-\frac{\mathcal{A}_{k}(z)}{h(z)}\biggr\rvert dz\cr
    =& \sum_{m=1}^{M}\frac{1}{\int_{\mathcal{R}_{m}}h(\zeta)d\zeta}\int_{\mathcal{R}_{m}^{2}}\frac{\mathcal{A}_{\ell}(z)}{h(z)}\ \bigl\lvert\mathcal{A}_{k}(\zeta)h(z)-\mathcal{A}_{k}(z)h(\zeta)\bigr\rvert d\zeta dz\cr
    \leq & \sum_{m=1}^{M}\frac{1}{\int_{\mathcal{R}_{m}}h(\zeta)d\zeta}\int_{\mathcal{R}_{m}^{2}}\frac{\mathcal{A}_{\ell}(z)}{h(z)}\ \Bigl[\bigl\lvert\mathcal{A}_{k}(\zeta)-\mathcal{A}_{k}(z)\bigr\rvert h(z) + \mathcal{A}_{k}(z)\bigl\lvert h(z)-h(\zeta)\bigr\rvert\Bigr] d\zeta dz.
\end{align}
We now utilize the \( \beta \)-Hölder smoothness of ${\cal A}_j(z)$. 
When $\beta\leq 1$, this implies that $|{\cal A}_j(x)-{\cal A}_j(z)|\leq \tilde{C}_1|x-z|^{\beta}$ for $x,z\in \mathrm{Int}({\cal Z}_k)$ and a constant $\tilde{C}_1>0$. When $\beta >1$,  the \( \beta \)-Hölder smoothness implies that ${\cal A}_j(z)$ has a continuous first derivative in the interior of ${\cal Z}$. Since the closure of ${\cal Z}$ is compact, the first derivative of  ${\cal A}_k(z)$ is bounded. It implies that $|{\cal A}_k(x)-{\cal A}_k(z)|\leq \tilde{C}_2|x-z|$ for a constant $\tilde{C}_2>0$. We combine these results to obtain: There is a constant $\tilde{c}_0$ such that for all $1\leq j\leq K$, 
\[
|{\cal A}_j(x)-{\cal A}_j(z)|\leq \tilde{c}_0|x-z|^{\beta \wedge 1}, \qquad \mbox{for any $x,z\in {\cal Z}$}. 
\]
Recalling that  $h(z)=\sum_{k=1}^{K}\mathcal{A}_{k}(z)$, the above inequality further implies that
\[
|h(x)-h(z)|\leq K\tilde{c}_0|x-z|^{\beta\wedge 1}, \qquad\mbox{for any }x, z\in {\cal Z}. 
\]
We plug these inequalities into \eqref{SigmaA-0} to derive that
\begin{align}\label{SigmaA-1}
    &\bigl\lvert \Sigma_{A}^{\text{net}}(k,\ell)-\Sigma_{\mathcal{A}}(k,\ell) \bigr\rvert\cr
    \leq &\sum_{m=1}^{M}\frac{1}{\int_{\mathcal{R}_{m}}h(\zeta)d\zeta}\int_{\mathcal{R}_{m}^{2}}K\tilde{c}_o \frac{\mathcal{A}_{\ell}(z)[ \mathcal{A}_{k}(z)+ h(z)]}{h(z)}\bigl[\operatorname{diam} (\mathcal{R}_{m})\bigr]^{\beta\wedge 1} d\zeta dz\cr
    \leq &\sum_{m=1}^{M}\frac{1}{\int_{\mathcal{R}_{m}}h(\zeta)d\zeta}\int_{\mathcal{R}_{m}^{2}}2K\tilde{c}_0 \mathcal{A}_{\ell}(z)\bigl[\operatorname{diam} (\mathcal{R}_{m})\bigr]^{\beta\wedge 1} d\zeta dz\cr
    = & \sum_{m=1}^{M}2 K\tilde{c}_0 \operatorname{Vol}(\mathcal{R}_{m})\ \bigl[\operatorname{diam}(\mathcal{R}_{m})\bigr]^{\beta\wedge 1}\frac{\int_{\mathcal{R}_{m}}\mathcal{A}_{\ell}(z)dz}{\int_{\mathcal{R}_{m}}h(z)dz}\cr
    \leq &C \operatorname{Vol} (\mathcal{Z}) \max_{1\leq m\leq M}\bigl[\operatorname{diam}(\mathcal{R}_{m})\bigr]^{\beta\wedge 1},
\end{align}
where in the second line and the last line, we have used $\max_k {\cal A}_k(z)\leq h(z)$. The claim follows by noticing that $\mathrm{Vol}({\cal Z})$ is upper bounded by a constant. 
\qed

\color{black}

\subsection{Proof of Theorem~\ref{thm:UB}}

Combining the claims in Lemma~\ref{lem:var-new}
and Corollary~\ref{cor:bias}, we find that with probability $1-o(n^{-2})$, for all $1\leq k\leq K$, 
\[
\max_{z\in {\cal Z}_{\delta_n}}|\widehat{\cal A}_k(z)- {\cal A}_k(z)|\leq C\Biggl( \epsilon^{\theta_d(\beta)} + \sqrt{\frac{\log(n)}{Nn\epsilon^d}}\Biggr). 
\]
The bound on the right hand side is minimized when $\epsilon^{\theta_d(\beta)}=\sqrt{\frac{\log(n)}{Nn\epsilon^d}}$; and for this choice of $\epsilon$, 
\[
\max_{z\in {\cal Z}_{\delta_n}}|\widehat{\cal A}_k(z)- {\cal A}_k(z)|\leq C \biggl(\frac{\log(n)}{Nn}\biggr)^{\frac{\theta_d(\beta)}{2\theta_d(\beta)+d}}. 
\]
Using the expression of $\theta_d(\beta)$ in Corollary~\ref{cor:bias}, we can easily see that the exponent is equal to $\theta_d^*(\beta)$ defined in the statement of this theorem. This proves the claim. \qed

\subsection{Proof of Theorem~\ref{thm:LB}} \label{subsec:LB-proof}

For brevity, we use ${\cal A}$ to represent ${\cal A}_1(\cdot), \ldots, {\cal A}_K(\cdot)$; and for 
any constants $c_1$-$c_5$, we use $\Phi_{n,N}^\star :=\Phi_{n,N}^\star(K,d,c_1,c_2,c_3,c_4,c_5)$ to denote the collection of all $({\cal A}, W)$ that satisfy 
 Assumptions \ref{assump1}-\ref{assump3}. Let 
 \[
 {\cal L}({\cal A}, \widetilde{\cal A})= \sum_{k=1}^K \int_{z\in {\cal Z}}|{\cal A}_k(z)-\widetilde{A}_k(z)|dz. 
 \]
 Our proof relies on a well-known technical tool for lower bound analysis: 
\begin{lemma}[Theorem 2.5 of \cite{tsybakov2009introduction}]
\label{lemma:textbookthm2.5-general-d}
    Suppose for an integer $J\geq 1$ and a scalar $\gamma\in (0,1/2)$, there exist  $(\mathcal{A}^{(s)},W^{(s)})\in \Phi_{n,N}^\star , 0\leq s\leq J$, such that the following conditions are satisfied:
    \begin{itemize}
        \item $\mathcal{L}(\mathcal{A}^{(s)}, \mathcal{A}^{(r)}) \geq 2C_0\left(\frac{1}{nN}\right)^{\frac{\beta}{2\beta+d}}$ for every $0\leq  s\neq r\leq J$;

        \item $KL(\mathcal{P}_s, \mathcal{P}_0) \leq \gamma \log (J)$ for every $1\leq s\leq (J)$, where $\mathcal{P}_s$ denotes the probability measure associated with $(\mathcal{A}^{(s)},W^{(s)})$ and $KL(\cdot,\cdot)$ denotes the Kullback–Leibler divergence. 
    \end{itemize}
    Then,  
    \begin{equation}
    \label{eq:thm-LB-ineq}
        \inf_{\widehat{\mathcal{A}} }\sup_{(\mathcal{A},W)\in\Phi^\star_{n,N}} \mathbb{P}\left(\mathcal{L}(\widehat{\mathcal{A}},\mathcal{A})\geq C_0\left(\frac{1}{nN}\right)^{\frac{\beta}{2\beta+d}} \right) \geq \frac{\sqrt{J}}{1+\sqrt{J}}\left(1-2\gamma-\sqrt{\frac{2\gamma}{\log J}}\right)
    \end{equation}
\end{lemma}

\noindent
Using Lemma~\ref{lemma:textbookthm2.5-general-d}, we divide our proofs into two steps:
\begin{itemize}
    \item Step 1: For some $J=J_n$ that tends to $\infty$ as $n\to\infty$, we construct $({\cal A}^{(s)}, W^{(s)})$, for $0\leq s\leq J$, such that they all belong to the parameter class $\Phi^{\star}_{n,N}$.
\item Step 2: We show that conditions in Lemma~\ref{lemma:textbookthm2.5-general-d} are satisfied. Therefore, \eqref{eq:thm-LB-ineq} holds. Since $J\to\infty$, the right hand sight is lower bounded by a constant. This gives the desirable claim.
\end{itemize}
In what follows, we present the two steps separately.

\bigskip

\noindent 
{\bf Step 1}: Construction of $({\cal A}^{(s)}, W^{(s)})$, for $0\leq s\leq J$. 

First we construct ${\cal A}^{(0)}$ and $W^{(0)}$. For brevity, we denote $\mathcal{A}^{(0)}=\mathcal{A}, W^{(0)}=W$, as long as there is no ambiguity. From the requirements in Assumptions~\ref{assump1}-\ref{assump3}, $W$ has to satisfy that $\lambda_{\min}(\Sigma)$ is lower bounded by a constant. Hence, we set $W$ as follows:  
\begin{equation} \label{LB-construct-W0}
    w_i = e_k, \quad \text{for all $1\leq k\leq K$ and $(k-1)\frac{n}{K}<i\leq k\frac{n}{K}$}. 
\end{equation}

To obtain ${\cal A}$, 
we first construct $1$-dimensional functions ${\cal A}_1(z), \ldots, {\cal A}_K(z)$ and then extend them to $d>1$ by a product form: 
\begin{equation} \label{LB-construct-prod}
    \mathcal A_k^{\text{prod}}(z): = \prod_{r=1}^d {\cal A}_k(z_k), \qquad \mbox{for }z=(z_1,\ldots, z_d)'. 
\end{equation}

It suffices to focus on $d=1$. 
The requirements on ${\cal A}$ imply: (a) each ${\cal A}_k$ is $\beta$-H{\"o}lder smooth on ${\cal Z}$; (b) the support of each ${\cal A}_k$ must be a strict subset of ${\cal Z}$ (this is needed for the existence of anchor region); (c) $h(z)=\sum_{k=1}^K {\cal A}_k(z)$ is uniformly lower bounded by a constant. In order to satisfy these three requirements, we utilize the bump function and the half-bump function:  
\beq \label{eq:LB-bump-function}
\Psi_0(z)=\exp\Bigl(\frac{1}{z^2-1}\Bigr)\cdot 1\{ -1<z<1\}, \qquad \Psi(z) = \Psi_0(z)\cdot 1\{0\leq z<1\}. 
\eeq
The bump function $\Psi_0(z)$ has a support $(-1,1)$ and is infinitely-order smooth on $\mathbb{R}$. The half-bump function $\Psi(z)$ has a support $[0,1)$ and is infinitely-order smooth on $[0,\infty)$.

\spacingset{1}
\begin{figure}[tb!]
    \centering
    \begin{subfigure}[b]{0.48\textwidth}
        \centering
        \includegraphics[width=\textwidth]{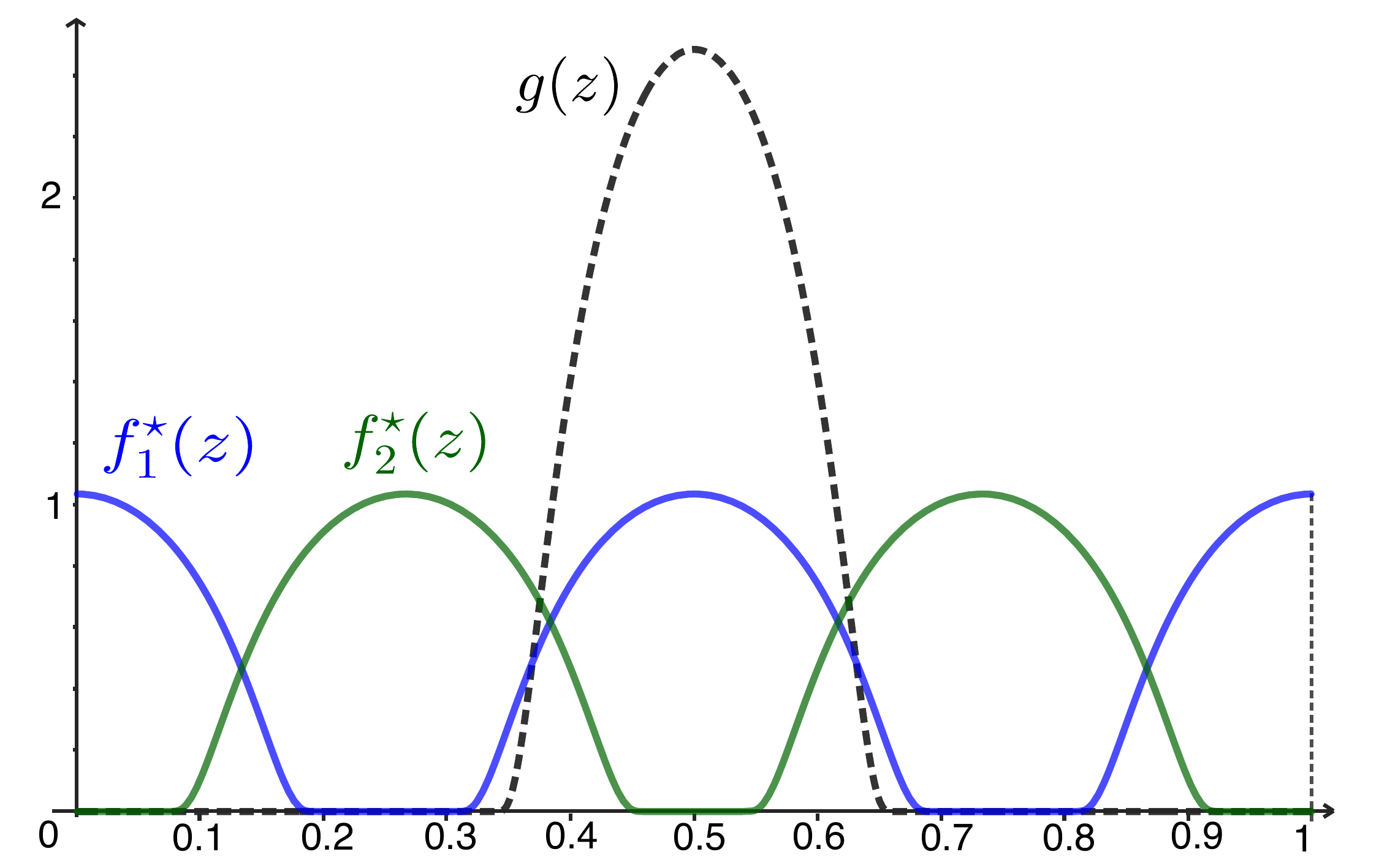} 
        \caption{Smooth base functions $f^{\star}_1(z), f^{\star}_2(z), g(z)$; the union of supports covers the whole interval $[0,1]$.}
    \end{subfigure}
    \hfill
        \begin{subfigure}[b]{0.48\textwidth}
        \centering
        \includegraphics[width=\textwidth]{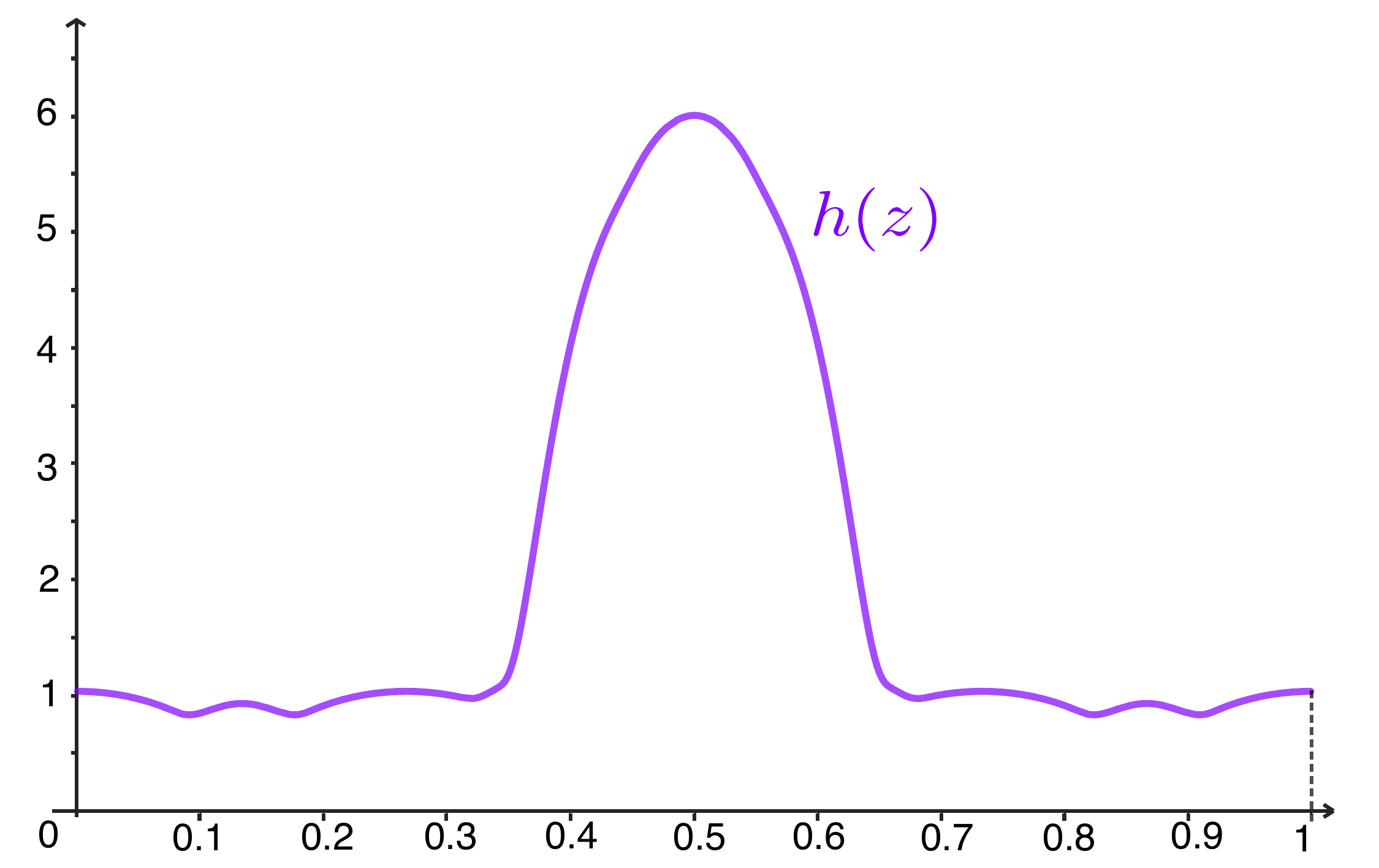} 
        \caption{Sum of densities $h(z)={\cal A}_1(z)+{\cal A}_2(z)$; it is uniformly lower bounded.}
    \end{subfigure}
    \vskip\baselineskip
        \begin{subfigure}[b]{0.48\textwidth}
        \centering
        \includegraphics[width=\textwidth]{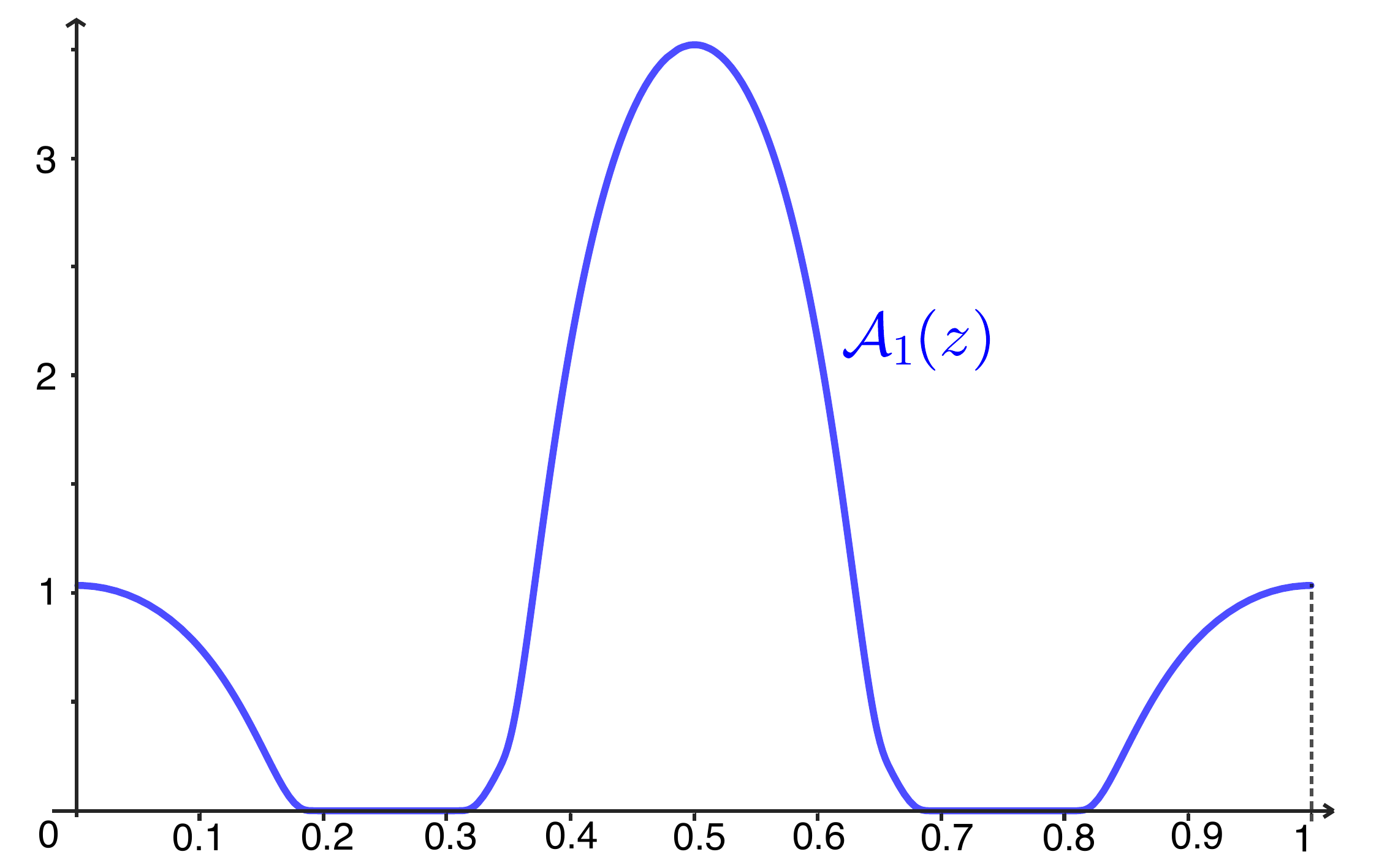} 
        \caption{Density function ${\cal A}_1(z)=f_{1}^{\star}(z)+g(z)$; it has anchor regions around $0$ and $1$.}
    \end{subfigure}
        \hfill
    \begin{subfigure}[b]{0.48\textwidth}
        \centering
        \includegraphics[width=\textwidth]{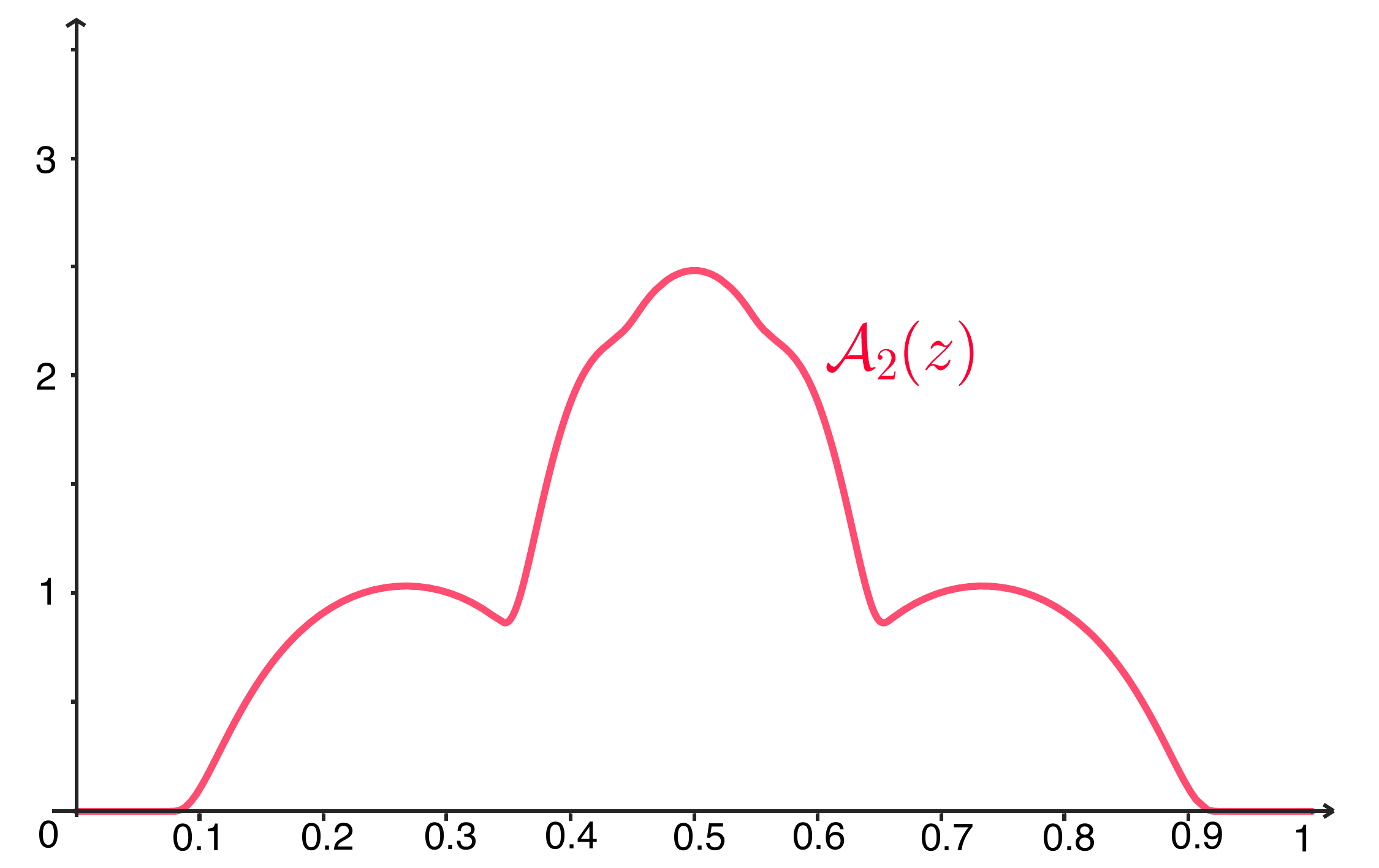} 
        \caption{Density function ${\cal A}_2(z)=f_{2}^{\star}(z)+g(z)$; it has anchor regions around $0.25$ and $0.75$.}
    \end{subfigure}
    \caption{An illustration of how we construct ${\cal A}^{(0)}$ for $d=1$ and $K=2$.}
    \label{fig:LB-example}
\end{figure}
\resetspacing

We first divide the interval ${\cal Z}=[0,1]$ into three disjoint intervals $I_1=[0,1/3)$, $ I_2=(2/3,1]$, and $I_0=[1/3,2/3]$. We then further divide $I_1$ into $2K-1$ disjoint intervals: $I_1=\cup_{k=1}^{2K-1}Q_k$, 
where $Q_1 = \bigl[0,\frac{1/2}{3(2K-3/2)}\bigr)$, and $Q_i = \bigl[\frac{i-3/2}{3(2K-3/2)},\frac{i-1/2}{3(2K-3/2)}\bigr)$, for $2\leq i\leq 2K$. Under these definitions, $Q_1$ has half of the length of other $Q_k$; $I_1=\cup_{i=1}^{2K-1}Q_i$, and $Q_{2K}\subset I_0$).
Now, 
for all $2\leq k\leq K$, let
\[
f_k(z) = C_1 l_1^\beta\Psi_0\Bigl(\frac{z-x_k}{l_1}\Bigr),\qquad  \mbox{where}\;
\begin{cases}
 \mbox{$x_k$ is the center of $P_k:= Q_{2k-2}\cup Q_{2k-1}\cup Q_{2k}$},\cr
 \ell_1=\frac{1}{4K-3} \mbox{ (which is the half length of $P_k$)}. 
\end{cases} 
\]
Since $\ell_1$ is a finite constant, each $f_k(z)$ is a function with a compact support on $P_k$ and is infinitely-order smooth on $\mathbb{R}$ (which is also $\beta$-H{\"o}lder smooth on ${\cal Z}$). Additionally, on each $Q_{2k-1}$, only $f_k(z)$ is nonzero, with all other $f_\ell(z)$ being zero (this will be useful for creating anchor regions in our construction of ${\cal A}_k(z)$ below). We also construct $f_1(z)$ by 
\[
f_1(z) = C_1\ell_1^\beta \Psi\Bigl( \frac{z}{\ell_1}\Bigr) + C_1\ell_1^\beta \Psi\Bigl( \frac{1/2-z}{\ell_1}\Bigr), 
\qquad  \mbox{where $\Psi(z)$ is the half-bump function in \eqref{eq:LB-bump-function}}. 
\]
Here, $f_1(z)$ is the union of two half-bump functions, one supported on $P_1=Q_1\cup Q_2$, and the other supported on $[1/2-\ell_1, 1/2]$. We find that $f_1(z)$ is infinitely-order smooth on $[0, 1/2]$ (but it is not smooth on $\mathbb{R}$), and in $Q_1$, only $f_1(z)$ is nonzero, with the other $f_k(z)$ being zero.
The reason of using the half-bump function, instead of the bump function, is to guarantee that $f_1(z)$ is lower bounded in a small half neighborhood of $0$ (this is crucial to guarantee that $h(z)$ is uniformly lower bounded under our construction below). Finally, let $\ell_2=1/6$ be the half length of $I_0$ and define:
\[
g(z) = C_2\ell_2^{\beta}\Psi_0\Bigl( \frac{z-1/2}{\ell_2}\Bigr), \qquad 
f_{k}^{\star}(z) = f_k(z)+f_k(1-z), \;\; \mbox{for }1\leq k\leq K, 
\]
Here, $g(z)$ is smooth on $\mathbb{R}$ and has a support on $[1/3,2/3]$; and $f_k(1-z)$ is a ``mirror'' of $f_k(z)$ from $I_1$ to $I_2$ (consequently, $f_{k}^{\star}(z)$ is symmetric with respect to $1/2$; we emphasize that the two half bumps in $f_1(z)$ combine into a single bump centering at $1/2$). 
The purpose of introducing $g(z)$ and mirroring $f_k(z)$ is to ensure that for each $z\in [0,1]$, at least one of $g(z), f_1(z),\ldots, f_K(z)$ is nonzero. 
We still need to choose the constants $C_1$ and $C_2$. We choose them such that 
\[
\int_0^1 f_{k}^{\star}(z)dz=1/2, \qquad  \int_0^1 g(z)dz=1/2.
\]
The functions, $g(\cdot), f_1(\cdot), \ldots, f_K(\cdot)$, are the baseline functions we use to construct ${\cal A}$. In a special case with $K=2$ and $d=1$, they are illustrated in Figure~\ref{fig:LB-example}(a). We now construct
\beq \label{LB-construct-A0}
{\cal A}_k(z) = g(z) + f_{k}^{\star}(z), \qquad\mbox{for }1\leq k\leq K. 
\eeq
In the special case of $K=2$ and $d=1$, the constructed ${\cal A}$ is plotted in Figure~\ref{fig:LB-example}(b).

Next, for some properly chosen $J=J_n\to\infty$, we construct $(\mathcal{A}^{(s)},W^{(s)})$, for $1\leq s\leq J$. For $W^{(s)}$, we simply let all of them equal to $W^{(0)}$ in \eqref{LB-construct-W0}. To construct ${\cal A}^{(s)}$, let 
\beq \label{LB-m-choise}
m=\left\lceil c_0{(nN)}^{\frac{1}{2\beta+d}}\right\rceil, \quad h_m=1/m, \quad x_j = \frac{2j-m-1}{m},\quad \mbox{for }1\leq j\leq m,
\eeq
for some constant $c_0>0$ to be determined. 
Essentially we divide $[-1,1]$ into $m$ disjoint intervals, where $h_m$ is the half length of each interval, and $x_1, \ldots, x_m$ are the center points of these intervals.
Applying the same partition to each coordinate of $[-1,1]^d$, we obtain a total of $m^d$ disjoint cubes. 
We use  $\vec{j}=(j_1,j_2,\ldots,j_d)'\in \{1,2,\ldots, m\}^d$ to index each such cube: For the cube associated with $\vec{j}$, its center point is located at $(x_{j_1}, x_{j_2},\ldots, x_{j_d})'$. We then create  a total of $m^d$ different multi-variate bump functions, one for each cube. These functions are indexed by $\vec{j}$ and defined as 
\beq \label{eq:LB-varphi-j}
\varphi_{\vec{j}}(z) = \varphi_{j_1,j_2,\cdots,j_d}(z) = h_m^{\beta} \prod_{r=1}^d 
\Psi_0\left(\frac{z_r-x_{j_r}}{h_m}\right). 
\eeq
where $\Psi_0(\cdot)$ is the uni-variate bump function in \eqref{eq:LB-bump-function}. Since each $\varphi_{\vec{j}}(z)$ is a product of $d$ uni-variate functions, we can calculate the derivatives explicitly. It suggests that these functions are $\beta$-H{\"o}lder smooth (but they are not infinitely-order smooth, as $h_m$ is not a constant). Each $\varphi_{\vec{j}}(z)$ is supported on the cube indexed by $\vec{j}$, so these bump functions have non-overlapping supports (this fact will be used in Step 2 of this proof). 
We further define:
\begin{equation}
\label{eq:LB-phi-j}
    \phi_{\vec{j}}(z) := \phi^+_{\vec{j}}(z) - \phi^-_{\vec{j}}(z), 
    \qquad \mbox{where}\;\; 
    \begin{cases}
        \phi^+_{\vec{j}}(z) = \frac{1}{16^{\beta}}\varphi_{\vec{j}}\Bigl(16 \bigl( z-\frac{7}{16} {\bf 1}_d\bigr)\Bigr),\cr
        \phi^-_{\vec{j}}(z) = \frac{1}{16^{\beta}}\varphi_{\vec{j}}\Bigl(16 \bigl( z-\frac{9}{16} {\bf 1}_d\bigr)\Bigr). 
    \end{cases}
\end{equation}
From $\varphi_{\vec{j}}(z)$ 
to $\phi_{\vec{j}}(z)$, we have made two changes: First, we create ``shifted and re-scaled versions" of $\varphi_{\vec{j}}(z)$, denoted by $\phi^{+}_{\vec{j}}(z)$ and $\phi^{-}_{\vec{j}}(z)$: The support of $\varphi_{\vec{j}}(z)$ is $[-1, 1]^d$. In comparison, the support of $\phi^+_{\vec{j}}(z)$ is $[\frac{3}{8}, \frac{1}{2}]^d$, and the support of $\phi^{-}_{\vec{j}}(z)$ is $[\frac{1}{2}, \frac{5}{8}]^{d}$. 
Second, we flip the sign on $\phi^-_{\vec{j}}(z)$ and add it to $\phi^+_{\vec{j}}(z)$. This gives $\phi_{\vec{j}}(z)$. 
With this construction, the functions $\phi_{\vec{j}}(z)$ satisfy the following properties: 
\begin{itemize}
    \item[(a)] Each $\phi_{\vec{j}}(z)$ is a union of two bumps with opposite signs. Hence, $\int_{[0,1]^d} \phi_{\vec{j}}(z)dz=0$. 
    \item[(b)] The $2m^d$ bumps (half of them are positive, half negative) have non-overlapping supports. 
    \item[(c)] The support of each $\phi_{\vec{j}}(z)$ is a subset of $[\frac{3}{8}, \frac{5}{8}]^d$, which is in the interior of $g(z)$ defined above. By \eqref{LB-construct-A0},  each ${\cal A}_k^{(0)}(z)$ is lower bounded by a constant in $[\frac{3}{8}, \frac{5}{8}]^d$. Therefore, in the support of each $\phi_{\vec{j}}(z)$, $\min_{1\leq k\leq K}{\cal A}_k^{(0)}(z)$ is lower bounded by a constant. 
 \end{itemize}
 
We now  use these functions $\phi_{\vec{j}}(z)$ to ``perturb" each ${\cal A}_k^{(0)}(z)$ to obtain ${\cal A}_k^{(s)}(z)$. 
%
%
The following lemma is a well-known result (e.g., see Lemma 2.9 of \cite{tsybakov2009introduction} for a proof):

\begin{lemma}[Varshamov–Gilbert]
\label{lemma:packing-number-for-m-general-d}
For any integer $A\geq 8$, there exists an integer $J\geq 2^{A/8}$, and vectors $\omega^{(0)}, \omega^{(1)},  \cdots, \omega^{(J)} \in \{0,1\}^{A}$ such that $\omega^{(0)}$ is a zero vector and that  $
        \|\omega^{(s)}- \omega^{(t)}\|_1 \geq A/8$ for all $0\leq s\neq t\leq J$. 
\end{lemma}

Recall that there are a total of $m^d$ different vectors $\vec{j}$, and we index them as $\vec{j}_1,\vec{j}_2,\ldots, \vec{j}_{m^d}$.
We apply Lemma~\ref{lemma:packing-number-for-m-general-d} with $A=m^d\geq c_0^d (nN)^{\frac{d}{2\beta+d}}$. 
Let $\omega^{(0)}, \omega^{(1)}, \ldots, \omega^{(J)}$ be the vector sequence in Lemma~\ref{lemma:packing-number-for-m-general-d}, where $J\geq 2^{m^d/8}\to\infty$. For each $1\leq s\leq J$ and $1\leq j\leq m^d$, let $\omega^{(s)}_j$ denote the $j$th entry of $\omega^{(s)}$.   We construct ${\cal A}^{(s)}$ as follows: 
\begin{equation}
\label{eq:LB-from-A0-to-As}
    \mathcal{A}^{(s)}_k(z) = \mathcal{A}^{(0)}_k(z) + \sum_{i=1}^{m^d} \omega_i^{(s)}\phi_{\vec{j}_i}(z), \qquad 1\leq s\leq J, \; 1\leq k\leq K. 
\end{equation}
Since ${\cal A}_k^{(0)}(z)$ and $\phi_{\vec{j}}(z)$ are both $\beta$-H{\"o}lder smooth on $[0,1]^d$, 
${\cal A}_k^{(s)}$ is also $\beta$-H{\"o}lder smooth on $[0,1]^d$. Property (c) of $\phi_{\vec{j}}(z)$ ensures that each ${\cal A}_k^{(s)}(z)$ is a nonnegative function, and Property (c) ensures that ensures that each ${\cal A}_k^{(s)}(z)$ has an integral of $1$.


Finally, we verify that $({\cal A}^{(s)}, W^{(s)})$ belongs to $\Phi^{\star}_{n,N}$ for all $0\leq s\leq J$. We pack this proof into the following lemma, which is proved in Section~\ref{subsec:proof-lem-condition-for-Aj-Wj}.

\begin{lemma}
\label{lemma:condition-for-Aj-Wj-general-d}
    For any $k$, sufficiently small $c_2$-$c_5$ and sufficiently large $c_1$, we have $(\mathcal{A}^{(s)},W^{(s)})\in \Phi^\star_{n,N}(K,d,c_1,c_2,c_3,c_4,c_5)$ for every $1\leq s\leq J$.
\end{lemma}

 \bigskip

 \noindent
{\bf Step 2}: Verification of the two conditions in Lemma~\ref{lemma:textbookthm2.5-general-d}.

First, we verify the first condition which requires a lower bound for $\mathcal{L}(\mathcal{A}^{(s)},\mathcal{A}^{(t)})$. By \eqref{eq:LB-from-A0-to-As}, for any $s\neq t$, 
\beq
\label{eq:separat-omega-phi}
    |{\cal A}_k^{(s)}(z)-{\cal A}_k^{(t)}(z)|=\biggl| \sum_{i=1}^{m^d} (\omega_i^{(s)}-\omega_i^{(t)})\cdot\phi_{\vec{j_i}}(z)\biggr| 
    = \sum_{i=1}^{m^d} |\omega_i^{(s)}-\omega_i^{(t)}|\Bigl(|\phi^+_{\vec{j}_i}(z)| + \phi^-_{\vec{j}_i}(z)| \Bigr).
\eeq
Here, the second equality follows from Property (b) of $\phi_{\vec{j}}(z)$. The integrals of all $2m^d$ bump functions, $\{ \phi^{\pm}_{\vec{j}_i}(z)\}_{i=1}^{m^d}$, are the same.   
Combining this with \eqref{eq:separat-omega-phi}, we obtain that 
\begin{align} \label{eq:LB-add1}
    \mathcal{L}(\mathcal{A}^{(s)},\mathcal{A}^{(t)}) &=
    \sum_{k=1}^K \int_{z\in {\cal Z}} |{\cal A}_k^{(s)}(z)- {\cal A}_k^{(k)}(z)|dz\cr
    &
    = \sum_{k=1}^K \sum_{i=1}^{m^d}|\omega_i^{(s)}-\omega_i^{(d)}|\left(\int_{z\in {\cal Z}} \phi^+_{\vec{j}_i}(z)dz + \int_{z\in {\cal Z}} \phi^+_{\vec{j}_i}(z)dz \right) \cr
    &= 2K \|\omega^{(s)}-\omega^{(t)}\|_1 \int_{z\in {\cal Z}} \phi_{\vec{j}_1}^+(z) dz. 
\end{align}
We then calculate $\int_{z\in {\cal Z}} \phi_{\vec{j}_1}(z)dz$. For any $b, c>0$, define $\Psi_{0}(z; b, c)=\frac{1}{b}\Psi_0(\frac{z-c}{b})$, where $\Psi_0(z)$ is the uni-variate bump function in \eqref{eq:LB-bump-function}. 
Combining this notation with \eqref{eq:LB-varphi-j}-\eqref{eq:LB-phi-j}, we find that 
\[
\phi^{+}_{\vec{j}}(z) =\frac{h_m^{\beta+d}}{16^{\beta+d}} \prod_{r=1}^d \Psi_{0}\Bigl(z_r;\, \frac{h_m}{16}, \,    \frac{7 + h_m x_{j_r}}{16}\Bigr).  
\]
%
The function $\Psi_0(z;b,c)$ has a property: When $[c-b, c+b] \subseteq [0,1]$, it holds that $\int_0^1\Psi_0(z; b, c)dz = \int_{c-b}^{c+b}\Psi_0(z; b,c)dz =\int_{-1}^1 \Psi_0(z)dz:=\|\Psi_0\|_1$ (the first equality is due to that the support of $\Psi_0(z; b,c)$ is  $[c-b, c+b]$, and the second equality is from a change of variable). As a result, 
%
\beq \label{eq:LB-add2}
\int_{z\in {\cal Z}}\phi^{+}_{\vec{j_1}}(z)dz = \frac{h_m^{\beta+d}}{16^{\beta+d}}\prod_{r=1}^d \left(\int_{0}^1 \Psi_0(z_r, b,c)dz_r\right) \Biggl|_{(b, c)=\bigl(\frac{h_m}{16},\frac{7 + h_mx_{j_1}}{16}\bigr)}=\frac{h_m^{\beta + d}}{16^{\beta + d}}\|\Psi_0\|_1^d.  
\eeq
Combining \eqref{eq:LB-add1} with \eqref{eq:LB-add2} gives:
\[
\mathcal{L}(\mathcal{A}^{(s)},\mathcal{A}^{(t)})\geq \frac{2K\|\Psi_0\|_1^d }{16^{\beta + d}}\cdot h_m^{\beta + d}\, \|\omega^{(s)}-\omega^{(t)}\|_1. 
\]
By Lemma~\ref{lemma:packing-number-for-m-general-d}, $\|\omega^{(s)}-\omega^{(d)}\|_1\geq m^d/8$. Additionally, $h_m=1/m$. It follows that
\beq \label{eq:LB-add3}
\mathcal{L}(\mathcal{A}^{(s)},\mathcal{A}^{(t)})\; \geq\;  \frac{2K\|\Psi_0\|_1^d }{8\times 16^{\beta  + d}} \cdot h_m^{\beta+d}m^{d}\; \geq\;  \underbrace{\frac{2K\|\Psi_0\|_1^d }{8\times 16^{\beta + d}\times c_0^{\beta}}}_{:= 2C_0}\cdot \Bigl( \frac{1}{nN}\Bigr)^{\frac{\beta}{2\beta + d}}. 
\eeq
This proves the first condition of Lemma~\ref{lemma:textbookthm2.5-general-d}.

Next, we check the second condition in \cref{lemma:textbookthm2.5-general-d}. 
For any two Poisson processes, their KL-divergence is given in the following lemma, which is also Theorem 5.2 of \cite{leskela2024information}:
\begin{lemma}[KL divergence for Poisson processes]
\label{lemma:kl-pp}
Suppose $P_\lambda$ and $P_\mu$ are Poisson processes on ${\cal Z}$ with intensity measures $\lambda(z)$ and $\mu(z)$. Then $
    KL(P_\lambda,  P_\mu) = \int_{z\in {\cal Z}} \Bigl[ \lambda(z) \log\bigl(\frac{\lambda(z)}{\mu(z)}\bigr) + \lambda(z) - \mu(z)\Bigr] dz$. 
    In particular, if $\int_{z\in {\cal Z}}\lambda (z)dz=\int_{z\in {\cal Z}}\mu (z)dz$, then $KL(P_\lambda, P_\mu) = \int_{z\in {\cal Z}} \lambda (z) \log\bigl(\frac{\lambda(z)}{\mu (z)}\bigr)dz$. 
\end{lemma}

In our problem, $\mathcal{P}_s$ is the probability measure given by $n$ independent Poisson processes, with intensity measures $N\Omega_i^{(s)}(\cdot)$, for $1\leq i\leq n$. 
Moreover, $\int_{z\in {\cal Z}}N\Omega_i^{(s)}(z)= \int_{z\in {\cal Z}}N\Omega_i^{(0)}(z)$. It gives: 
\begin{align} \label{eq:LB-add4}
KL({\cal P}_s, {\cal P}_0) &= \sum_{i=1}^n \int_{z\in {\cal Z}} N \Omega_i^{(s)}(z)\log\Biggl( \frac{\Omega_i^{(s)}(z)}{\Omega_i^{(0)}(z)}\Biggr) dz\cr
&= \sum_{i=1}^n \int_{z\in [3/8,\, 5/8]^d} N \Omega_i^{(s)}(z)\log\Biggl( \frac{\Omega_i^{(s)}(z)}{\Omega_i^{(0)}(z)}\Biggr)dz,
\end{align}
where the first line comes from independence across $n$ Poisson processes, and the second line is due to that $\Omega_i^{(s)}(\cdot)$ differs from $\Omega_i^{(0)}(\cdot)$ only in the region ${\mathcal C}_0 :=[3/8,\, 5/8]^d$ (this has been explained in Property (c) of $\phi_{\vec{j}}(z)$). 

We fix $1\leq s\leq J$. Write $f_i(z)=\Omega^{(0)}_i(z)$ and $\delta_i(z)=\frac{\Omega_i^{(s)}(z)-\Omega_i^{(0)}(z)}{f_i(z)}$ for short. Property (c) of $\phi_{\vec{j}}(z)$ has guaranteed that $\min_{k}{\cal A}_k^{(s)}(z)$ is uniformly lower bounded in ${\cal C}_0$. It follows that 
\beq \label{eq:LB-add5}
\inf_{z\in {\cal C}_0}\{f_i(z)\} = \inf_{z\in {\cal C}_0 } \biggl\{ \sum_{k=1}^K \pi_i(k){\cal A}_k^{(0)}(z)\biggr\} \geq \inf_{z\in {\cal C}_0} \Bigl\{ \min_{k}{\cal A}_k^{(0)}(z)\Bigr\} \geq \tilde{c}, 
\eeq
for some constant $\tilde{c}>0$. Consequently, 
\beq \label{eq:LB-add6}
\sup_{z\in {\cal C}_0}\{|\delta_i(z)|\} \leq \tilde{c}^{-1} \sup_{z\in {\cal C}_0} \biggl\{ \sum_{k=1}^K \pi_i(k)\bigl| {\cal A}_k^{(s)}(z) - {\cal A}_k^{(0)}(z)\bigr|\biggr\} \leq Ch_m^{\beta}, 
\eeq
where the last inequality is a consequence of the definition of ${\cal A}_k^{(s)}(z)$ in \eqref{eq:LB-from-A0-to-As}, Property (b) of $\phi_{\vec{j}}(z)$ (i.e., these functions have non-overlapping supports), and the explicit expression of bump functions in \eqref{eq:LB-varphi-j}-\eqref{eq:LB-phi-j}. Using \eqref{eq:LB-add5}-\eqref{eq:LB-add6}, we find that 
\begin{align} \label{eq:LB-add7}
\Omega_i^{(s)}(z)\log\Biggl( \frac{\Omega_i^{(s)}(z)}{\Omega_i^{(0)}(z)}\Biggr) &= f_i(z)[1+\delta_i(z)]\cdot \log(1+ \delta_i(z))\cr
&\leq f_i(z)[1+\delta_i(z)]\Bigl[\delta_i(z)-\delta^2_i(z)/2 + \widetilde{C}|\delta_i(z)|^3 \Bigr]\cr
&\leq f_i(z)\Bigl[ \delta_i(z) +\delta_i^2(z)/2  + C|\delta_i(z)|^3 \Bigr]. 
\end{align}
Here, the second line is from the Taylor expansion of $\log(1+t)$. Specifically, $\log(1+t)=t-\frac{t^2}{2!}+\frac{2 t^3}{3!(1+t_*)^3}$, for some $t^*$ between $0$ and $t$; when $t$ is in a small neighborhood of $0$, the last term is upper bounded by $\widetilde{C}|t|^3$, for a universal constant $\widetilde{C}>0$. 
We plug \eqref{eq:LB-add7} into \eqref{eq:LB-add4}. It follows that 
\begin{align*}
    KL({\cal P}_s, {\cal P}_0)\leq N \sum_{i=1}^n \int_{z\in {\cal C}_0} f_i(z)\Bigl[ \delta_i(z) +\delta_i^2(z)/2  + C|\delta_i(z)|^3 \Bigr]dz. 
\end{align*}
First,  $\int_{z\in {\cal C}_0}f_i(z)\delta_i(z)dz = \int_{z\in {\cal C}_0}[\Omega_i^{(s)}(z)-\Omega_i^{(0)}(z)]dz= 0$. Second, similar to \eqref{eq:LB-add5}, we can show that $|f_i(z)|$ is uniformly upper bounded by a constant. Third, $|\delta_i(z)|^3\leq |\delta_i(z)|^2$ (following \eqref{eq:LB-add6}). Combining these arguments, we obtain: 
\beq \label{eq:LB-add8}
    KL({\cal P}_s, {\cal P}_0) \leq C N\sum_{i=1}^n \int_{z\in {\cal C}_0} \delta_i^2(z)dz. 
\eeq
We now bound the right hand side of \eqref{eq:LB-add8}. 
Using the definition of $\delta_i(z)$ and the lower bound in \eqref{eq:LB-add5} for $f_i(z)$, we obtain:  
\begin{align*}
\delta_i^2(z) & \leq \tilde{c}^{-2} \bigl[ \Omega_i^{(s)}(z) -\Omega_i^{(0)}(z) \bigr]^2 \; =\;   \tilde{c}^{-2} \Biggl( \sum_{k=1}^K \pi_i(k)\Bigl[ {\cal A}_k ^{(s)}(z) - {\cal A}_k^{(0)}(z)\Bigr] \Biggr)^2 \cr
&\leq \tilde{c}^{-2}\sum_{k=1}^K \bigl[ {\cal A}_k ^{(s)}(z) - {\cal A}_k^{(0)}(z)\bigr]^2 = \tilde{c}^{-2} K \Biggl( \sum_{t=1}^{m^d} [\omega_t^{(s)}-\omega_t^{(0)}] \cdot \phi_{\vec{j}_t}(z)\Biggr)^2  \cr
& \leq \tilde{c}^{-2} K \sum_{t=1}^{m^d} [\omega_t^{(s)}-\omega_t^{(0)}]^2 \cdot \phi^2_{\vec{j}_t}(z).
\end{align*}
Here, the inequality in the second line uses the Cauchy-Schwarz inequality and that $\sum_{k=1}^K \pi_i^2(k)\leq \sum_{k=1}^K\pi_i(k)\leq 1$; and the equality in this line is from \eqref{eq:LB-from-A0-to-As}. In the last line, we have used Property (b) of $\phi_{\vec{j}}(z)$; specifically, since these bump functions  have non-overlapping supports, we can change the square of sum to the sum of squares. 
It follows from the above inequality that 
\beq \label{eq:LB-add9}
\int_{z\in {\cal C}_0}\delta_i^2(z)dz  \leq  \tilde{c}^{-2}K\cdot \|\omega^{(s)}-\omega^{(0)}\|^2   \max_{1\leq t\leq m^d} \int_{z\in {\cal C}_0}\phi^2_{\vec{j}_t}(z)dz. 
\eeq
For any $\vec{j}$, by \eqref{eq:LB-varphi-j}-\eqref{eq:LB-phi-j}, $|\phi^{\pm}_{\vec{j}}(z)|\leq (h_m/16)^{\beta}$. It follows that 
\[
\int_{z\in {\cal C}_0}\phi^2_{\vec{j}}(z)dz = 2 \int_{z\in {\cal C}_0}\bigl[\phi^{+}_{\vec{j}} (z)\bigr]^2 dz \leq  \text{Vol}(\mathcal C_0)\cdot 2\bigl[\max_{z\in \mathcal C_0}\phi^{+}_{\vec{j}} (z)\bigr]^2
\leq 
\frac{2 \text{Vol}(\mathcal C_0) }{16^{2\beta }}h_m^{2\beta } .  
\]
Combining this with \eqref{eq:LB-add9} and using that $h_m=1/m$ and $\|\omega^{(s)}-\omega^{(0)}\|^2\leq m^d$, we obtain 
\beq \label{eq:LB-add10}
\int_{z\in {\cal C}_0}\delta_i^2(z)dz  \leq  \frac{2 K \text{Vol}(\mathcal C_0) }{\tilde{c}^{2} 16^{2\beta}} \cdot m^d h_m^{2\beta}\leq  \frac{2 K \text{Vol}(\mathcal C_0) }{\tilde{c}^{2} 16^{2\beta}}\,  m^{d - 2\beta}.  
\eeq
Combining \eqref{eq:LB-add10} into \eqref{eq:LB-add8}, we find that for a constant $C^*_{\beta,d}$ that only depends on $(\beta, d)$,
\[
KL({\cal P}_s, {\cal P}_0) \leq C^{*}_{\beta, d}\cdot  Nn\cdot m^{-2\beta}. 
\]
By \eqref{LB-m-choise}, $m=\lceil c_0 (Nn)^{\frac{1}{2\beta + d}}\rceil$, which implies $Nn\leq c_0^{-(2\beta + d)} m^{2\beta+d}$. It follows that $KL({\cal P}_s, {\cal P}_0)\leq C^*_{\beta,d}c_0^{-(2\beta+d)}m^d$. 
Meanwhile, $\log(J)\geq \log(2^{m^d/8})\geq \log(2)\cdot m^d/8$. We immediately have
\beq \label{eq:LB-add11}
KL({\cal P}_s, {\cal P}_0) \leq \frac{8C^{*}_{\beta, d}}{c_0^{2\beta+d}\log(2)}\log(J):=\gamma\log(J). 
\eeq
We can choose a proper $c_0$ such that $\gamma<1/2$. This proves the second condition in Lemma~\ref{lemma:textbookthm2.5-general-d} and also completes the proof of this theorem.\qed

\subsubsection{Proof of Lemma~\ref{lemma:condition-for-Aj-Wj-general-d}}
\label{subsec:proof-lem-condition-for-Aj-Wj}

Denote the class 
\[
\Sigma_d(\beta,\gamma) := \{f\in \mathbb{R}^d\to \mathbb{R}: |f^{\lfloor\beta\rfloor}(x)-f^{\lfloor\beta\rfloor}(y)|\leq \gamma\cdot \|x-y\|^{\beta-\lfloor\beta\rfloor}\}
\]
where $\|\cdot\|$ denotes the natural norm in Euclidean space. We also use $\Sigma(\beta,\gamma)$ to represent $\Sigma_1(\beta,\gamma)$.
For Assumption \ref{assump1}, we first notice that by construction, $f_k^\star(z_r)+g(z_r) \in \Sigma(\beta, 2C_1+C_2)$ for every $1\leq r\leq d$, where $C_1$ and $C_2$ only depend on $\beta$. Since $f_k^\star(z_r)+g(z_r) $ is also upper bounded, we have $\mathcal A_k(z) \in \Sigma_d(\beta,c_1/3)$ when $c_1$ is sufficiently large. For each $\vec{j}$, when $c_1$ is sufficiently large, we know that  $\phi^+_{\vec{j}}(z)\in \Sigma_d(\beta,c_1/3)$ since it is supported on $\Delta_{\vec{j}}:=\prod_{r=1}^d \left(\frac{3+(j_r-1)/m}{8},\frac{3+j_r/m}{8}\right)$. Since $\Delta_{\vec{j}}$ are disjoint for different $\vec{j}$, we know that the perturbation $\sum_{\vec{j}} \phi^+_{\vec{j}}(z) \in \Sigma_d(\beta,c_1/3)$. Similarly, we can show that $\sum_{\vec{j}} \phi^-_{\vec{j}}(z) \in \Sigma_d(\beta,c_1/3)$ as well. Combined with \eqref{eq:LB-from-A0-to-As}, we know $\mathcal A_k^{(s)}\in \Sigma_d(\beta,c_1)$ for sufficiently large $c_1$.

By construction, the anchor region for each $\mathcal{A}_k^{(s)}$is 
\begin{equation*}
    \mathcal S_k = 
    \begin{cases}
        \left(\frac{2k-5/2}{3(2K-3/2)},\frac{2k-3/2}{3(2K-3/2)}\right)^d \cup \left(1-\frac{2k-3/2}{3(2K-3/2)},1-\frac{2k-5/2}{3(2K-3/2)}\right)^d, \quad k\geq 2,\cr
        \left(0,\frac{2k-3/2}{3(2K-3/2)}\right)^d \cup \left(1-\frac{2k-3/2}{3(2K-3/2)},1\right)^d, \quad k=1.
    \end{cases}
\end{equation*}
Therefore, the volume of each anchor region is at least $\frac{1}{3(2K-3/2)}\geq c_2$ when choosing $c_2$ to be sufficiently small.

Now for any pair $(k,l)$, we have 
\[
\Sigma_\mathcal{A}(k,l) = \int_{\mathcal{Z}}\frac{\mathcal A_k(z)\mathcal A_l(z)}{h(z)}dz.
\]
For $z\in \mathcal C_0$, we have 
\[
\mathcal A_k(z) = 
\begin{cases}
    g(z),\quad k\geq 2, \cr
    g(z)+C_1\ell_1^\beta\Psi_0\left(\frac{z-1/2}{\ell_1}\right), \quad k=1.
\end{cases}
\]
Therefore, we have 
\[
\mathcal A_k(z) \geq g(z), \forall k, \quad \sum_{k=1}^K \mathcal{A}_k(z) \leq Kg(z) + C_1\ell_1^\beta\|\Psi_0\|_{\infty}.
\]
Therefore, for sufficiently small $c_5$, 
\[
\Sigma_{\mathcal A}(k,l) \geq \int_{\mathcal C_0} \frac{\mathcal A_k(z)\mathcal A_l(z)}{h(z)}dz \geq \left(\int_{3/8}^{5/8}\frac{g(z)^2}{Kg(z)+C_1\ell_1^\beta\|\Psi_0\|_{\infty}} dz\right)^d \geq 2c_5.
\]
Also, we can rewrite $\Sigma_{\mathcal{A}}(k,l)$ as the sum
\[
\Sigma_{\mathcal{A}}(k,l) = 
\begin{cases}
    \underbrace{\int_{(0,1)^d-\cup_{k=1}^K \mathcal S_k} \frac{\mathcal{A}_k(z)}{\sqrt{h(z)}}\cdot \frac{\mathcal{A}_l(z)}{\sqrt{h(z)}} dz}_{G(k,l)}, \quad k\neq l,
    \cr
    \int_{\mathcal S_k} \frac{\mathcal{A}_k(z)}{\sqrt{h(z)}}\cdot \frac{\mathcal{A}_k(z)}{\sqrt{h(z)}} dz + \underbrace{\int_{(0,1)^d-\cup_{k=1}^K \mathcal S_k} \frac{\mathcal{A}_k(z)}{\sqrt{h(z)}}\cdot \frac{\mathcal{A}_k(z)}{\sqrt{h(z)}} dz}_{G(k,k)}, \quad k=l.
\end{cases}
\]
Therefore, for sufficiently small $c_5$, we have
\begin{equation*}
    \Sigma_\mathcal{A} \geq \text{diag}(2c_5,2c_5,\cdots,2c_5) + G \geq \text{diag}(2c_5,2c_5,\cdots,2c_5).
\end{equation*}

From the construction of $W$ we know
\begin{equation*}
    \lambda_{min}(\Sigma_W) \geq 1/K > c_4,
\end{equation*}
where $c_4$ is chosen to be sufficiently small.
Since the minimal value of $h(z)$ when $d=1$ is achieved and lower bounded, for sufficiently small $c_3$, we have $h(z)>c_3$.

Now we need to verify Assumption \ref{assump3} for each $(\mathcal{A}^{(s)},W^{(s)})$ since $W^{(s)}=W$, it is sufficient to prove that
\begin{equation*}
    ||\Sigma_{\mathcal{A}^{(s)}}-\Sigma_\mathcal{A}||_{\infty} \leq O\left(\left(\frac{1}{nN}\right)^{\frac{\beta}{2\beta+d}}\right)
\end{equation*}
For every $(k,r)$, since $\Sigma_{\mathcal{A}^{(s)}}$ and $\Sigma_\mathcal{A}$ only differ on $\mathcal C_0$, we have
\begin{align*}
    |\Sigma_{\mathcal{A}^{(s)}}(k,r)-\Sigma_\mathcal{A}(k,r)| &\leq 
    \int_{\mathcal C_0} \left|
    \frac{\mathcal{A}_k^{(s)}(z)^2}{\sum_{k=1}^K \mathcal{A}_k^{(s)}(z)} -
    \frac{\mathcal{A}_k(z)^2}{\sum_{k=1}^K \mathcal{A}_k(z)}
    \right| dz 
\end{align*}
By the construction of $\mathcal{A}_k^{(s)}$ and $h_m=O((nN)^{-\frac{1}{2\beta+d}})$, we have 
\begin{equation*}
    \max_z |\mathcal{A}_k^{(s)}(z)-\mathcal{A}_k(z)|\leq O\left(\left(\frac{1}{nN}\right)^{\frac{\beta}{2\beta+d}}\right) 
\end{equation*}
Since $\mathcal{A}_k(z)\geq C^{***}_{\beta,d}$ on $\mathcal C_0$ for some constant $C^{***}_{\beta,d}>0$, we have $\mathcal{A}_k(z)\geq C^{***}_{\beta,d}/2$ on that interval for sufficiently large $nN$. Furthermore, 
\begin{align*}
    &\left|
    \frac{\mathcal{A}_k^{(s)}(z)}{\sum_{k=1}^K \mathcal{A}_k^{(s)}(z)} -
    \frac{\mathcal{A}_k(z)}{\sum_{k=1}^K \mathcal{A}_k(z)}
    \right| = \left|
    \frac{\sum_{r\neq k}\mathcal{A}_r(z)\mathcal{A}^{(s)}_k(z)- \mathcal{A}_k(z)\mathcal{A}^{(s)}_r(z)}{\left(\sum_{k=1}^K \mathcal{A}_k(z)\right)\left(\sum_{k=1}^K \mathcal{A}_k^{(s)}(z)\right)}
    \right| \\
    &\leq 
    \frac{\left|\sum_{r\neq k}\mathcal{A}_r(z)\mathcal{A}^{(s)}_k(z)-\mathcal{A}_r(z)\mathcal{A}_k(z)\right|+\left| \mathcal{A}_k(z)\mathcal{A}_r(z)-\mathcal{A}_k(z)\mathcal{A}^{(s)}_r(z)\right|}{\left(\sum_{k=1}^K \mathcal{A}_k(z)\right)\left(\sum_{k=1}^K \mathcal{A}_k^{(s)}(z)\right)}\\
    &\leq \frac{\sum_{r\neq k}
    \mathcal{A}_r(z)\left|\mathcal{A}^{(s)}_k(z)-\mathcal{A}_k(z)\right|+ \mathcal{A}_k(z)\left|\mathcal{A}_r(z)-\mathcal{A}^{(s)}_r(z)\right|}{\left(\sum_{k=1}^K \mathcal{A}_k(z)\right)\left(\sum_{k=1}^K \mathcal{A}_k^{(s)}(z)\right)}\\
    &=O\left(\left(\frac{1}{nN}\right)^{\frac{\beta}{2\beta+d}}\right) 
\end{align*}
Therefore, 
\begin{align*}
    |\Sigma_{\mathcal{A}^{(s)}}(k,r)-\Sigma_\mathcal{A}(k,r)| &\leq 
    \int_{\mathcal C_0} \left|
    \frac{\mathcal{A}_k^{(s)}(z)^2}{\sum_{k=1}^K \mathcal{A}_k^{(s)}(z)} -
    \frac{\mathcal{A}_k(z)^2}{\sum_{k=1}^K \mathcal{A}_k(z)}
    \right| dz \\
    &\leq I + II
\end{align*}
where 
\begin{align*}
    I &= \int_{\mathcal C_0} \left|
    \frac{\mathcal{A}_k^{(s)}(z)\mathcal{A}_k(z)}{\sum_{k=1}^K \mathcal{A}_k(z)} -
    \frac{\mathcal{A}_k(z)^2}{\sum_{k=1}^K \mathcal{A}_k(z)}
    \right| dz = O\left(\left(\frac{1}{nN}\right)^{\frac{\beta}{2\beta+d}}\right) 
\end{align*}
and 
\begin{align*}
    II = \int_{\mathcal C_0} \left|
    \frac{\mathcal{A}_k^{(s)}(z)\mathcal{A}_k^{(s)}(z)}{\sum_{k=1}^K \mathcal{A}_k^{(s)}(z)} -
    \frac{\mathcal{A}_k^{(s)}(z)\mathcal{A}_k(z)}{\sum_{k=1}^K \mathcal{A}_k(z)}
    \right| dz = O\left(\left(\frac{1}{nN}\right)^{\frac{\beta}{2\beta+d}}\right)
\end{align*}
Thus we have completed the proof. \qed

\subsection{Proof of Theorem~\ref{thm:estW} and Corollary~\ref{cor:estW}}

In the proof of Lemma~\ref{lem:var-new}, we have verified that the reduced model on hyperword counts satisfy all the regularity conditions in \cite{ke2022using}. 
Therefore, we can apply Theorem~3.5 of \cite{ke2022using} (by letting $\beta_n\asymp 1$ there) to obtain that for any $\delta\in (0,n^{-1})$, with probability $1-n\delta$, conditioning on $N_1, N_2, \ldots, N_n$, 
\[
\|\hat{w}_i - w_i\|_1\leq C\sqrt{\frac{M\log(n)}{\bar{N}n}} + C\sqrt{\frac{-\log(n\delta)}{N_i}}, \qquad \mbox{simultaneously for }1\leq i\leq N_i,
\]
where the probability is $(1-n\delta)$, instead of $(1-\delta)$ as in \cite{ke2022using}, because we have used the probability union bound to guarantee the ``simultaneous'' claim for all $i$. 
In \eqref{lem-var-2}, we have shown that $N/2\leq N_i\leq 3N/2$ for all $1\leq i\leq n$, with probability $1-o(n^{-2})$. Combining this with the above inequality, letting $\delta=n^{-4}$, and using $M\asymp 1/\epsilon^d$, we obtain that with probability $1-o(n^{-2})$,
\[
\max_{1\leq i\leq n}\|\hat{w}_i - w_i\|_1\leq C\sqrt{\frac{\log(n)}{Nn\epsilon^d}} + C\sqrt{\frac{\log(n)}{N}}. 
\]
This proves Theorem~\ref{thm:estW}. For Corollary~\ref{cor:estW}, since $n\epsilon^d\to\infty$ under the additional assumptions, the right hand side becomes $CN^{-1/2}\sqrt{\log(n)}$. \qed


%
%

\section{Supplementary Results for the AP Dataset} \label{supp:AlgAP}

\subsection{Tuning Parameters and Implementation Details}\label{supp:tuningap}




After preprocessing, there are approximately $390,000$ word embeddings in total. In this study, we specify the number of hyperwords as $M=600$ to address the bias-variance trade-off, as analyzed in \cref{sec:Method}. In general, we recommend setting $M$ between $0.05\%$ and $0.2\%$ of the total number of embeddings. Within this range, we observe that the experimental results obtained from TRACE remain stable.

Before the net-rounding step, we first perform dimension reduction on the embeddings. Specifically, we sub-sample \( 20\% \) of the embeddings to fit a UMAP model and apply the learned transformation to all embeddings. The output dimension is set to \( d = 10 \), the number of neighbors to \( 10 \), and the minimum distance to \( 0.1 \) (see \cref{supp:umap} for more details on these UMAP parameters). We recommend setting \( d \) to be roughly comparable to the anticipated number of topics in the corpus, retaining sufficient thematic information while balancing signal strength and computational efficiency.

For the net-rounding step, we use the mini-batch k-means clustering algorithm from the \texttt{scikit}-\texttt{learn} Python library, with a batch size set to \( 10\% \) of the total number of embeddings, to obtain \( M = 600 \) hyperwords. To ensure algorithmic stability, the initialization is set to \texttt{random}, with \(\texttt{n\_init}=100\), meaning the algorithm selects the best solution from \( 100 \) random initializations based on inertia (i.e., within-cluster sum of squares) \citep{celebi2013comparative}. After net-rounding, the Topic-SCORE algorithm is used to extract topic structures from the hyperword counts in each document. For further details on implementing Topic-SCORE, please refer to \cref{supp:topicscore}.

\subsection{Number of Topics}\label{supp:numberoftopics}

\begin{figure}[htb]
  \centering
  \includegraphics[width=.75\textwidth]{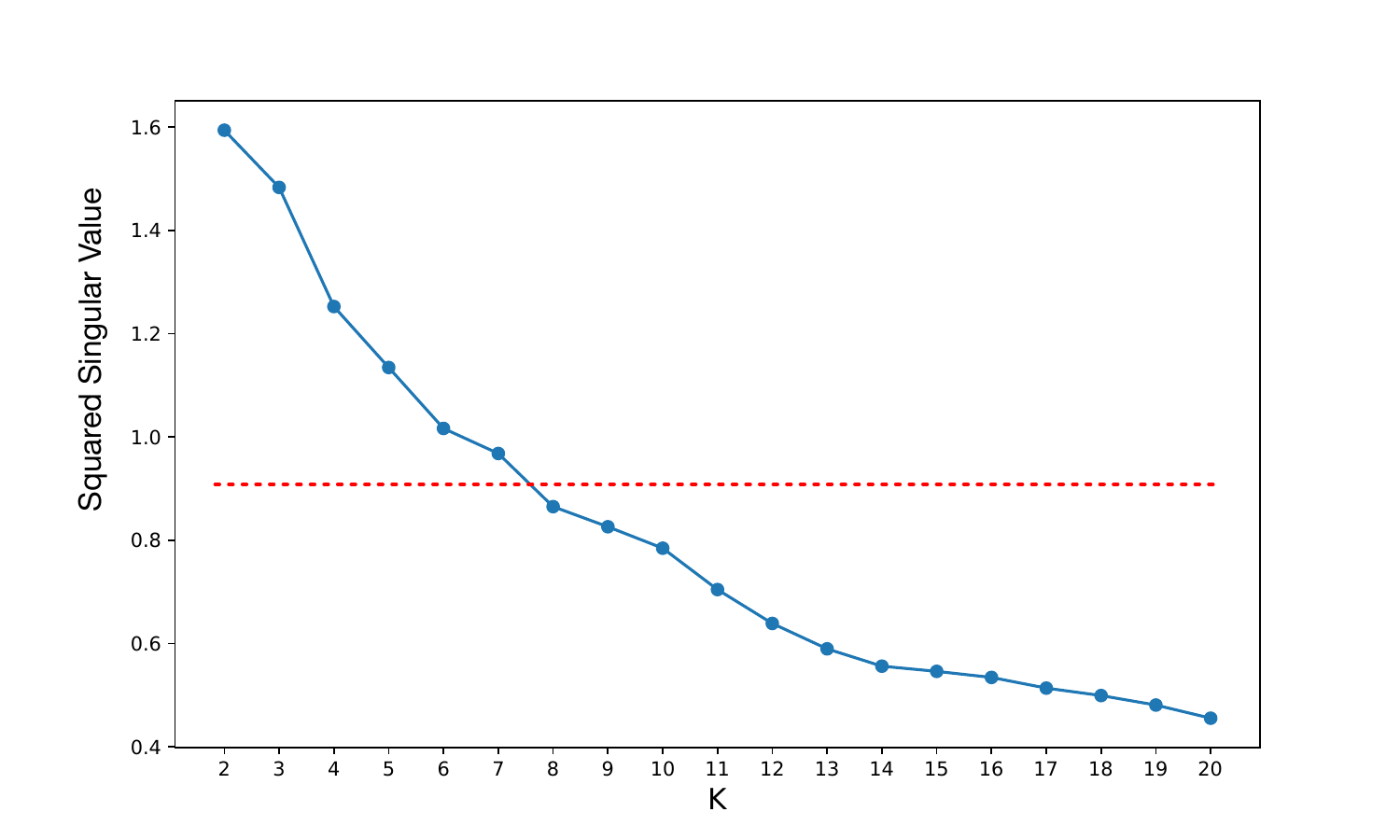} 
  \caption{Scree plot for singular values computed from the AP dataset.}\label{fig:screeplot}
\end{figure}

To determine the number of topics for a text corpus, we recommend examining the scree plot of the singular values from the hyperword frequency matrix. \cref{fig:screeplot} displays the squared singular values, plotted in descending order, for the AP dataset after following the steps described in \cref{supp:tuningap}. Two noticeable gaps in the singular values can be observed: one between 3 and 4, and another between 7 and 8. Notably, selecting \( K = 3 \) topics separates the corpus into distinct categories of \textit{Finance}, \textit{Geopolitics}, and \textit{Entertainment}. 

Interestingly, this result differs slightly from those obtained using Topic-SCORE applied to the word frequency matrix, which identifies \textit{Law \& Crime} instead of \textit{Entertainment}. On one hand, even for \( K = 3 \), the results from TRACE appear more appropriate, as \textit{Law \& Crime} shares some overlap with \textit{Geopolitics}, whereas \textit{Entertainment} is more distinct from the other two topics. Additionally, the corpus contains a substantial number of documents in the \textit{Entertainment} category, further supporting TRACE’s separation of this topic.

This difference may stem from the nature of the word distributions within these topics. The word distribution for \textit{Entertainment} is more diffuse, making its structure harder to discern without external embedding information. In contrast, \textit{Law \& Crime} exhibits a more concentrated word distribution, dominated by terms like \textit{conviction}, \textit{guilty}, \textit{charge}, or \textit{gun}, \textit{shot}, \textit{police}. This concentrated distribution makes it easier for Topic-SCORE to identify its topic structure without relying on external embeddings.

On the other hand, selecting \( K = 7 \) instead of \( K = 3 \) reveals \textit{Law \& Crime} among the four additional topics, as shown in \cref{fig:AP-anchor} in the main paper. This observation suggests that \( K = 7 \) may be a more suitable choice for the number of topics.

{
To explore the relationship between topics at different levels of granularity, we present a hierarchical organization of the topics discovered by TRACE (with UMAP dimension $d = 10$) across varying values of $K$, as shown in Table~\ref{tab:varying-K}. Since TRACE involves some randomness, the table reflects the result from a single run. While independent runs exhibit some variability, recurring patterns are observed. For example, a general topic such as \textit{Geopolitics} may manifest as more focused topics like \textit{Political Leadership} or \textit{Regional Conflicts}, and \textit{Entertainment} may split into subtopics such as \textit{Movies}, \textit{Sports}, and \textit{Music \& Arts}. These observations illustrate the model’s ability to uncover nuanced and interpretable topic hierarchies.

\spacingset{.8}
\begin{table}[tb]
\centering
\caption{\small Hierarchical structure of topics from $K=7$ to $K=20$ in AP dataset. ``X'' indicates that the topic does not appear in certain $K$. ``*'' indicates that this topic may change from time to time in different runs.}
\label{tab:varying-K}
\scalebox{0.9}{
\begin{tabular}{@{} l l l @{}}
\toprule
\textbf{$K=7$}           & \textbf{$K=10$}               & \textbf{$K=20$}             \\
\midrule
\multirow{5}{*}{Finance}
  & \multirow{2}{*}{Financial Market}   & Stock Market           \\
  &                                    & Currency                   \\
\cmidrule(l{2pt}r{2pt}){2-3}
  & Corporation                        & Corporation                 \\
\cmidrule(l{2pt}r{2pt}){2-3}
  & \multirow{2}{*}{Agriculture}     & Agriculture                \\
  &                                    & Environment     \\
\midrule
Domestic Politics          & Election                          & Election                   \\
\midrule
Geopolitics                & Geopolitics                       & Political Leadership*       \\
\midrule
\multirow{2}{*}{Law \& Crime}
  & \multirow{2}{*}{Law \& Crime}     & Judicial Process           \\
  &                                    & Crime                      \\
\midrule
Natural Events             & Natural Events                    & Meteorology*                \\
\midrule
\multirow{3}{*}{Transportation Safety}
  & \multirow{3}{*}{Transportation Safety} & Aviation Safety               \\
  &                                    & Maritime Safety                   \\
  &                                    & Space Exploration          \\
\midrule
\multirow{3}{*}{Entertainment}
  & \multirow{3}{*}{Entertainment}    & Movies                       \\
  &                                    & Sports                     \\
  &                                    & Music \& Arts                       \\
\midrule
\multirow{2}{*}{X}
  & \multirow{2}{*}{Health Care}      & Health Care                \\
  &                                    & Diseases                    \\
\midrule
X                  & X                         & Education                  \\
\midrule
X                   & X                          & Religion*                   \\
\bottomrule
\end{tabular}
}
\end{table}
\resetspacing
}

{
\subsection{Number of UMAP dimension $d$}
\label{supp:umap-d}

The UMAP reduced dimension $d$ serves as a hyperparameter that influences the quality of input embeddings to TRACE and may consequently affect the performance of our method.
However, it is not merely a hyperparameter that may influence TRACE’s ability to discover topic structures and estimate topic weights—it also defines the latent space in which we perform density estimation. Throughout the paper, we fix $d = 10$ while choosing $K = 7$ for the Associated Press dataset and $K = 13$ for the MADStat dataset.

We also examine the robustness of the method when $d$ is modestly larger or smaller than $K$. To this end, we conducted a new experiment on the AP dataset using TRACE with $K = 7$ topics and three different embedding dimensions: $d = 3$, $10$, and $20$. For each configuration, we repeated the experiment five times, extracted the top ten representative words for each topic, and then assigned topic labels accordingly.

The results appear in Tables~\ref{tab:top10-d=3}, \ref{tab:top10-d=10}, and \ref{tab:top10-d=20}. Four topics, \emph{Finance}, \emph{Domestic Politics}, \emph{Geopolitics}, and \emph{Natural Events}, emerge consistently across all settings, indicating robustness over the choice of $d$. However, different $d$ produces slightly varying topic summaries.
At low dimension ($d=3$), the UMAP embedding is too coarse to capture the corpus’s nuance, yielding fragmented or ambiguous topics; indeed, some runs produce uninterpretable clusters (Table~\ref{tab:top10-d=3}). As $d$ grows, we observe that the representative words for each topic become more concentrated and coherent. When $d=10$, Table \ref{tab:top10-d=10} yields a narrower scope of representative words but there is still a substantial overlap among multiple runs. For example, Topic~6 spans both aerospace terms (\emph{shuttle}, \emph{rocket}) and aviation terms (\emph{airline}, \emph{pilot}), even though it clearly pertains to transportation. At $d=20$, however, topics become sharply focused and exhibit minimal overlap across runs (Table~\ref{tab:top10-d=20}).
Overall, embeddings of moderate dimensionality, on the order of $K$, strike the best balance between topic separation and interpretability.

\spacingset{.8}
\begin{table}[tbp]
\centering
\caption{\small Top 50 anchor words for each topic of AP from TRACE.}
\scalebox{.95}{
\begin{tabular}{cp{2.5cm}|p{14cm}}
\toprule
& {\small Topics} & {\small Runs (10 representative top words)} \\
\midrule
{\small 1} & {\small Finance} &
{\small Run 1: index, exchange, trade, stock, market, composite, expire, issue, yen, percent; \newline
Run 2: yen, index, exchange, trade, stock, composite, expire, dollar, pound, mark; \newline
Run 3: yen, ounce, index, composite, exchange, bullion, lire, issue, trader, share; \newline
Run 4: index, exchange, composite, expire, unchanged, trading, stock, quote, issue, percent; \newline
Run 5: index, exchange, composite, unchanged, stock, trade, market, issue, rate, equity}\\
\midrule
{\small 2} & {\small \parbox[t]{3cm}{Domestic\\Politics}} &
{\small Run 1: communist, socialist, democrat, party, gop, marxist, leftist, opposition, democracy, republic; \newline
Run 2: campaign, election, candidate, primary, nomination, runoff, caucus, party, elect, rival; \newline
Run 3: presidential, primary, election, voter, turnout, campaign, candidate, democratic, republican, ballot; \newline
Run 4: gop, election, campaign, democrat, primary, republican, party, candidate, race, opponent; \newline
Run 5: democratic, republican, campaign, party, election, candidate, runoff, incumbent, district, nomination} \\
\midrule
{\small 3} & {\small Geopolitics} &
{\small Run 1: terrorist, rebel, militant, guerrilla, militia, insurgency, radical, attack, resistance, front; \newline
Run 2: israel, palestinian, gaza, jordan, armenia, iraqi, iraq, kuwait, saudi, syria; \newline
Run 3: lebanon, soldier, veteran, marine, troop, force, armed, iraq, combat, patrol; \newline
Run 4: damage, injury, shooting, assault, murder, fatal, bullet, casualty, wound, kill; \newline
Run 5: opposition, communist, marxist, socialist, capital, truce, peace, unity, liberation, movement} \\
\midrule
{\small 4} & {\small \parbox[t]{3cm}{Law \&\\Crime}} &
{\small Run 1: court, judge, jury, appeal, trial, supreme, district, prosecutor, defense, plea; \newline
Run 2: charge, plead, convict, guilty, count, indictment, suspect, sentence, bail, prosecution; \newline
Run 3: jury, appeal, court, judge, trial, magistrate, counsel, justice, verdict, testimony; \newline
Run 4: supreme, district, circuit, appeal, judge, court, trial, legal, judicial, justice; \newline
Run 5: convict, charge, plea, sentence, bail, defendant, punishment, fine, probation, court} \\
\midrule
{\small 5} & {\small \parbox[t]{3cm}{Natural\\Events}} &
{\small Run 1: quake, earthquake, avalanche, fault, scale, storm, flood, rain, wind, thunderstorm; \newline
Run 2: mexico, downtown, mexican, venezuela, central, latin, western, northern, drought, heat; \newline
Run 3: winter, cold, snow, heat, thunderstorm, cloud, hail, rainfall, storm, tropical; \newline
Run 4: winter, snow, storm, rain, temperature, cold, hot, degree, shower, front; \newline
Run 5: utah, alaska, yellowstone, desert, snow, rain, heat, temperature, winter, hail} \\
\midrule
{\small 6} & {\small \parbox[t]{3cm}{Transportation\\Safety}} &
{\small Run 1: shuttle, mission, rocket, satellite, nasa, flight, crew, launch, spacecraft, booster; \newline
Run 4: boeing, airbus, jet, flight, carrier, airline, airport, runway, faa, pilot; \newline
Run 5: navy, pilot, flight, marine, crew, attendant, charter, takeoff, wing, transport} \\
\midrule
{\small 7} & {\small Entertainment} &
{\small Run 2: symphony, orchestra, musician, record, band, concert, opera, composer, performance, score; \newline
Run 3: dance, ballet, music, concert, orchestra, theater, play, actor, singer, performance; \newline
Run 4: tape, film, movie, video, recorder, director, comedy, act, singer, performance; \newline
Run 5: award, medal, prize, ceremony, festival, theater, show, music, performance, entertainer} \\
\midrule
{\small 8} & {\small Unknown} &
{\small Run 1: love, humor, flow, symbol, balance, test, competition, absence, shape, voice; \newline
Run 2: nancy, bush, love, test, humor, element, panic, phase, disaster, challenge; \newline
Run 3: harvard, yale, cambridge, bishop, king, pope, pontiff, thomas, ray, owen} \\
\bottomrule
\end{tabular}
}
\label{tab:top10-d=3}
\end{table}
\resetspacing

\spacingset{.8}
\begin{table}[tbp]
\centering
\caption{\small Representative words for each theme at $d=10$, grouped by topic across five runs.}
\label{tab:top10-d=10}
\scalebox{0.95}{
\setlength\tabcolsep{4pt}
\begin{tabular}{cp{2.5cm}|p{14cm}}
\toprule
& {\small Topics} & {\small Runs (10 representative top words)} \\
\midrule
{\small 1} & {\small Finance} &
{\small Run 1: volume, jones, index, exchange, market, trade, stock, wall, dollar, rise; \newline
Run 2: unchanged, jones, trade, list, insider, issue, drop, climb, slip, surge; \newline
Run 3: volume, unchanged, jones, dow, market, exchange, index, trade, stock, currency; \newline
Run 4: index, composite, dealer, trader, brokerage, market, exchange, volume, trade, stock; \newline
Run 5: index, exchange, composite, unchanged, stock, trade, market, issue, rate, equity} \\
\midrule
{\small 2} & {\small \parbox[t]{3cm}{Domestic\\Politics}} &
{\small Run 1: gop, democratic, republican, democrats, democrat, republicans, campaign, election, candidate; \newline
Run 2: elect, run, nomination, campaign, candidate, primary, race, opponent, party, ticket; \newline
Run 3: primary, campaign, election, voter, ballot, candidate, democrat, gop, runoff, caucus; \newline
Run 4: gop, republican, democratic, campaign, election, candidate, party, primary, district, vice; \newline
Run 5: democratic, republican, campaign, party, election, candidate, runoff, incumbent, district, nomination} \\
\midrule
{\small 3} & {\small Geopolitics} &
{\small Run 1: terrorist, rebel, militant, guerrilla, militia, insurgency, radical, attack, resistance, front; \newline
Run 2: occupy, rebel, truce, peace, plo, swapo, unita, resistance, uprising, liberation; \newline
Run 3: africa, south, african, angola, zimbabwe, anc, plo, liberation, front, unity; \newline
Run 4: south, africa, angola, plo, unita, inkatha, zimbabwe, natal, resistance, front; \newline
Run 5: guerrilla, rebel, movement, liberation, path, insurgency, radical, attack, front, unity} \\
\midrule
{\small 4} & {\small \parbox[t]{3cm}{Law \&\\Crime}} &
{\small Run 1: circuit, supreme, appeal, court, judge, jury, trial, district, justice, prosecutor; \newline
Run 2: charge, plead, guilty, convict, accuse, indictment, trial, count, sentence, bail; \newline
Run 3: trial, indictment, defense, justice, plea, court, judge, prosecutor, sentencing, verdict; \newline
Run 4: supreme, district, circuit, appeal, judge, trial, court, law, justice, hearing; \newline
Run 5: plead, guilty, conviction, charge, acquit, sentence, bail, trial, court, judge} \\
\midrule
{\small 5} & {\small \parbox[t]{3cm}{Natural\\Events}} &
{\small Run 1: creek, river, mountain, lake, beach, valley, peak, basin, hill, plain; \newline
Run 2: manhattan, brooklyn, coast, valley, park, canyon, plain, hill, river, city; \newline
Run 3: creek, city, coast, river, mountain, lake, beach, valley, peak, manhattan; \newline
Run 4: springs, coast, coastal, beach, hills, florida, valley, canyon, ridge, park; \newline
Run 5: creek, city, coast, river, mountain, lake, beach, valley, peak, basin} \\
\midrule
{\small 6} & {\small \parbox[t]{3cm}{Transportation\\Safety}} &
{\small Run 1: force, guard, fly, charter, landing, flight, route, passenger, plane, transport; \newline
Run 2: ground, air, douglas, aerospace, launch, shuttle, mission, satellite, booster, flight; \newline
Run 4: airline, airport, carrier, runway, faa, pilot, flight, charter, transport, engine; \newline
Run 5: navy, pilot, flight, marine, crew, attendant, aircraft, passenger, transport, charter} \\
\midrule
{\small 7} & {\small Entertainment} &
{\small Run 1: wife, show, dance, group, band, rock, music, concert, singer, song; \newline
Run 2: art, artist, painting, painter, ballet, music, concert, theater, performance, exhibition; \newline
Run 3: film, movie, director, hollywood, screen, entertainment, picture, blockbuster, production, videotape; \newline
Run 4: wife, music, song, play, band, art, concert, dance, painter, exhibition; \newline
Run 5: film, movie, hollywood, picture, blockbuster, entertainment, director, production, shoot, screen} \\
\midrule
{\small 8} & {\small Education} &
{\small Run 3: club, endowment, foundation, harvard, society, cambridge, yale, scholarship, college, campus} \\
\bottomrule
\end{tabular}
}
\end{table}
\resetspacing

\spacingset{.8}
\begin{table}[tbp]
\centering
\caption{\small Representative words for each theme at $d=20$, grouped by topic across five runs.}
\label{tab:top10-d=20}
\scalebox{0.95}{
\setlength\tabcolsep{4pt}
\begin{tabular}{cp{2.5cm}|p{14cm}}
\toprule
& {\small Topics} & {\small Runs (10 representative top words)} \\
\midrule
{\small 1} & {\small Finance} &
{\small Run 1: issue, index, exchange, composite, volume, stock, trade, market, dollar, rise; \newline
Run 2: point, percent, cent, index, exchange, price, gain, drop, climb, slump; \newline
Run 3: yen, dollar, mark, silver, pound, bullion, ounce, franc, unrest, rate; \newline
Run 4: lire, troy, bullion, ounce, yen, settle, finish, broker, decline, surge; \newline
Run 5: issue, index, exchange, composite, dow, street, stock, rate, equity, bond} \\
\midrule
{\small 2} & {\small \parbox[t]{3cm}{Domestic\\Politics}} &
{\small Run 1: term, gop, democratic, republican, election, candidate, campaign, party, primary, caucus; \newline
Run 2: campaign, election, candidate, primary, nomination, run, ticket, elect, opponent, rival; \newline
Run 3: gop, majority, minority, primary, caucus, senator, representative, vote, ballot, turnout; \newline
Run 4: gop, dole, havel, gov, rep, senator, congress, chamber, policy, state; \newline
Run 5: republican, democratic, campaign, party, election, candidate, district, run, bid, convention} \\
\midrule
{\small 3} & {\small Geopolitics} &
{\small Run 1: arafat, avril, yitzhak, aoun, aristide, plo, liberation, truce, unity, movement; \newline
Run 2: africa, south, african, angola, mandela, zimbabwe, plo, unita, liberation, swapo; \newline
Run 3: south, mandela, africa, african, angola, zimbabwe, liberation, unity, front, revolt; \newline
Run 4: occupy, south, plo, african, angola, rebel, truce, peace, confrontation, unrest; \newline
Run 5: occupy, aristide, west, likud, yasser, conflict, peace, truce, rally, movement} \\
\midrule
{\small 4} & {\small \parbox[t]{3cm}{Law \&\\Crime}} &
{\small Run 1: defense, trial, indictment, courtroom, sentencing, justice, judge, plea, prosecutor, tribunal; \newline
Run 2: jury, juror, attorney, counsel, represent, appeal, law, judge, court, verdict; \newline
Run 3: jury, juror, appeal, circuit, supreme, judge, justice, trial, counsel, court; \newline
Run 5: plead, guilty, try, innocent, conviction, acquit, sentence, penalty, probation, bail} \\
\midrule
{\small 5} & {\small \parbox[t]{3cm}{Natural\\Events}} &
{\small Run 1: california, angeles, yosemite, island, san, francisco, santa, clara, oakland, riverside; \newline
Run 2: mile, mph, miles, foot, inch, yard, scale, survey, depth, elevation; \newline
Run 3: mph, park, manhattan, brooklyn, river, canyon, national, valley, plain, lake; \newline
Run 4: coast, coastal, beach, hills, fall, ridge, park, village, canyon, lake; \newline
Run 5: quake, earthquake, fault, storm, weather, rain, wind, flood, magnitude, drought} \\
\midrule
{\small 6} & {\small \parbox[t]{3cm}{Transportation\\Safety}} &
{\small Run 1: missile, rocket, missiles, crew, crewman, force, guard, submarine, launch, mission; \newline
Run 2: navy, marine, naval, petty, board, forces, unit, guard, veteran, service; \newline
Run 3: official, office, eastern, medellin, orion, airline, crash, accident, disaster, collision; \newline
Run 4: station, engine, road, train, rail, highway, bridge, transit, vehicle, track; \newline
Run 5: chrysler, general, ford, hyundai, boeing, lockheed, mcdonnell, aircraft, airline, airport} \\
\midrule
{\small 7} & {\small Entertainment} &
{\small Run 1: publication, newspaper, journal, magazine, editorial, reporter, column, press, newspaper, media; \newline
Run 2: art, musician, piano, music, artist, theater, orchestra, symphony, performance, exhibition; \newline
Run 3: kennedy, disney, mickey, presley, henson, actor, star, entertainer, cinema, hollywood; \newline
Run 4: disney, mickey, nixon, presley, henson, painting, artist, gallery, exhibit, performance; \newline
Run 5: luther, king, thomas, martin, ryan, band, concert, music, festival, award} \\
\midrule
{\small 8} & {\small Education} &
{\small Run 4: student, university, campus, professor, course, degree, college, research, faculty, academic} \\
\bottomrule
\end{tabular}
}
\end{table}
\resetspacing
}

\subsection{Bandwidth Selection via Maximum Entropy Criterion}\label{supp:entropy}

Differential entropy quantifies the extent to which the probability density of a continuous random variable is spread out. Distributions with high differential entropy, such as the uniform distribution, allocate their probability density over a wide range of values, reflecting greater uncertainty and requiring more information to describe a realization of the variable. In contrast, distributions with low differential entropy concentrate their probability density within a narrower range, indicating less uncertainty \citep{jaynes1957information,cover2012elements}.

In this study, we interpret the estimated mixed-membership vectors of word embeddings as points within a simplex, with the vertices representing different topics. These points are treated as samples drawn from a continuous distribution (e.g., a Dirichlet distribution) over the simplex. A more dispersed point cloud indicates higher uncertainty in the underlying distribution, leading to better topic separation, more equitable topic proportions, and greater diversity in topic representations. Specifically, we aim to avoid scenarios where documents cluster excessively around a single topic, an equal mixture of all topics, or any fixed-weight combination of topics. Additionally, we seek to prevent cases where a subset of topics disproportionately dominates the document set. As a result, a higher entropy estimate is desirable to achieve these objectives.

\begin{figure}[htb]
  \centering
  \includegraphics[width=\textwidth]{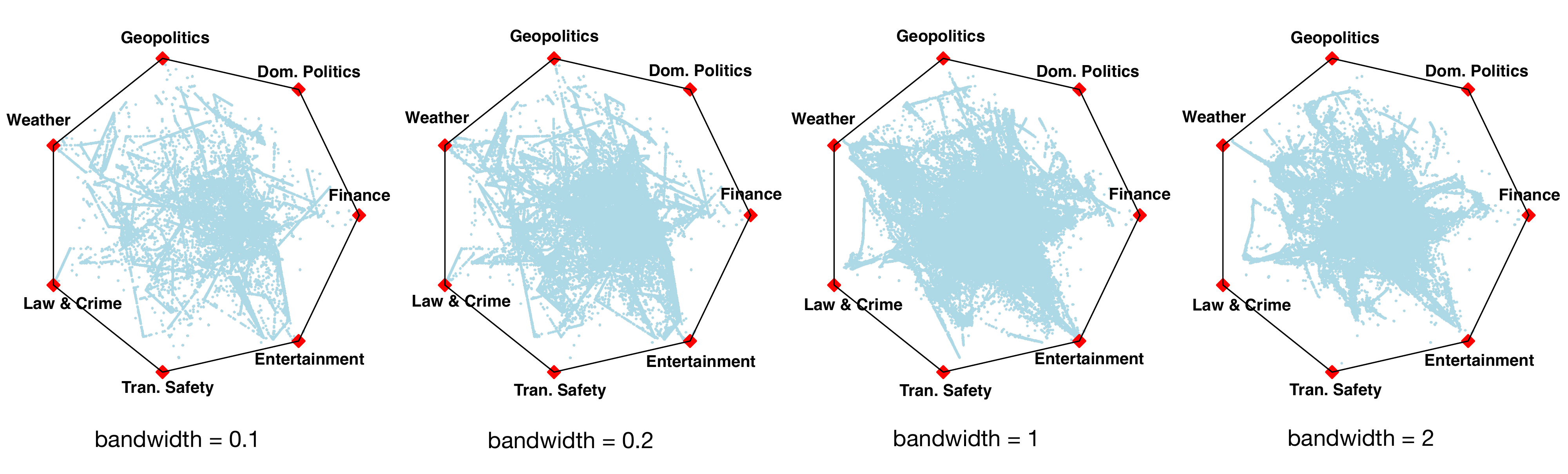} 
  \caption{Comparison of heptagon plots with different bandwidths.}\label{fig:bwcompare}
\end{figure}

\begin{figure}[htb]
  \centering
  \includegraphics[width=.75\textwidth]{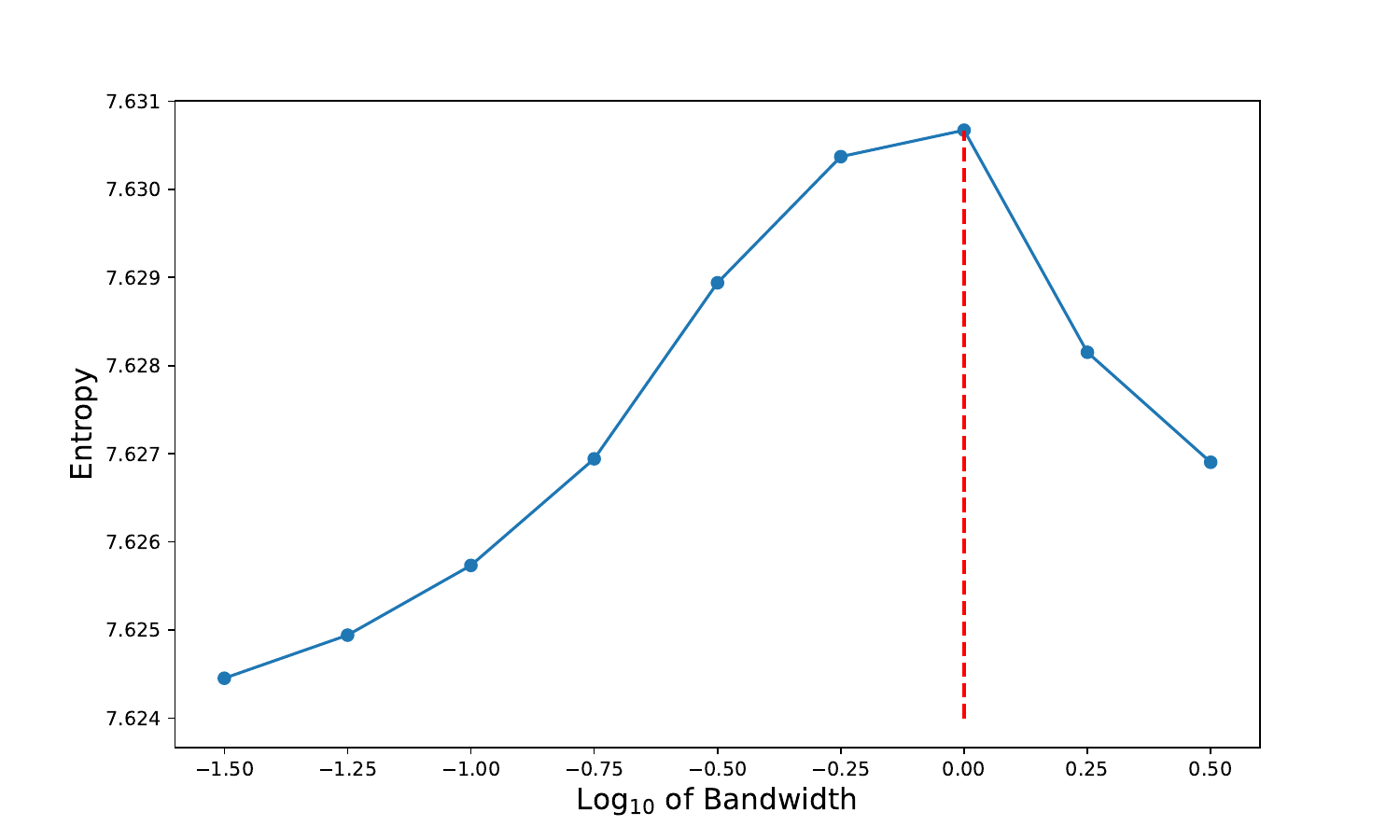} 
  \caption{Entropy estimates of point clouds corresponding to different bandwidths.}\label{fig:entropyplot}
\end{figure}

To illustrate, we also visualize the mixed-membership vectors of word embeddings from the AP dataset as points within a heptagon. \cref{fig:bwcompare} shows these point clouds under varying bandwidth settings. With a large bandwidth, the point cloud contracts toward the center of the heptagon, indicating reduced diversity of topic mixtures. Conversely, with a small bandwidth, the points cluster in fewer locations but remain distributed across the heptagon, reflecting a bumpier underlying distribution. Both scenarios result in lower entropy estimates for the point cloud, as depicted in \cref{fig:entropyplot}, suggesting reduced uncertainty. Based on these observations, we select an optimal bandwidth of \( 1 \).

Specifically, \cref{fig:entropyplot} is generated using the well-established \( k \)-nearest neighbor (kNN) entropy estimator \citep{berrett2019efficient} with parameter \( k=25 \). The kNN entropy estimator is a nonparametric method for estimating the entropy of a distribution, particularly advantageous for high-dimensional data where traditional density estimation methods face challenges. It leverages the distances to the \( k \)-nearest neighbors of each data point to assess the spread or randomness of the distribution.

Given a set of \( n \) points \( X = \{x_1, x_2, \ldots, x_n\} \) in \( d \)-dimensional space, the kNN entropy estimator for the empirical distribution, \( H_{\text{empirical}} \), is calculated as:
\[
H_{\text{empirical}} \approx -\psi(k) + \psi(n) + \frac{d}{n} \sum_{i=1}^n \log\left(\epsilon_i\right),\quad \text{where}
\]
\begin{itemize}
\setlength{\itemsep}{-5pt}
\item $k$ is the number of nearest neighbors (the choice depends on $n$ and $d$);
\item $\epsilon_{\mathrm{i}}$ is the distance from point $x_i$ to its $k$-th nearest neighbor in the dataset;
\item $d$ is the dimensionality of the space; and
\item $\psi$ is the digamma function $\psi(x)=\frac{d}{d z} \log (\Gamma(x))$, where $\Gamma(x)$ is the gamma function. 
\end{itemize}
For our goal of selecting the bandwidth that maximizes entropy, it is sufficient to focus on the final term, 
\(
\frac{d}{n} \sum_{i=1}^n \log \left(\epsilon_i\right)
\),
where \(d\) represents \(K-1\), the number of topics minus one, and \(n\) denotes the number of distinct word embeddings. For consistency, we always downsample \(n\) to \(50,000\) in our estimation.

\subsection{Additional Results}\label{supp:apadditional}



\cref{tb:AP-trace-more} summarizes the top 50 words ranked by anchorness across the seven topics of the AP dataset. While most of these words are unique to specific topics, a few exceptions stand out due to their relevance to multiple contexts. For example, the word \textit{coast} appears in both \textit{Natural Events} and \textit{Transportation Safety}, as it is pertinent to both weather conditions and maritime incidents occurring near coastal areas. Similarly, the word \textit{crash} is commonly used in distinct contexts, such as a \textit{market crash} in \textit{Finance} or an \textit{airplane crash} in \textit{Transportation Safety}.



\cref{tb:AP-trace-region-more} lists 20 representative words from each anchor region labeled in \cref{fig:AP-topics} in the main paper. Interestingly, multiple anchor regions can exist within the same topic, reflecting differences in subtopic emphasis or the syntactic roles of words in those regions. For instance, in the topic \textit{Finance}, Region 1 predominantly contains nouns associated with financial concepts and assets, while Region 2 is characterized by verbs describing dynamic market changes. Similarly, within the topic \textit{Natural Events}, Region 5 primarily includes words related to directions or locations, whereas Region 6 focuses on terms describing natural phenomena or weather conditions.



\begin{figure}[tbp]
  \centering
  \includegraphics[width=.6\textwidth]{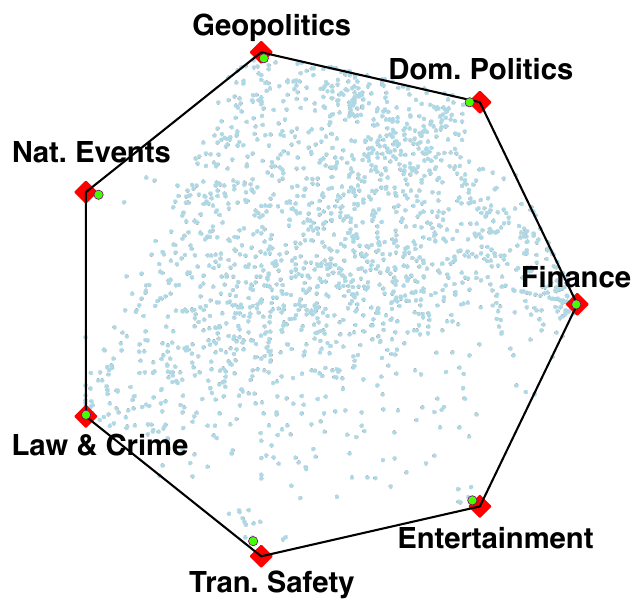} 
  \caption{Heptagon plot with all documents of AP obtained from TRACE.}\label{fig:document-heptagon}
\end{figure}

Following the approach described in \cref{subsec:AP}, we visualize the positions of all documents within the topic heptagon, as shown in \cref{fig:document-heptagon}. Each vertex of the heptagon corresponds to a topic, and each dot represents a document, with its position determined by the document’s mixed-membership topic weight vector. The distribution of documents across the heptagon indicates that the topics identified by TRACE are both balanced and informative, capturing the complexity of documents that exhibit mixtures of topics.

For each topic, a representative pure document is highlighted in green in \cref{fig:document-heptagon}, with its specific content detailed in \cref{tb:AP-doc}. These documents clearly represent the seven topics without ambiguity, illustrating the interpretability and coherence of the discovered topics.

\spacingset{.8}
\begin{table}[tbp]
\centering
\caption{\small Top 50 anchor words for each topic of AP from TRACE.} \label{tb:AP-trace-more}
\scalebox{.92}{
\begin{tabular}{cl|p{14cm}}
\toprule
& {\small Topics} & {\small Top 50 anchor words for each topic of AP from TRACE} \\
\midrule
{\small 1}& {\small Finance}& {\footnotesize index, exchange, composite, issue, unchanged, volume, street, wall, market, dollar, marketplace, economy, broker, trader, dealer, brokerage, dow, yield, rate, inflation, rise, fall, drop, come, climb, lose, decline, increase, jump, gain, advance, grow, slow, tumble, soar, edge, double, pick, slip, rose, plunge, surge, move, retreat, add, shoot, expand, go, recover, risen}\\ 
\midrule
{\small 2}& {\small Domestic Politics}& {\footnotesize gop, dole, havel, sen, gov, rep, secretary, maj, term, democratic, republican, democrat, senator, democrats, republicans, vice, house, party, conservative, bush, senate, congressman, subcommittee, committee, chamber, congressional, representative, capitol, dukakis, dukaki, duracell, durenberger, darman, duarte, campaign, favorite, rival, run, seek, mate, opponent, incumbent, running, predecessor, campaigning, candidate, successor, challenger, bid, nominee}\\ 
\midrule
{\small 3}& {\small Geopolitics}& {\footnotesize west, likud, yitzhak, peres, shimon, yasser, tel, arafat, pere, perestroika, aristide, ershad, aoun, avril, middle, israel, palestinian, israeli, bank, gaza, palestinians, jerusalem, east, israelis, palestine, sharon, aviv, iran, plo, liberation, nations, interior, revolution, coup, rule, state, dictatorship, dictator, government, authoritarian, sandinista, duvali, central, regime, communist, iraq, iraqi, morocco, iranian, lebanese}\\ 
\midrule
{\small 4}& {\small Natural Events}& {\footnotesize beach, coast, coastal, spring, springs, florida, hills, cape, lake, ridge, creek, fall, faithful, hill, monte, valley, manhattan, park, brooklyn, point, yellow, great, river, national, central, west, plain, height, canyon, grand, peak, side, grande, mount, old, fork, temperature, winter, high, degree, cold, january, heat, cloudy, snow, hot, sun, cool, shower, warm}\\ 
\midrule
{\small 5}& {\small Law \& Crime}& {\footnotesize conviction, try, plead, guilty, convict, acquit, innocent, prosecute, find, clear, indict, charge, prosecutor, indictment, trial, prosecution, defense, sentencing, plea, courthouse, justice, sentence, courtroom, verdict, federal, magistrate, tribunal, testimony, fine, punish, warrant, penalty, punishment, term, execution, maximum, minimum, execute, time, reward, probation, bail, accuse, suspicion, count, connection, offense, suspect, allege, repute}\\ 
\midrule
{\small 6}& {\small Transportation Safety}& {\footnotesize coast, accident, disaster, crash, collapse, collision, fatal, tragedy, fatality, navy, naval, marine, petty, ship, sea, port, vessel, boat, maritime, shipping, sailor, canal, harbor, ocean, bay, carrier, fleet, warship, uss, island, waterway, plane, water, ashore, frigate, dock, craft, tanker, midway, cruise, submarine, shore, deck, fisherman, hull, fishing, swimming, coral, shark, force}\\ 
\midrule
{\small 7}& {\small Entertainment}& {\footnotesize award, honor, prize, nobel, medal, winner, peace, honoraria, nominate, good, nominee, academy, awards, history, record, historical, fame, tradition, historian, heritage, culture, legacy, past, cultural, hit, historic, choose, name, dance, museum, show, song, collection, sing, singe, library, archaeologist, singing, picture, museums, ballet, tour, concert, choir, photograph, play, date, perform, vocal, dancing}\\ 
\bottomrule
\end{tabular}}
\end{table}
\resetspacing

\spacingset{.8}
\begin{table}[tbp]
\centering
\caption{\small 20 representative words for each labeled region of AP from TRACE.} \label{tb:AP-trace-region-more}
\scalebox{.92}{
\begin{tabular}{l|p{12cm}}
\toprule
{\small Regions} & {\small 20 representative words for each labeled region of \cref{fig:AP-topics} in the main paper, obtained from TRACE} \\
\midrule
{\small 1 (Finance)} & {\footnotesize level, price, order, production, stock, inflation, cent, issue, start, economy, sale, demand, future, value, export, economic, margin, wage, earning, charge}\\ 
\midrule
{\small 2 (Finance)}& {\footnotesize rise, climb, drift, sink, percent, grow, weaken, edge, retreat, shoot, move, jump, drop, advance, come, push, go, add, slow, slip}\\ 
\midrule
{\small 3 (Domestic Politics)}& {\footnotesize gop, communist, senator, bob, carl, minority, party, chess, gore, bush, democratic, rep, house, democrat, sen, republican, daniel, independent, district, majority}\\ 
\midrule
{\small 4 (Geopolitics)}& {\footnotesize morocco, invade, army, japan, yugoslavia, north, kuwait, germany, invasion, france, terrorism, afghanistan, pentagon, gulf, regime, india, lebanon, poland, arabia, hondura}\\ 
\midrule
{\small 5 (Natural Events)}& {\footnotesize northwestern, east, south, british, wake, southern, north, case, southeast, northeast, midwest, midw, west, eastern, northern, southwestern, carolina, condition, front, florida}\\ 
\midrule
{\small 6 (Natural Events)}& {\footnotesize thunderstorm, system, sustain, earthquake, est, ice, atmosphere, front, warm, white, rainfall, moon, cool, solar, climate, problem, strength, air, frozen, heat}\\ 
\midrule
{\small 7 (Law \& Crime)}& {\footnotesize conviction, fine, innocent, connection, arrest, suspect, offender, find, away, allege, cite, indictment, witness, day, convict, charge, prosecutor, fire, prosecute, contempt}\\ 
\midrule
{\small 8 (Transportation Safety)}& {\footnotesize accident, aircraft, wing, first, nose, pilot, capacity, fly, jump, master, captain, control, operator, logan, plane, attendant, handling, board, cockpit, crew}\\ 
\midrule
{\small 9 (Entertainment)}& {\footnotesize fare, release, tune, piece, art, speaker, conduct, culture, horn, sing, back, version, perform, cry, sing, french, score, style, ballet, piano}\\ 
\bottomrule
\end{tabular}}
\end{table}
\resetspacing

\spacingset{.8}
\begin{table}[tbp]
\centering
\caption{\small Representative document for each topic of AP from TRACE.} \label{tb:AP-doc}
\scalebox{.92}{
\begin{tabular}{cl|p{14cm}}
\toprule
& {\small Topics} & {\small Representative document for each topic of AP from TRACE, as marked in \cref{fig:document-heptagon} (Cut to 100 words)} \\
\midrule
{\small 1}& {\small Finance}& {\footnotesize Share prices ended slightly higher on London's Stock Exchange Friday after a lackluster session. Share prices picked up small gains in early trading as investors took advantage of Thursday's sharp decline to pick up some bargains. But as soon as the buying dried up the market drifted lower, failing to gain support from Wall Street's performance. The Financial Times-Stock Exchange 100-share index was up 4.4 points, or 0.2 percent, at 2,040.6 at the close. The Financial Times 30-share index rose 7.7 points to close at 1,582.6. The Financial Times 500-share index rose 1.15 points to close at 1,085.87. Volume was...}\\ 
\midrule
{\small 2}& {\small \parbox[t]{2.8cm}{ Domestic\\Politics}} & {\footnotesize There will be no organized union boost behind a single candidate in Saturday's Democratic caucuses in Michigan, a state where union members can wield more clout than almost anywhere else. While national labor leaders are assuming Michael Dukakis will be the eventual nominee, they are prevented from endorsing him by what appears to be growing rank-and-file support for Jesse Jackson, who has gotten more union votes than any of the other candidates in primaries so far. Richard Gephardt also has considerable union support. None of the Democratic candidates appears to have won the hearts or votes of a majority of...}\\ 
\midrule
{\small 3}& {\small Geopolitics}& {\footnotesize The leaders of Greek and Turkish Cypriots in Cyprus will meet in New York next month for talks aimed at resolving problems in the divided nation, Secretary-General Javier Perez de Cuellar said today. U.N. officials said Georges Vassiliou, president of the Republic of Cyprus, and Rauf Denktash, president of the Turkish Republic of Northern Cyprus, will visit New York for talks Aug. 24. ``The leaders of the two sides in Cyprus have accepted my proposal for resumption of talks to negotiate a settlement of all aspects of the Cyprus problem,'' Perez de Cuellar said in a brief news conference...}\\ 
\midrule
{\small 4}& {\small Natural Events}& {\footnotesize A band of rain showers extended from Texas to Vermont on Friday, and rain also fell along the northern Pacific Coast. A cold snap in Alaska streched into its 15th day with no relief expected until at least Monday. The overnight low was minus 58 degrees at Fort Yukon. Rain showers and thunderstorms reached from west-central Texas across northern Texas, southeast Oklahoma, northern Arkansas and southwest Missouri. Thunderstorms in Texas brought hail to Waxahachee, Granview, Ferris and Peoria. Rain showers stretched from northeast Louisiana to eastern Mississippi, across northern Alabama, northern Georgia, northwestern South Carolina, Tennessee, Kentucky, West Virginia...}\\ 
\midrule
{\small 5}& {\small Law \& Crime}& {\footnotesize Two mothers tipped police to drugs in their sons' bedrooms, leading to the arrest of one of the teen-agers and confiscation of cocaine from both, police said. The recent arrest of a 14-year-old Landover boy came after his mother told police that she had found the drug in the youth's room, said Carol Landrum, a spokesman for the police department. The son was charged with possession of a controlled dangerous substance with intent to distribute and was being held at Boys Village in Cheltenham, Landrum said. Police said they seized 19 grams of cocaine, which officers said could sell for...}\\ 
\midrule
{\small 6}& {\small \parbox[t]{2.8cm}{ Transportation\\Safety}} & {\footnotesize A TWA jetliner bound for New York returned to Cairo International Airport after taking off Sunday because a telephone caller claimed there was a bomb on board, an airport official said. The official, speaking on condition of anonymity, said the jet returned to the air at 1 p.m. (6 a.m. EDT), 5 hours later, after the plane was evacuated and its passengers and luggage searched. No bomb was found. He said someone called minutes after the plane took off for Paris en route to New York. The anonymous caller said there was a bomb on the plane and hung up...}\\  
\midrule
{\small 7}& {\small Entertainment}& {\footnotesize ``Deep'' (RCA-BMG)---Peter Murphy: The ``deep'' of former Bauhaus singer Peter Murphy's fourth solo project could refer to his unfathomable lyrics, but Murphy has said the title reflects both how he feels about his work deeply and his perception of the music, which he calls ``deep rock.'' It sounds more like doomsday disco. Murphy's old outfit, a post-punk British band with artistic pretensions, was among the first to mix infectious dance grooves with brooding, introspective lyrics. The style lives on in this recording and in records by The Cure, The The and others favored by masses of black-garbed club hoppers...}\\ 
\bottomrule
\end{tabular}}
\end{table}
\resetspacing




Lastly, we demonstrate the advantage of incorporating contextualized word embeddings into statistical topic models by comparing TRACE with Topic-SCORE, applied directly to the word frequency matrix of the AP dataset. When \( K=7 \), TRACE outperforms Topic-SCORE in terms of topic separation and coherence. As shown in \cref{tb:AP-topic-score}, three well-defined topics obtained from TRACE—\textit{Natural Events}, \textit{Transportation Safety}, and \textit{Domestic Politics}—are not clearly distinguished by Topic-SCORE without the use of contextualized word embeddings. Moreover, Topic-SCORE splits the \textit{Geopolitics} topic into two subtopics that share related traits, while also identifying two additional ``ambiguous'' topics that are challenging to summarize. For instance, Topic 6 includes a mix of anchor words related to economics (e.g., \textit{salary}, \textit{mortgage}), healthcare (e.g., \textit{medicare}, \textit{arthritis}), and survey analysis (e.g., \textit{survey}, \textit{poll}, \textit{respondent}), reflecting a lower level of topic coherence.

In addition, we demonstrate that TRACE preserves semantic knowledge from pre-trained language models in its mixed-membership topic weight vectors for each word. To investigate this, we compare the vectors generated by TRACE and Topic-SCORE for pairs of synonyms. For this experiment, we set \( K=3 \), as the interpretability of Topic-SCORE’s results diminishes at \( K=7 \), making it difficult to establish clear correspondence between the topics identified by the two methods. In Topic-SCORE, each word is assigned a unique position in the topic triangle, whereas TRACE generates multiple embeddings for each word, effectively capturing its contextual variability. For each pair of synonyms, we manually select embeddings from TRACE where the words are used interchangeably in context.

\begin{figure}[tbp]
  \centering
  \begin{subfigure}[t]{.45\textwidth}
    \centering
    \hspace*{10pt}
    \includegraphics[width=\textwidth]{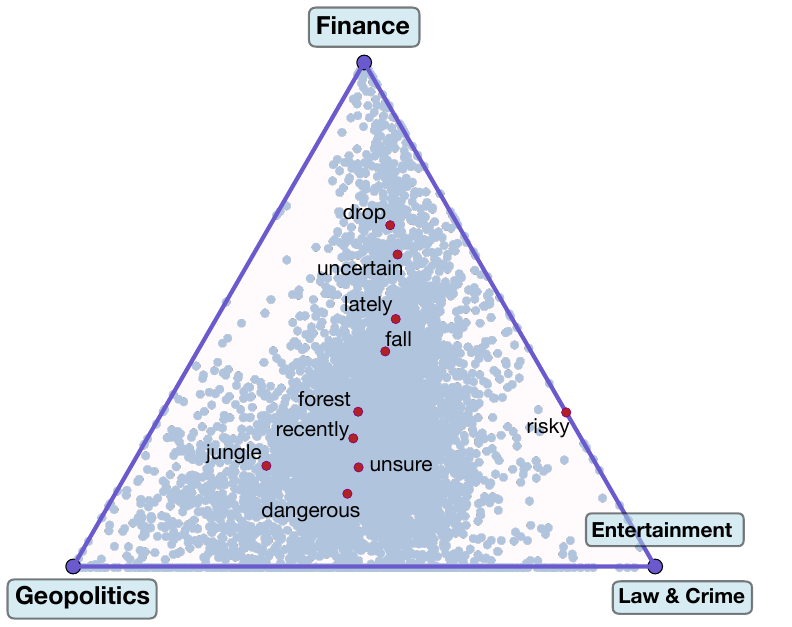} 
    \label{fig:synonymts}
    \vspace*{-20pt}
    \caption{Topic-SCORE.}
  \end{subfigure}~
  \begin{subfigure}[t]{.49\textwidth}
    \centering
    \hspace*{15pt}
    \includegraphics[width=\textwidth]{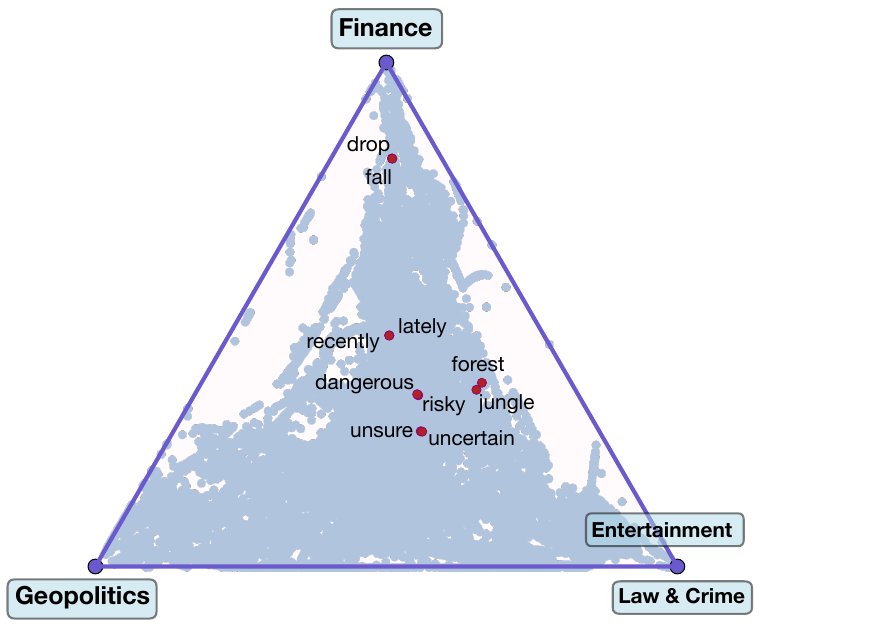} 
    \label{fig:synonymtrace}
    \vspace*{-20pt}
    \caption{TRACE.}
  \end{subfigure}
  \spacingset{1}
  \caption{Triangle plots showing the positions of synonyms from Topic-SCORE and TRACE.}
  \resetspacing
  \label{fig:synonyms}
\end{figure}

As shown in \cref{fig:synonyms}, embedding pairs from TRACE almost always overlap in the topic triangle, effectively capturing the semantic similarity between synonyms. By contrast, Topic-SCORE positions some synonym pairs far apart in the topic triangle. This discrepancy may stem from the low frequency of certain words or the averaging effect of representing each word with a single mixed-membership vector aggregated across contexts.

\spacingset{.8}
\begin{table}[tbp]
\centering
\caption{\small Top 50 anchor words for each topic of AP from Topic-SCORE.} \label{tb:AP-topic-score}
\scalebox{.92}{
\begin{tabular}{cl|p{14cm}}
\toprule
& {\small Topics} & {\small Top 50 anchor words for each topic of AP from Topic-SCORE} \\
\midrule
{\small 1}& {\small Finance}& {\footnotesize guilder, lire, franc, bullion, midmorning, troy, zurich, ounce, silver, yen, trader, profittaking, feeder, dollar, belly, pound, bullish, tokyo, gold, nikkei, unleaded, cent, dealer, future, clr, cdy, mercantile, oat, europe, fixed, bushel, currency, bid, london, witter, sentiment, commodity, traded, thursday, pressured, mixed, goldman, trading, anticipation, midday, stockindex, gallon, retreated}\\ 
\midrule
{\small 2}& {\small Geopolitics}& {\footnotesize gorbachev, ligachev, yeltsin, reformer, popov, mikhail, russian, boris, sakharov, ryzhkov, perestroika, baltic, citizenship, lithuania, preparation, republics, politburo, regan, sajudis, alexander, estonian, latvia, photo, ugly, waldheim, ussoviet, estonia, cardinal, stalin, lithuanian, oval, historian, socialism, vladimir, shevardnadze, jaruzelski, lukanov, stalin, kim, khrushchev, longrange, oberg, rust, delegate, kremlin, soviet, glasnost, embraced}\\ 
\midrule
{\small 3}& {\small Law \& Crime}& {\footnotesize clashed, knife, natal, masked, gunfire, belfast, police, nida, wounding, provincial, sikh, crawford, extremist, mafia, policemen, dhaka, gunshot, shooting, unite, suspicious, caliber, gunter, accidentally, gunman, injured, protestant, gang, irish, inkatha, bangladesh, bullet, shining, aftershock, avalanche, lima, wounded, clinic, village, ira, sikh, slogan, hindu, handgun, stranded, shot, patrol}\\ 
\midrule
{\small 4}& {\small Entertainment}& {\footnotesize homes, nbc, fox, cheer, primetime, nielsen, jackpot, abc, ranking, nest, fec, who, christie, wonder, sotheby, lottery, ptl, ames, interstate, khashoggi, drawing, chicken, bidder, satellite, pride, trustee, whirlpool, diamond, pick, coach, property, casino, pattern, million, monet, branch, sipc, dubbed, conrail, divisions, video, empty, ashland, poultry, pound, shell, cosmetics, secondquarter, fundraising}\\ 
\midrule
{\small 5}& {\small Geopolitics}& {\footnotesize qatar, torture, relevant, resolution, norway, ratification, gerard, herbert, sweden, united, arbitration, salvadoran, maurice, envoy, schwarzkopf, demonstrate, states, simon, kenneth, netherlands, mariel, redman, winner, cambodia, binding, trafficker, mulroney, britain, copyright, deported, montreal, extradited, khmer, desire, cartel, obligation, stone, unfortunately, detainee, harry, great, treaty, eec, beef, decide, subsidy, tariff, nobel, cuban}\\ 
\midrule
{\small 6}& {\small (Unknown)}& {\footnotesize averaged, durable, graduate, gallon, disagreed, arthritis, survey, mortgage, respondent, counted, adjusted, survey, sixmonth, poll, failure, averaging, depressed, revised, master, flow, robertson, user, percent, sample, adult, medicare, liability, oneyear, turnout, salary, wine, employed, smallest, machinery, offer, gnp, normal, magazine, builder, gte, error, select, fortune, margin, census, wage, pay, spite, cdc}\\ 
\midrule
{\small 7}& {\small (Unknown)}& {\footnotesize perry, bcspehealth, getz, steiger, impeachment, milstead, privacy, extortion, benedict, care, mecham, carpenter, patricia, ackerman, southwell, juror, jury, writes, abortion, fat, suing, mecham, nixon, husband, patronage, toussaint, chair, menorah, imagine, barry, parole, bro, joy, souter, mofford, abused, friedrick, gown, anne, itll, discrimination, yates, webb, circuit, nominate, defendant, breakfast, blier}\\ 
\bottomrule
\end{tabular}}
\end{table}
\resetspacing

\section{Supplementary Results for the MADStat Dataset} \label{supp:Algmadstat}

\subsection{Tuning Parameters and Number of Topics}

For the MADStat dataset, approximately \( 3,500,000 \) word embeddings are obtained following the preprocessing steps described in \cref{subsec:MADStat}. The number of hyperwords is set to \( M=2400 \), adhering to the practical recommendation in \cref{supp:tuningap}. We then randomly select \( 10\% \) of the word embeddings to train a UMAP model with parameters \( d=10 \), \texttt{n\_neighbors}=100, and \texttt{min\_dist}=0.1. The learned UMAP transformation is subsequently applied to project all embeddings in the corpus. As in previous experiments, we apply mini-batch k-means with \texttt{n\_init}=100 random initializations to partition the projected embeddings into \( M=2400 \) hyperwords.

\begin{figure}[htb]
  \centering
  \includegraphics[width=.75\textwidth]{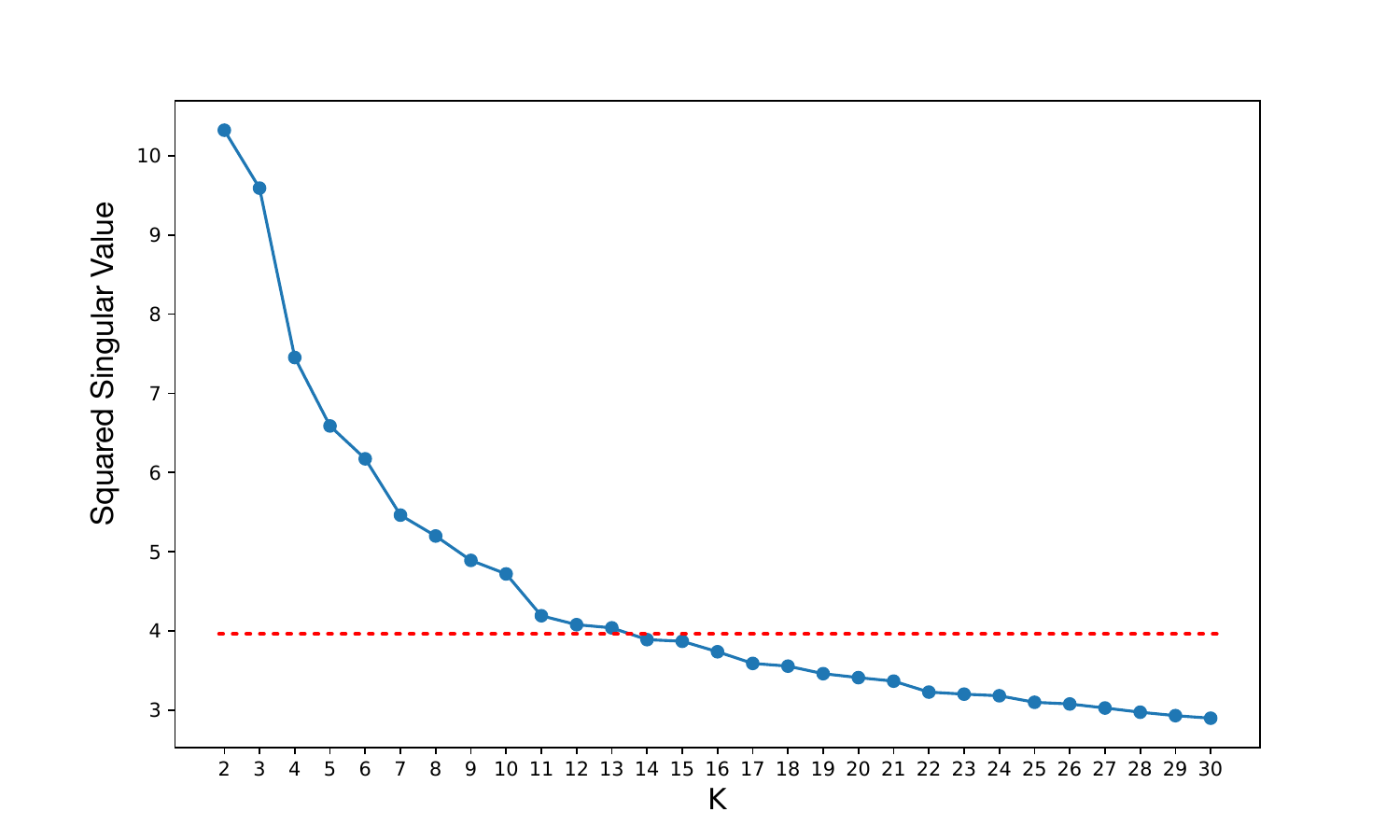} 
  \caption{Scree plot for singular values computed from the MADStat dataset.}
  \label{fig:screeplot2}
\end{figure}

\cref{fig:screeplot2} shows the scree plot for the singular values of the hyperword frequency matrix. In this case, noticeable drops in the singular values occur after \( K=3 \) and \( K=10 \). However, to demonstrate TRACE's ability to extract richer and finer-grained topic information compared to \cite{ke2023recent}, we set \( K=13 \) by thresholding the squared singular values at \( 4 \).

\begin{figure}[htb]
    \centering
    \begin{subfigure}{0.31\textwidth} 
        \centering
        \includegraphics[width=\linewidth]{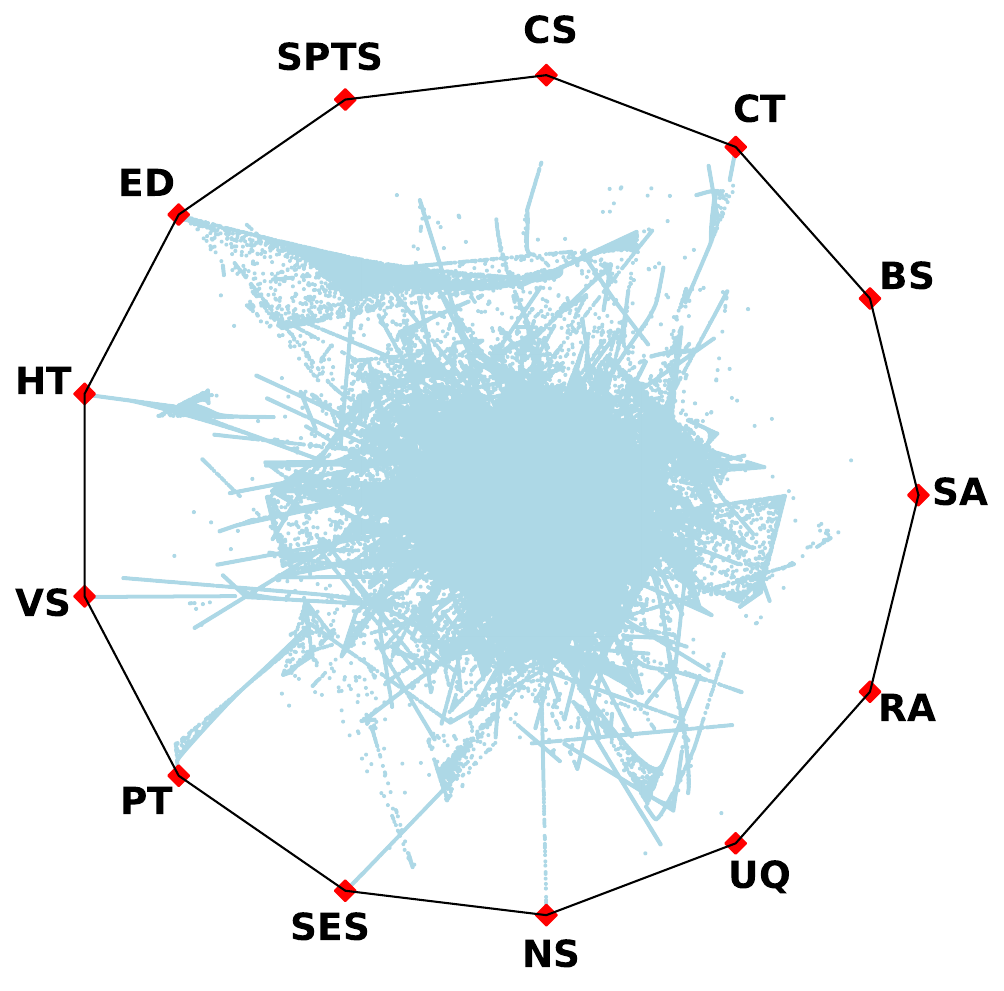} 
        \caption{bandwidth=0.02}
    \end{subfigure}
    \hfill
    \begin{subfigure}{0.31\textwidth}
        \centering
        \includegraphics[width=\linewidth]{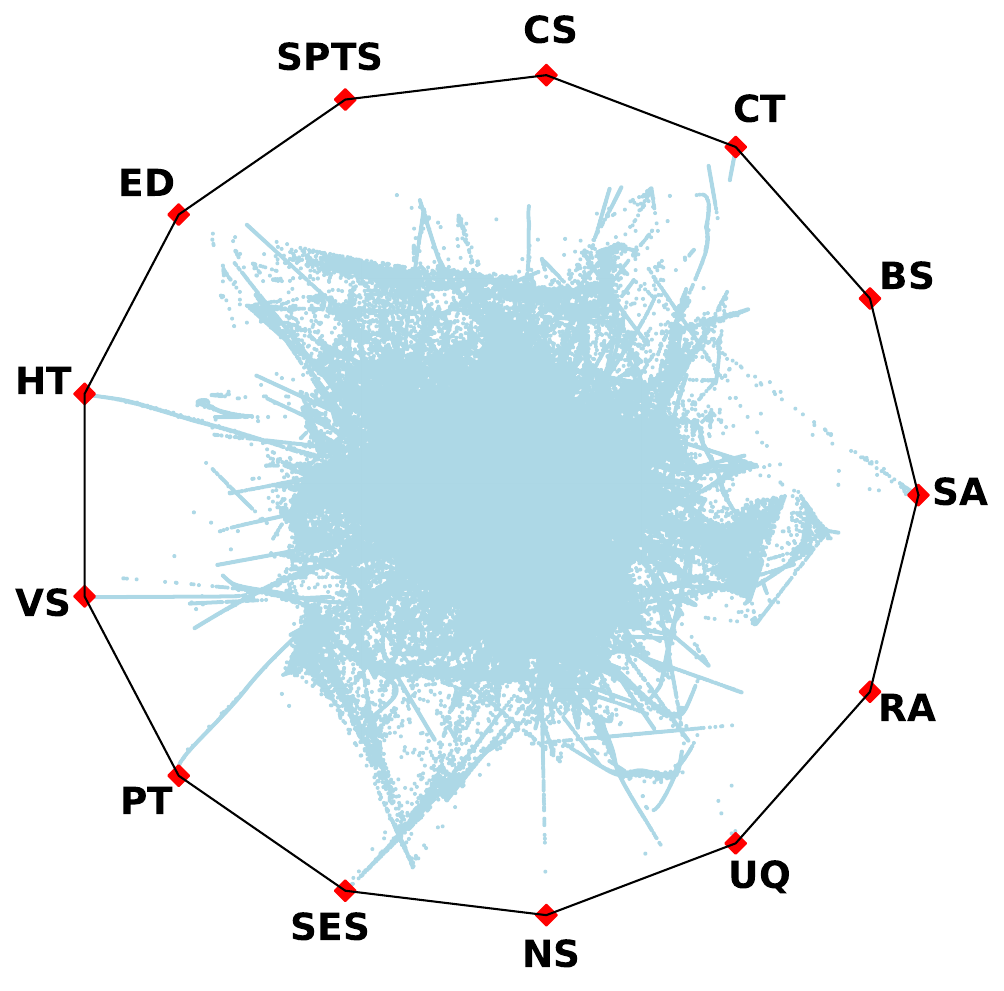}
        \caption{bandwidth=0.1}
    \end{subfigure}
    \hfill
    \begin{subfigure}{0.31\textwidth}
        \centering
        \includegraphics[width=\linewidth]{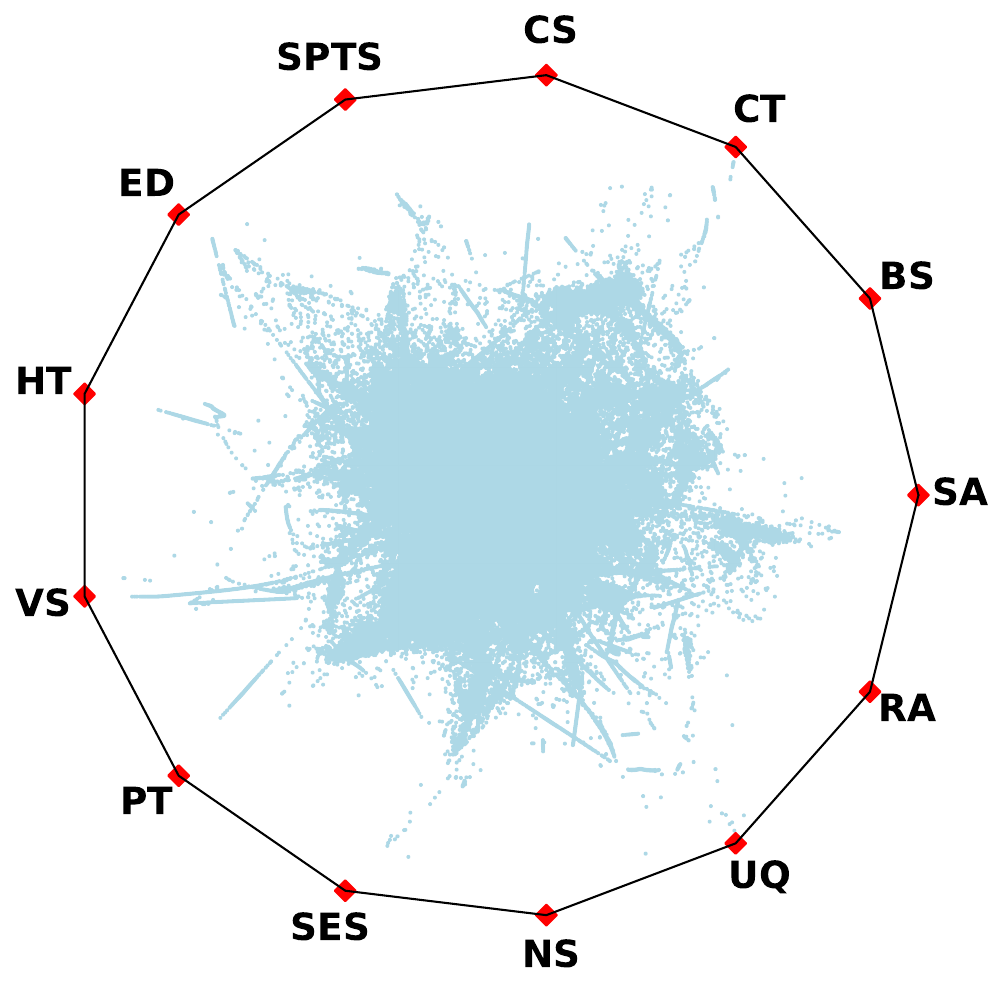}
        \caption{bandwidth=0.5}
    \end{subfigure}
    
    \caption{Comparison of tridecagon plots with different bandwidths.}
    \label{fig:SA_bandwidth}
\end{figure}

Lastly, the bandwidth is selected as \( 0.1 \) using the maximum entropy criterion introduced in \cref{supp:entropy}. \cref{fig:SA_bandwidth} visualizes the mixed-membership vectors for word embeddings under different bandwidths (\( 0.02 \), \( 0.1 \), and \( 0.5 \)), highlighting the impact of this parameter on the results.

\subsection{Additional Results}

\begin{sidewaysfigure}
\centering
\hspace*{40pt}
\includegraphics[width=.95\textwidth]{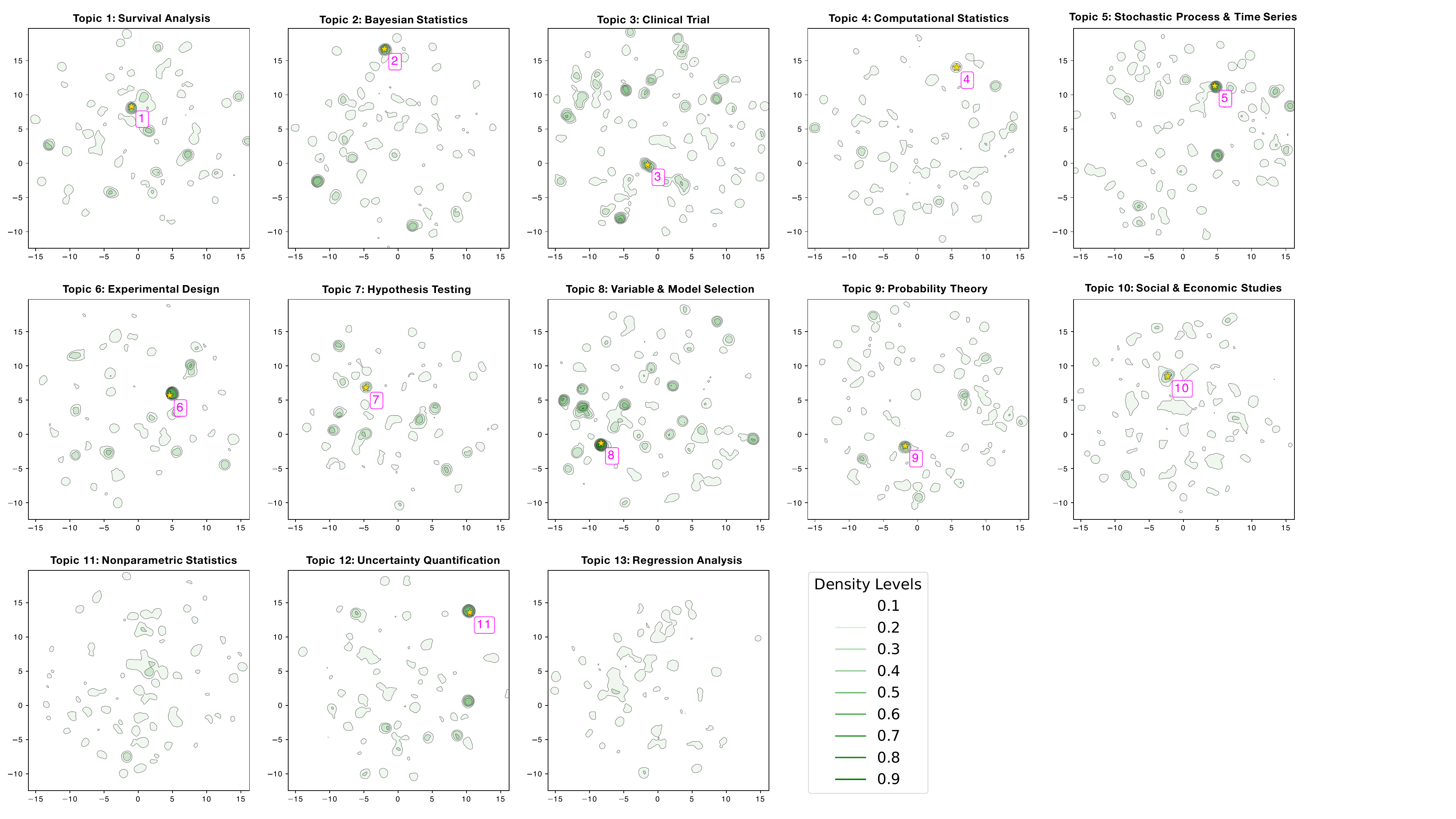}
\spacingset{1}
\caption{Estimated topic measures on the MADStat dataset ($K=13$). For each topic, we compute $\widehat{\cal B}_k(z)=\widehat{\cal A}_k(z)/[\sum_{\ell=1}^K \widehat{\cal A}_\ell(z)]$ and plot its contour in a projected two-dimensional space (the projection is only made for visualization). Eleven anchor regions are marked in the plots, each indicating a group of nearly-anchor words. Some representative words in each numbered region are given in the plots.} \label{fig:MADStat-topics}
\resetspacing
\end{sidewaysfigure}

Similar to \cref{fig:AP-topics} in the main paper, we visualize the contours of the re-normalized topic densities \( \widehat{\mathcal{B}}_1(\cdot), \ldots, \widehat{\mathcal{B}}_{13}(\cdot) \) for the MADStat dataset in \cref{fig:MADStat-topics}. The embeddings are distributed across multiple anchor regions within each topic, and these regions often differ significantly between topics. Some of the anchor regions are labeled in \cref{fig:MADStat-topics}, and their 15 representative words are listed in \cref{tb:MADStat-region}.

\cref{tb:MADStat-trace} presents the top 50 anchor words for each topic identified by TRACE in \cref{subsec:MADStat}. Most words are unique to their corresponding topics, demonstrating strong topic coherence. A few words, however, appear across multiple topics, reflecting TRACE's flexibility in leveraging rich contextual information from language models. For instance, the word \textit{cure} is characteristic of both \textit{Survival Analysis} and \textit{Clinical Trials}, while the word \textit{variance} is prevalent in both \textit{Nonparametric Statistics} and \textit{Regression Analysis}.

\begin{figure}[tbp]
  \centering
  \begin{subfigure}[t]{1.0\textwidth}
    \centering
    \includegraphics[width=\textwidth,height=.4\textheight]{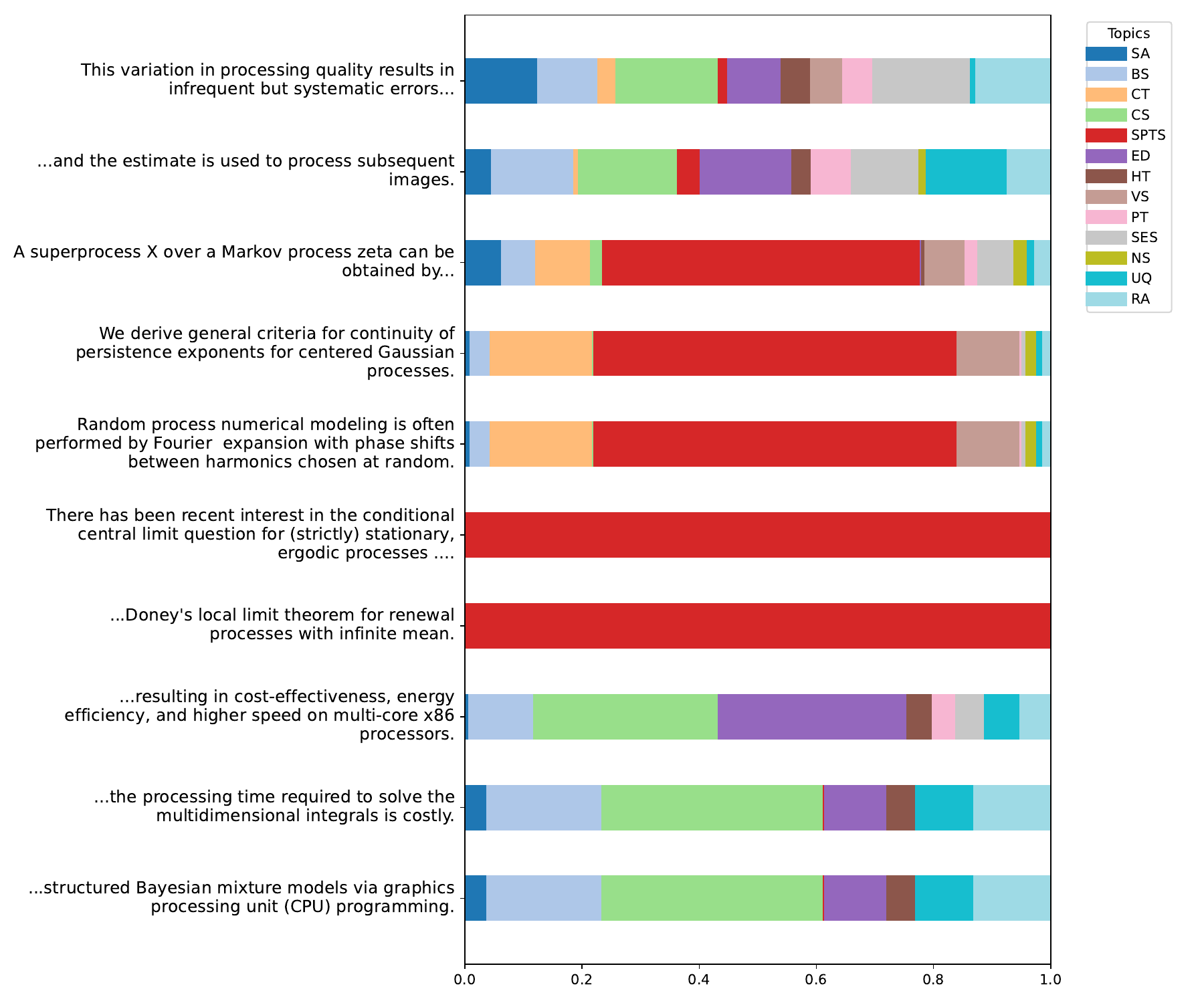} 
    \label{fig:madstatprocess}
    \vspace*{-30pt}
    \caption{\emph{Process}.}
  \end{subfigure}

  \vspace*{20pt}
  
  \begin{subfigure}[t]{1.0\textwidth}
    \centering
    \includegraphics[width=\textwidth,height=.4\textheight]{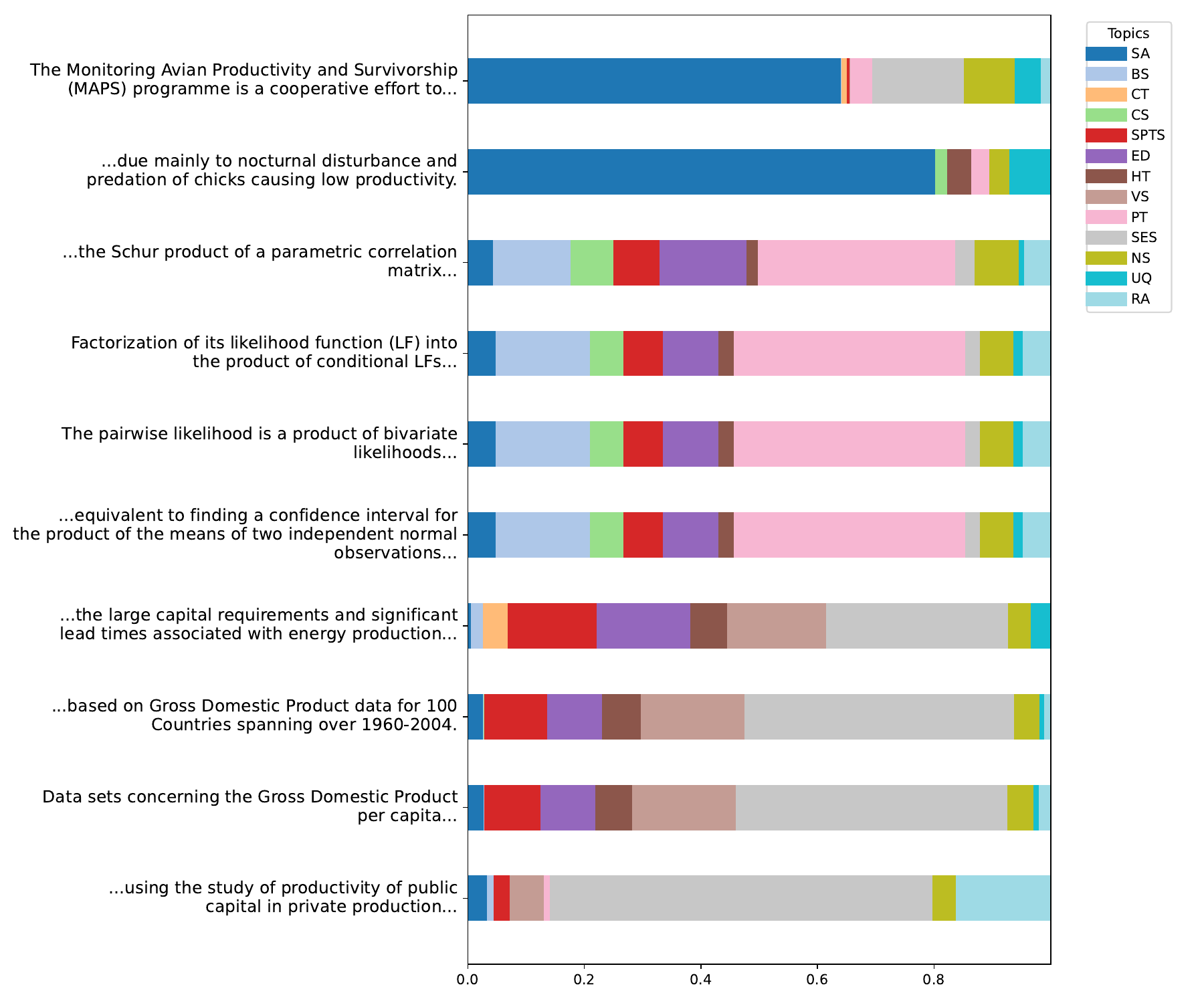} 
    \label{fig:madstatproduct}
    \vspace*{-30pt}
    \caption{\emph{Product}.}
  \end{subfigure}
  \caption{Mixed-membership vectors for example words in different contexts of MADStat.}
  \label{fig:barplots}
\end{figure}

Understanding contextual information is essential, as common anchor words may carry entirely different meanings across topics. \cref{fig:barplots} provides examples of two such words, \textit{process} and \textit{product}, each analyzed with ten contexts. The word \textit{process}, for example, has dominating weights in \textit{Stochastic Process \& Time Series} when referring to a random sequence under study, as in \textit{renewal processes}, \textit{ergodic processes}, or \textit{Markov processes}. However, it shows a significant weight in \textit{Computational Statistics} when used in the context of \textit{processing units}. Additionally, it can distribute relatively evenly across multiple topics when appearing in phrases like \textit{processing results} or \textit{process images}. Similarly, the word \textit{product} holds its largest weight in \textit{Probability Theory} when used as a noun denoting the multiplication operation. In contrast, its derivative, \textit{productivity}, shows higher weights in \textit{Survival Analysis} or \textit{Social \& Economic Studies}, reflecting its usage in those contexts.  

Finally, \cref{tb:MADStat-doc} presents the titles of five representative documents for each topic in the MADStat dataset as identified by TRACE, illustrating the clarity and coherence of the topics. These documents are selected based on the rankings of the largest components in their mixed-membership vectors. Upon examination, each document strongly aligns with its assigned topic, underscoring the effectiveness of TRACE in capturing the distinct thematic structure of the dataset. Additionally, the diversity within these examples highlights TRACE's ability to identify meaningful and well-separated topics in complex, real-world datasets.

\clearpage

\clearpage

\spacingset{.8}
\begin{table}[htb!]
\centering
\caption{Top 50 anchor words for each topic of MADStat from TRACE.}
\label{tb:MADStat-trace}
\scalebox{.8}{
\begin{tabular}{cl|p{17cm}}
\toprule
& {\small Topics} & {\small Top 50 anchor words for each topic of MADStat from TRACE} \\
\midrule
{\small 1} & {\small \parbox[t]{3cm}{Survival\\Analysis}} &
{\footnotesize cure, survival, fitness, survive, reproduction, nest, immunity, hazard, frailty, proportional, stress, failure, reproductive, defect, censor, cause, cox, scale, crude, baseline, transplant, violation, shock, fail, event, damage, condition, transplantation, conservation, tumor, time, radiation, cumulative, cancer, leukemia, lung, colon, prostate, transplanted, breast, reduce, augmented, delay, second, accelerate, life, lifetime, vital, die, death} \\ 
\midrule
{\small 2} & {\small \parbox[t]{3cm}{Bayesian\\Statistics}} &
{\footnotesize posterior, hypergeometric, hyperparameter, bayesian, prior, bayes, brownian, reference, dirichlet, default, subjective, gibbs, sampler, mcmc, msc, svm, cdf, frequentist, frequent, lead, intrinsic, predictive, mixture, predictiveness, opinion, alarm, monte, belief, inherent, chain, jump, objective, standard, principle, hierarchical, hierarchy, representative, carlo, divergence, diverge, pseudo, divergent, invalid, cauchy, inappropriate, point, perspective, augmentation, augment, unreliable} \\ 
\midrule
{\small 3} & {\small \parbox[t]{3cm}{Clinical\\Trial}} &
{\footnotesize trial, clinical, placebo, interim, randomize, randomization, safety, compliance, clinician, therapeutic, phase, vaccine, efficacy, forest, arm, endpoint, treatment, therapy, ethical, practical, benefit, medication, treat, patient, physician, contribution, cure, protocol, participant, management, clinic, interior, pharmaceutical, drug, antiretroviral, med, generic, surrogate, assignment, intervention, compound, course, stage, study, replacement, propensity, prevention, effectiveness, adverse, administration} \\ 
\midrule
{\small 4} & {\small \parbox[t]{3cm}{Computational\\Statistics}} &
{\footnotesize svm, machine, learning, learn, iteration, boost, pca, optimization, round, descent, differentiation, multiplication, algorithm, classifier, gradient, algorithmic, electronic, online, complexity, computer, training, train, collaborative, noisy, reconstruction, induction, burden, software, package, code, programming, program, language, source, routine, environment, write, version, solve, heuristic, histogram, heaviness, computational, computing, quick, simple, fast, problem, best, strong} \\ 
\midrule
{\small 5} & {\small \parbox[t]{3.8cm}{Stochastic Process\\\& Time Series}} &
{\footnotesize process, flow, reaction, levy, stationary, diffusion, jump, disturbance, persistent, drift, transfer, motion, movement, transition, wiener, move, acceleration, ergodic, ergodicity, stable, unstable, drive, martingale, brownian, predictable, cycle, intensity, stationarity, spectral, fourier, harmonic, pathway, path, span, trajectory, stochastic, conservative, arma, stochasticity, hilbert, index, memory, regular, continuous, operator, differential, innovation, continual, point, noise} \\ 
\midrule
{\small 6} & {\small \parbox[t]{3cm}{Experimental\\Design}} &
{\footnotesize design, plan, layout, designer, universal, mutual, block, latin, frame, factorial, balanced, resolution, drawing, light, balance, band, spiked, orthogonal, diagonal, diagonalization, box, array, project, factorization, factorize, reflect, grade, optimality, factor, optimal, incomplete, lacking, incompleteness, fractional, elemental, nest, nesting, brownian, replication, replicate, cross, experimental, neighbor, fraction, square, blind, ideal, row, entry, column} \\ 
\midrule
{\small 7} & {\small \parbox[t]{3cm}{Hypothesis\\Testing}} &
{\footnotesize test, examination, check, testing, alternative, null, wilcoxon, hypothesis, option, powerful, power, rank, precedence, possibility, homogeneity, homoscedastic, homogenous, wald, candidate, rating, substitute, supplement, hypothesize, neyman, sign, mise, spectra, heuristic, bonferroni, mahalanobis, lemma, hotelling, omnibus, side, redundant, structure, selective, pitman, ranking, versus, rate, score, alternate, significance, validation, pearson, verification, type, scoring, ratio, symmetry} \\ 
\midrule
{\small 8} & {\small \parbox[t]{3.2cm}{Variable/Model\\Selection}} &
{\footnotesize selection, variable, selectivity, selector, lasso, auc, bic, aic, akaike, decide, select, choose, choice, model, oracle, penalty, tune, predictor, predictable, procedure, protocol, shrinkage, criterion, fdr, routine, subset, operation, option, case, predictive, candidate, landmark, process, device, penalize, penalization, average, prediction, applicant, principle, influential, simultaneous, false, guarantee, discovery, joint, true, control, coefficient, testing} \\ 
\midrule
{\small 9} & {\small \parbox[t]{3cm}{Probability\\Theory}} &
{\footnotesize variable, vector, tensor, identical, random, attraction, distribute, spaced, disperse, spread, scatter, integer, inequality, lattice, constant, independent, sum, product, number, lemma, conjecture, mutual, positive, negative, field, dependent, element, exclusive, series, value, sequence, eigenvalue, eigenvector, efron, equivariant, moment, origin, scale, scaling, randomness, equivariance, array, union, moment, walk, definite, symmetric, exchangeability, exchangeable, infinite} \\ 
\midrule
{\small 10} & {\small \parbox[t]{3.2cm}{Social \& Economic\\Studies}} &
{\footnotesize state, country, nation, community, world, united, develop, north, union, european, american, england, british, international, national, regional, city, income, county, use, usage, utilization, consumption, consumer, occupational, unemployment, firm, household, force, course, public, employment, political, management, manager, institution, government, university, nutritional, food, unemployed, nutrition, institutional, generation, year, customer, day, productivity, season, hour} \\ 
\midrule
{\small 11} & {\small \parbox[t]{3.2cm}{Nonparametric\\Statistics}} &
{\footnotesize minimax, square, mean, mise, mse, loss, mle, bandwidth, ridge, least, ordinary, contamination, contaminate, error, gain, order, violation, irregularity, absolute, density, kernel, bias, maximize, maximizer, minimize, minimizer, conservation, integrate, embed, asymptotic, variance, unknown, unspecified, asymmetric, asymmetry, ascertain, hill, derivative, mars, expression, acyclic, parameter, lose, isotonic, rate, smoothness, influence, slope, dominate, uncertain} \\ 
\midrule
{\small 12} & {\small \parbox[t]{3.5cm}{Uncertainty\\Quantification}} &
{\footnotesize tolerance, coverage, cover, interval, confidence, credible, belief, bootstrap, pivotal, matern, crude, percentile, blind, region, band, banded, hierarchy, nominal, digital, jackknife, jackknifing, rational, conservative, conservativeness, regional, calibration, calibrate, construct, form, design, width, small, wide, develop, naive, volume, narrow, simultaneous, empirical, domain, area, formulate, method, coefficient, credibility, quantile, edgeworth, size, form, short} \\ 
\midrule
{\small 13} & {\small \parbox[t]{3.5cm}{Regression\\Analysis}} &
{\footnotesize gee, square, miss, intercept, missingness, lose, statement, measurement, omit, measure, defective, actual, auxiliary, error, violation, irregularity, imputation, impute, least, variance, imprecise, misspecification, misclassification, mix, ordinary, logistic, linear, random, ridge, linearly, linearity, linearization, linearize, response, reaction, jackknife, jackknifing, respond, instrumental, transform, correct, effect, compensate, covariance, calibration, calibrate, bias, respondent, ordinal, ordinality} \\ 
\bottomrule
\end{tabular}}
\end{table}

\begin{table}[htb!]
\centering
\caption{15 representative words for each labeled region from TRACE.} \label{tb:MADStat-region}
\scalebox{.92}{
\begin{tabular}{l|p{12cm}}
\toprule
{\small Regions} & {\small 15 representative words for each region of \cref{fig:MADStat-topics} from TRACE} \\
\midrule
{\small 1 (Survival Analysis)} & {\footnotesize frailty, diagnostic, die, survival, compatibility, variable, brain, parent, latent, evolutionary, living, fertility, behavioral, process, healthy}\\ 
\midrule
{\small 2 (Bayesian Statistics)}& {\footnotesize characterize, posterior, justify, distribution, classify, connect, characterization, variable, conceptualize, describe, character, illustrate}\\ 
\midrule
{\small 3 (Clinical Trial)}& {\footnotesize treated, rank, mean, cure, medication, replacement, therapeutic, base, upper, none, background, topical, systematic, treatment, matrix}\\ 
\midrule
{\small 4 (Computational Statistics)}& {\footnotesize advanced, convenient, ideal, computable, open, objective, fast, cent, penalty, algorithm, computer, curve, oriented, able, aided}\\ 
\midrule
{\small 5 (Stochestic Process \& Time Series)}& {\footnotesize mixing, cycle, filter, parameterize, trait, proxy, step, process, transition, descriptive, reaction, pathway, constant, flow, diagnostic}\\ 
\midrule
{\small 6 (Experimental Design)}& {\footnotesize common, frequently, generalize, design, vast, criterion, term, tend, subject, quick, framework, rarely, recurrence, general, often}\\ 
\midrule
{\small 7 (Hypothesis Testing)}& {\footnotesize interval, simulation, derivation, consistent, far, rest, methodology, lattice, parameter, markov, examination, test, applied, effect, unit}\\ 
\midrule
{\small 8 (Variable/Model Selection)}& {\footnotesize prior, selection, justify, inverse, unit, unify, wilks, consist, extend, biomarker, selection, agreement, design, harmonization, nomination}\\ 
\midrule
{\small 9 (Probability Theory)}& {\footnotesize believe, state, summary, outline, express, point, demonstrate, pointed, notice, validate, reveal, describe, observe, verify, trace}\\ 
\midrule
{\small 10 (Social \& Economic Studies)}& {\footnotesize share, portfolio, account, stock, trade, asset, opportunity, rate, market, methodology, price, analytic, demand, offer, device}\\ 
\midrule
{\small 11 (Uncertainty Quantification)}& {\footnotesize confidence, wilks, throughout, across, credibility, within, span, weibull, along, ratio, belief, apart, iterated, term, wald}\\ 
\bottomrule
\end{tabular}}
\end{table}
\resetspacing

\clearpage

\clearpage

\spacingset{.8}
\begin{table}[tbp]
\centering
\caption{\small Titles of five representative documents for each topic in MADStat from TRACE.}
\label{tb:MADStat-doc}
\scalebox{.65}{
\begin{tabular}{cl|p{17.5cm}}
\toprule
&{\small Topics} & {\small Titles of five representative documents for each topic in MADStat from TRACE} \\
\midrule
1 & {\small \parbox[t]{2.5cm}{Survival\\Analysis}} &
{\small Survival Analysis Under Cross-sectional Sampling: Length Bias And Multiplicative Censoring\newline 
Modelling Postfledging Survival And Age-specific C Breeding Probabilities In Species With Delayed Maturity\newline
 Multiple Imputation Methods For Modelling Relative Survival Data\newline
 Modeling Heterogeneity In Nest Survival Data\newline
 A Martingale Approach To The Copula-graphic Estimator For The Survival Function Under Dependent Censoring}\\ 
\midrule
2 & {\small \parbox[t]{2.5cm}{Bayesian\\Statistics}} &
{\small Sensitivity Of Some Standard Bayesian Estimates To Prior Uncertainty---A Comparison\newline
 A Note On Nonexistence Of Posterior Moments\newline
 Implications Of Reference Priors For Prior Information And For Sample Size\newline
 A Note On Noninformative Priors For Weibull Distributions\newline
 Infinitesimal Sensitivity Of Posterior Distributions}\\ 
\midrule
3& {\small \parbox[t]{2.5cm}{Clinical\\Trial}} &
{\small Data Monitoring Committees---The Moral Case For Maximum Feasible Independence\newline
 Up-front Versus Sequential Randomizations For Inference On Adaptive Treatment Strategies\newline
 Data Monitoring Boards In The Pharmaceutical-industry\newline
 How Study Design Affects Outcomes In Comparisons Of Therapy. II: Surgical\newline
 Estimation Of Treatment Effect For The Sequential Parallel Design}\\ 
\midrule
4& {\small \parbox[t]{2.5cm}{Computational\\Statistics}} &
{\small A New Approximate Maximal Margin Classification Algorithm\newline
 Simulated Stochastic Approximation Annealing For Global Optimization With A Square-root Cooling Schedule\newline
 The Interplay Of Optimization And Machine Learning Research\newline
 Algorithms For Sparse Linear Classifiers In The Massive datasetting\newline
 Languages For Statistics And Data Analysis}\\ 
\midrule
5& {\small \parbox[t]{2.5cm}{Stochastic\\ Process\\\& Time Series}} &
{\small Spectral Representations Of Sum- And Max-stable Processes\newline
 Levy-driven Carma Processes\newline
 Spatial Autoregressive And Moving Average Hilbertian Processes\newline
 Levy Processes Conditioned On Having A Large Height Process\newline
 Euler(p,q) Processes And Their Application To Non Stationary Time Series With Time Varying Frequencies}\\ 
\midrule
6& {\small \parbox[t]{2.5cm}{Experimental\\Design}} &
{\small Partially Efficiency Balanced Designs\newline
 Alias balanced and alias partially balanced fractional $2^{m}$ factorial designs of resolution $2l+1$\newline
 Construction Of Group Divisible Designs And Rectangular Designs From Resolvable And Almost Resolvable Balanced Incomplete Block Designs\newline
 Development Of Research In Experimental Design In India\newline
 A Trigonometric Approach To Quaternary Code Designs With Application To One-eighth And One-sixteenth Fractions}\\ 
\midrule
7& {\small \parbox[t]{2.5cm}{Hypothesis\\Testing}} &
{\small A Consensus Combined P-value Test And The Family-wide Significance Of Component Tests\newline
 Linear Rank Tests For Independence In Bivariate Distributions---Power Comparisons By Simulation\newline
 Non-parametric Change-point Tests For Long-range Dependent Data\newline
 Smooth Tests Of Goodness Of Fit---An Overview\newline
 A Test For Detecting Outlying Cells In The Multinomial Distribution And 2-way Contingency-tables}\\ 
\midrule
8& {\small \parbox[t]{2.5cm}{Variable/Model\\Selection}} &
{\small The Influence Of Variable Selection---A Bayesian Diagnostic Perspective\newline
 Stability Selection\newline
 Some Models Of Genetic Selection\newline
 Influential Data Cases When The $C_{p}$ Criterion Is Used For Variable Selection In Multiple Linear Regression\newline
 Variable Selection In Model-based Discriminant Analysis}\\ 
\midrule
9& {\small \parbox[t]{2.5cm}{Probability\\Theory}} &
{\small Strong Law Of Large Numbers For 2-exchangeable Random Variables\newline
 Some Asymptotic Results On Extremes Of Incomplete Samples\newline
 Joint Limit Laws Of Sample-moments Of A Symmetric Distribution\newline
 A Characterization Of Joint Distribution Of Two-valued Random Variables And Its Applications\newline
 From Moments Of Sum To Moments Of Product}\\ 
\midrule
10& {\small \parbox[t]{2.5cm}{Social \&\\ Economic\\ Studies}} &
{\small Comovements In Stock-prices In The Very Short Run\newline
 Sovereign Credit Ratings, Market Volatility, And Financial Gains\newline
 A Conversation With Morton Kramer\newline
 Official Labor Statistics---A Historical-perspective\newline
 A Conversation With V. P. Godambe}\\ 
\midrule
11& {\small \parbox[t]{2.5cm}{Nonparametric\\Statistics}} &
{\small On Shrinking Minimax Convergence In Nonparametric Statistics\newline
 Optimal Smoothing In Adaptive Location Estimation\newline
 Asymptotics For The Transformation Kernel Density Estimator\newline
 Beta Kernel Smoothers For Regression Curves\newline
 Optimum Kernel Estimators Of The Mode}\\ 
\midrule
12& {\small \parbox[t]{2.5cm}{Uncertainty\\Quantification}} &
{\small Conservative Confidence Intervals For A Single Parameter\newline
 The Monotone Boundary Property And The Full Coverage Property Of Confidence Intervals For A Binomial Proportion\newline
 Exact Confidence Coefficients Of Simultaneous Confidence Intervals For Multinomial Proportions\newline
 Some Simple Corrected Confidence Intervals Following A Sequential Test\newline
 More On Shortest And Equal Tails Confidence Intervals}\\ 
\midrule
13& {\small \parbox[t]{2.5cm}{Regression\\Analysis}} &
{\small On Small Sample Properties Of The Mixed Regression Predictor Under Misspecification\newline
 Polynomial Regression With Errors In The Variables\newline
 Doubly Robust Estimates For Binary Longitudinal Data Analysis With Missing Response And Missing Covariates\newline
 Some Effects Of Ignoring Correlated Measurement Errors In Straight-line Regression And Prediction\newline
 Consistent Estimation In An Implicit Quadratic Measurement Error Model}\\ 
\bottomrule
\end{tabular}}
\end{table}
\resetspacing

\spacingset{1.2}

\bibliography{topic}
\bibliographystyle{chicago}

\end{document}